\documentclass[aps,pra,reprint,amsmath,amssymb,superscriptaddress,onecolumn,longbibliography, notitlepage,nofootinbib]{revtex4-1}

\usepackage{mathtools}
\usepackage{braket}
\usepackage{siunitx}
\usepackage{listings}
\usepackage[hidelinks]{hyperref}

\usepackage{xparse}

\NewDocumentCommand{\codeword}{v}{%
\texttt{\textcolor{black}{#1}}%
}

\usepackage{xcolor}

\begin{document}

\title{Deep physical neural networks enabled by a backpropagation algorithm for arbitrary physical systems}
\author{Logan~G.~Wright}
\thanks{Equal contribution}
\email{lgw32@cornell.edu; to232@cornell.edu}
\affiliation{School of Applied and Engineering Physics, Cornell University, Ithaca, NY 14853, USA}
\affiliation{NTT Physics and Informatics Laboratories, NTT Research, Inc., Sunnyvale, CA 94085, USA}
\author{Tatsuhiro~Onodera}
\thanks{Equal contribution}
\email{lgw32@cornell.edu; to232@cornell.edu}
\affiliation{School of Applied and Engineering Physics, Cornell University, Ithaca, NY 14853, USA}
\affiliation{NTT Physics and Informatics Laboratories, NTT Research, Inc., Sunnyvale, CA 94085, USA}
\author{Martin~M.~Stein}
\affiliation{School of Applied and Engineering Physics, Cornell University, Ithaca, NY 14853, USA}
\author{Tianyu~Wang}
\affiliation{School of Applied and Engineering Physics, Cornell University, Ithaca, NY 14853, USA}
\author{Darren~T.~Schachter}
\affiliation{School of Electrical and Computer Engineering, Cornell University, Ithaca, NY 14853, USA}
\author{Zoey~Hu}
\affiliation{School of Applied and Engineering Physics, Cornell University, Ithaca, NY 14853, USA}
\author{Peter~L.~McMahon}
\email{pmcmahon@cornell.edu}
\affiliation{School of Applied and Engineering Physics, Cornell University, Ithaca, NY 14853, USA}

\begin{abstract}
Deep neural networks have become a pervasive tool in science and engineering. However, modern deep neural networks' growing energy requirements now increasingly limit their scaling and broader use. We propose a radical alternative for implementing deep neural network models: \textit{Physical Neural Networks}. We introduce a hybrid physical-digital algorithm called \textit{Physics-Aware Training} to efficiently train sequences of controllable physical systems to act as deep neural networks. This method automatically trains the functionality of any sequence of real physical systems, directly, using backpropagation, the same technique used for modern deep neural networks. To illustrate their generality, we demonstrate physical neural networks with three diverse physical systems---optical, mechanical, and electrical. Physical neural networks may facilitate unconventional machine learning hardware that is orders of magnitude faster and more energy efficient than conventional electronic processors.
\end{abstract}

\maketitle

\section{Introduction}

Like many historical developments in artificial intelligence \cite{brooks1991intelligence,hooker2020hardware}, the widespread adoption of deep neural networks (DNNs) was enabled in part by synergistic hardware. In 2012, building on numerous earlier works, Krizhevsky et al. showed that the backpropagation algorithm for stochastic gradient descent (SGD) could be efficiently executed with graphics-processing units to train large convolutional DNNs \cite{krizhevsky2012imagenet} to perform accurate image classification. Since 2012, the breadth of applications of DNNs has expanded, but so too has their typical size. As a result, the computational requirements of DNN models have grown rapidly, outpacing Moore's Law \cite{schwartz2019green,patterson2021carbon}. Now, DNNs are increasingly limited by hardware energy efficiency. 

The emerging DNN energy problem has inspired special-purpose hardware: DNN `accelerators' \cite{reuther2020survey,markovic2020physics}. Several proposals push beyond conventional electronics with alternative physical platforms \cite{markovic2020physics}, such as optics \cite{wetzstein2020inference,shen2017deep,hamerly2019large, shastri2021photonics} or memristor crossbar arrays \cite{prezioso2015training,hu2014memristor}. These devices rely on approximate analogies between the hardware physics and the mathematical operations in DNNs (Fig.~1A). Consequently, their success will depend on intensive engineering to push device performance toward the limits of the hardware physics, while carefully suppressing parts of the physics that violate the analogy, such as unintended nonlinearities, noise processes, and device variations.

More generally, however, the controlled evolutions of physical systems are well-suited to realizing deep learning models. DNNs and physical processes share numerous structural similarities, such as hierarchy, approximate symmetries, redundancy, and nonlinearity. These structural commonalities explain much of DNNs' success operating robustly on data from the natural, physical world \cite{lin2017does}. As physical systems evolve, they perform, in effect, the mathematical operations within DNNs: controlled convolutions, nonlinearities, matrix-vector operations and so on. We can harness these physical computations by encoding input data into the initial conditions of the physical system, then reading out the results by performing measurements after the system evolves (Fig.~1C). Physical computations can be controlled by adjusting physical parameters. By cascading such controlled physical input-output transformations, we can realize trainable, hierarchical physical computations: deep physical neural networks (PNNs, Fig.~1D). As anyone who has simulated the evolution of complex physical systems appreciates, physical transformations are typically faster and consume less energy than their digital emulations: processes which take nanoseconds and nanojoules frequently require seconds and joules to digitally simulate. PNNs are therefore a route to scalable, energy-efficient, and high-speed machine learning. 

\begin{figure}
    \centering
    \includegraphics[width=0.95\textwidth]{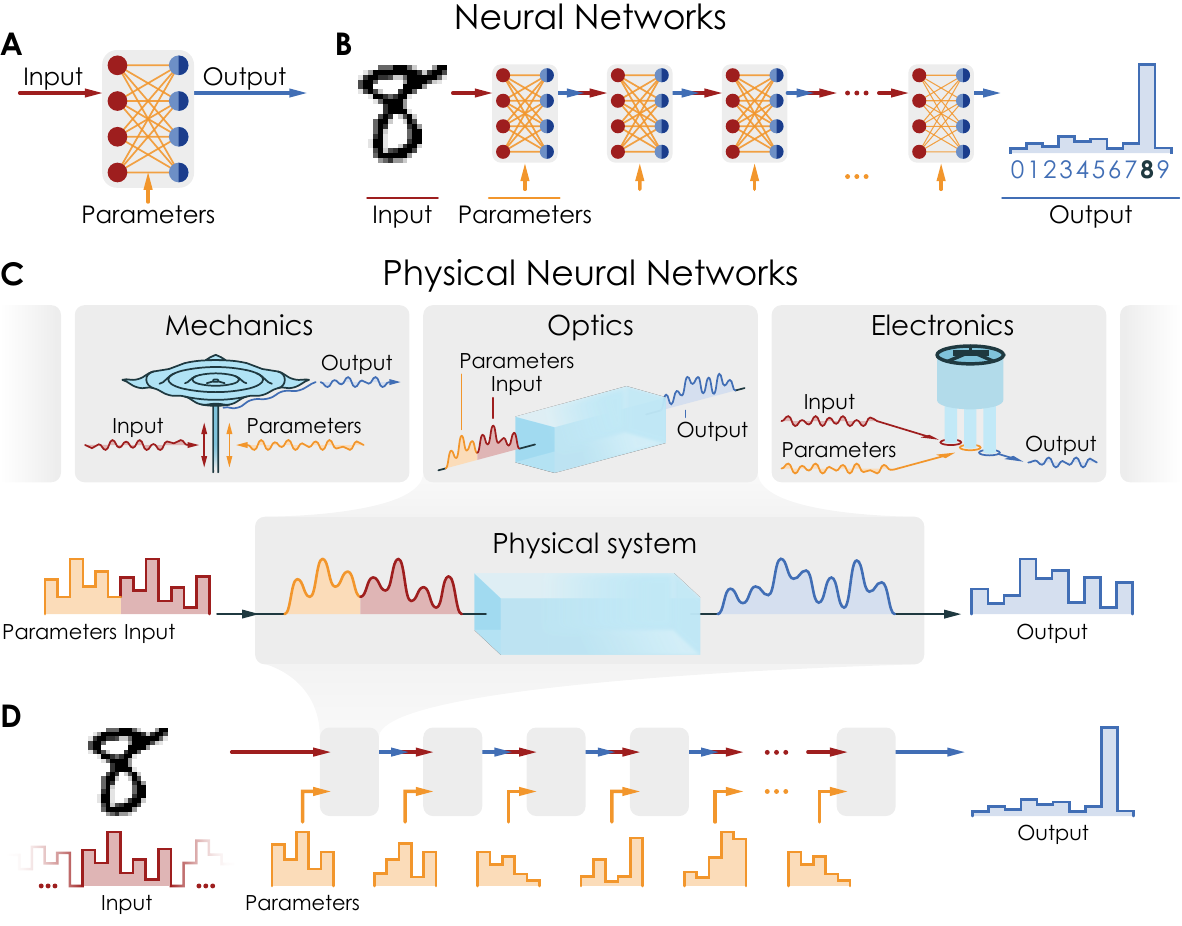}
    \caption{\textbf{Introduction to physical neural networks.} \textbf{A.} Conventional artificial neural networks  are built from an operational unit (the layer) that involves a trainable matrix-vector multiplication followed by element-wise nonlinear activation such as the rectifier (ReLU). The weights of the matrix-vector multiplication are adjusted during training so that the ANN implements a desired mathematical operation. \textbf{B.} By cascading these operational units together, one creates a deep neural network (DNN), which can learn a multi-step (hierarchical) computation. \textbf{C.} When physical systems evolve, they perform computations. We can partition the controllable properties of the system into input data, and control parameters. By changing control parameters, we can alter the physical transformation performed on the input data. In this paper, we consider three examples. In a mechanical system, input data and parameters are encoded into the time-dependent force applied to a suspended metal plate. The physical computation results in an input- and parameter-dependent multimode oscillation pattern, which is read out by a microphone. In a nonlinear optical system, a pulse passes through a nonlinear $\chi^{(2)}$ medium, producing a frequency-doubled output. Input data and parameters are encoded in the amplitude of frequency components of the pulse, and outputs are obtained from the frequency-doubled pulse's spectrum. In an electronic system, analog signals encode input data and parameters, which interact in nonlinear electronics to produce an output signal. \textbf{D.} Just as hierarchical information processing is realized in DNNs by a sequence of trainable nonlinear functions, we can construct deep physical neural networks (PNNs) by cascading layers of trainable physical transformations. In these deep PNNs, each physical layer implements a controllable function, which can be of a more general form than that of a conventional DNN layer.} 
    \label{fig1}
\end{figure}

Theoretical proposals for physical learning hardware have recently emerged in various fields, such as optics \cite{hughes2019wave,Wu2020neuromorphicmetasurface,furuhata2021physical,nakajima2020neural}, spintronic nano-oscillators \cite{Romera2018spintronicPNNvowel}, nanoelectronic devices \cite{Chen2020geneticalgorithmSi,euler2020deep}, and small-scale quantum computers \cite{Mitarai2018quantumcircuitlearning,Benedetti2019parameterizedQCL,havlivcek2019supervised,fosel2021quantum}. A related trend is physical reservoir computing  \cite{Tanaka2019reservoircomputingreview,Nakajima2020reservoircomputingreview}, in which the information transformations of a physical system `reservoir' are not trained but are instead linearly combined by a trainable output layer. Reservoir computing harnesses generic physical processes for computation, but its training is inherently shallow: it does not allow the hierarchical process learning that characterizes modern deep neural networks. In contrast, the newer proposals for physical learning hardware \cite{hughes2019wave,Wu2020neuromorphicmetasurface,Mitarai2018quantumcircuitlearning,Benedetti2019parameterizedQCL,Chen2020geneticalgorithmSi,euler2020deep,Romera2018spintronicPNNvowel,stern2020supervised,furuhata2021physical,nakajima2020neural} overcome this by training the physical transformations themselves. 

There have been few experimental studies on physical learning hardware, however, and those that exist have relied on gradient-free learning algorithms \cite{Chen2020geneticalgorithmSi,Romera2018spintronicPNNvowel,bueno2018reinforcement}. While these works have made critical steps, it is now appreciated that gradient-based learning algorithms, such as the backpropagation algorithm, are essential for the efficient training and good generalization of large-scale DNNs \cite{poggio2020theoretical}. To solve this problem, proposals to realize backpropagation on physical hardware have appeared \cite{hughes2019wave,lin2018all,Wu2020neuromorphicmetasurface,euler2020deep,furuhata2021physical, hermans2015trainable,hughes2018training,lopez2021self}. While inspirational, these proposals nonetheless often rely on restrictive assumptions, such as linearity or dissipation-free evolution. The most general proposals \cite{hughes2019wave,Wu2020neuromorphicmetasurface,furuhata2021physical,euler2020deep} may overcome such constraints, but still rely on performing training \textit{in silico}, i.e., wholly within numerical simulations. Thus, to be realized experimentally, and in scalable hardware, they will face the same challenges as hardware based on mathematical analogies: intense engineering efforts to force hardware to precisely match idealized simulations.

Here we demonstrate a universal framework to directly train arbitrary, real physical systems to execute deep neural networks, using backpropagation. We call these trained hierarchical physical computations \textit{physical neural networks} (PNNs). Our approach is enabled by a hybrid physical-digital algorithm, \textit{physics-aware training} (PAT). PAT allows us to efficiently and accurately execute the backpropagation algorithm on any sequence of physical input-output transformations, directly \textit{in situ}. We demonstrate the universality of this approach by experimentally performing image classification using three distinct systems: the multimode mechanical oscillations of a driven metal plate, the analog dynamics of a nonlinear transistor oscillator, and ultrafast optical second harmonic generation. We obtain accurate hierarchical classifiers, which utilize each system’s unique physical transformations, and which inherently mitigate each system’s unique noise processes and imperfections. While PNNs are a radical departure from traditional hardware, they are easily integrated into modern machine learning. We show that PNNs can be seamlessly combined with conventional hardware and neural network methods via physical-digital hybrid architectures, in which conventional hardware learns to opportunistically cooperate with unconventional physical resources using PAT. Ultimately, PNNs provide a basis for hardware-physics-software codesign in artificial intelligence \cite{markovic2020physics}, routes to improving the energy efficiency and speed of machine learning by many orders of magnitude, and pathways to automatically designing complex functional devices, such as functional nanoparticles \cite{peurifoy2018nanophotonic}, robots \cite{veit2016residual}, and smart sensors \cite{Zhou2020smartsensor,Martel2020smartsensor,Mennel2020smartsensor,Duarte2008singlepixelimaging}.

\section{Results}

Figure 2 shows an example physical neural network based on the propagation of broadband optical pulses in quadratic nonlinear optical media, i.e. ultrafast second harmonic generation (SHG)\footnote{Technically speaking, propagation of intense broadband pulses in quadratic nonlinear optical media gives rise not only to second-harmonic generation, but also to sum-frequency and difference-frequency processes among the multitude of optical frequencies. For simplicity, we refer to these complex nonlinear optical dynamics as SHG.}. The ultrafast SHG process realizes a physical computation, a nonlinear transformation of the input near-infrared pulse's spectrum ($\sim$ 800 nm center-wavelength) to a new optical spectrum in the ultraviolet ($\sim$ 400 nm center-wavelength). To use and control this physical computation, input data and parameters are encoded into sections of the near-infrared pulse's spectrum by modulating its frequency components using a pulse shaper (Fig.~2A). This modulated pulse then propagates through a nonlinear optical crystal, producing a broadband ultraviolet pulse, whose spectrum is measured to read-out the computation result. 

A task routinely used to demonstrate novel machine learning hardware is the classification of spoken vowels according to their formant frequencies \cite{shen2017deep,Romera2018spintronicPNNvowel,voweldata}. In this task, the machine learning device must learn to predict spoken vowels from 12-dimensional input data vectors of formant frequencies extracted from audio recordings. 

To realize vowel classification with SHG, we construct a multi-layer SHG-PNN (Fig.~1B) where the input data for the first physical layer (i.e., the first physical transformation) consists of the vowel formant frequency vector. At the final layer, the SHG-PNN's predicted vowel is read out by taking the maximum value of 7 spectral bins in the measured ultraviolet spectrum (Fig.~1C). In each layer, the output spectrum is digitally renormalized before being passed to the next layer, along with a trainable digital rescaling (i.e., $\vec{x}_{i+1} = a\vec{y}_i/ \text{max}(\vec{y}_i) + b$). Thus, the classification is achieved almost entirely using the trained nonlinear optical transformations. The results of Fig.~2 show that the SHG-PNN adapts the physical transformation of ultrafast SHG into a hierarchical physical computation, which consists of five sequential, trained nonlinear transformations that cooperate to produce a final optical spectrum that accurately predicts vowels (Fig.~1B,D). 

\begin{figure}
    \centering
    \includegraphics[width=0.95\textwidth]{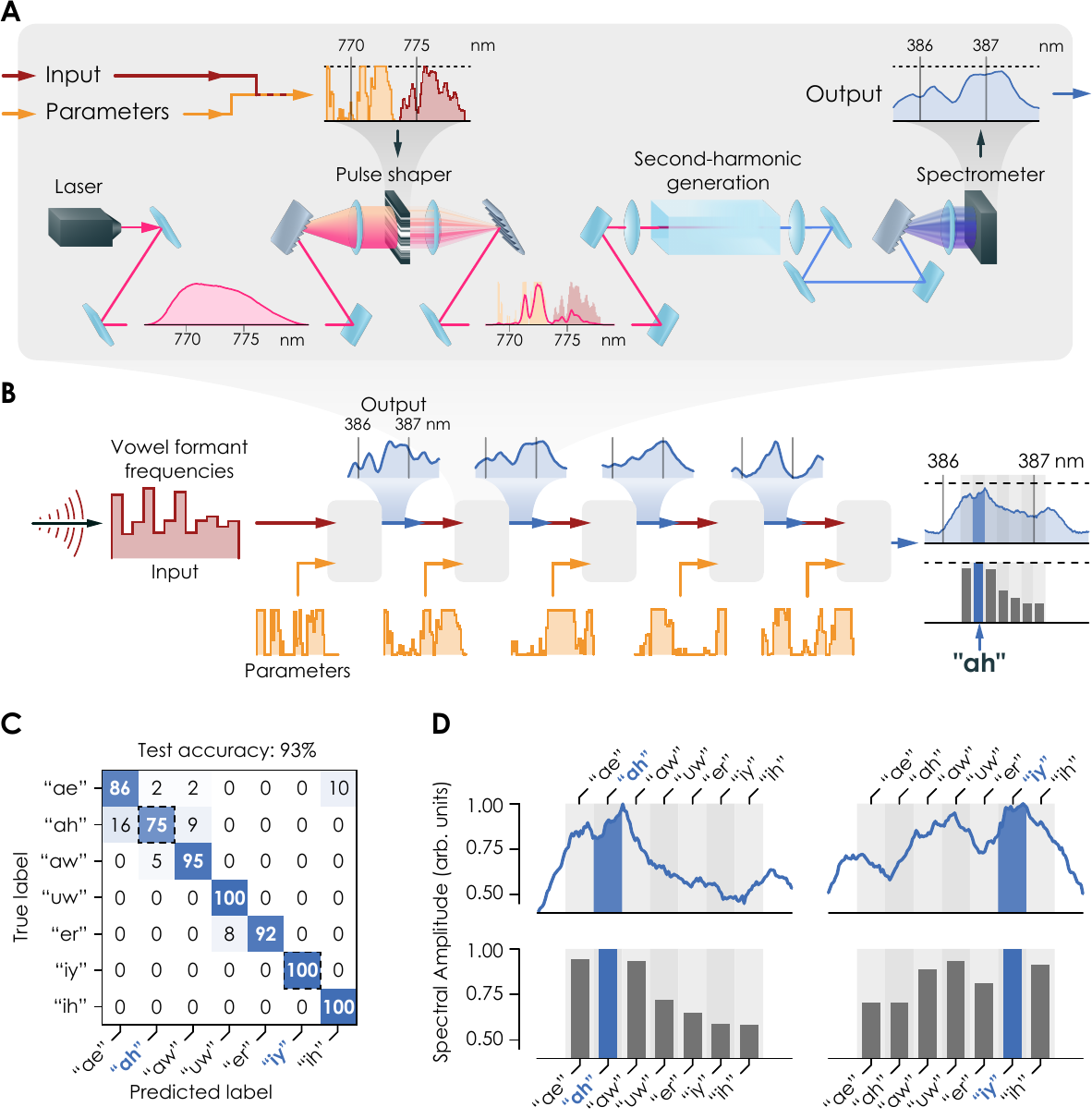}
    \caption{\textbf{An example physical neural network, implemented experimentally using broadband optical second-harmonic generation (SHG) with shaped femtosecond pulses. A.} Inputs to the system are encoded into the spectrum of a laser pulse using a pulse shaper, based on a diffraction grating and a digital micromirror device (see Supplementary Section 2). To control the physical transformations implemented by the broadband nonlinear interactions, we assign a portion of the pulse’s spectrum for trainable parameters (yellow). The result of the physical computation is obtained from the spectrum of the frequency-doubled (blue, $\sim$ 390 nm) output of the SHG process.  \textbf{B.} To construct a deep PNN with femtosecond SHG, we take the output of a SHG process and use it as the input to a second SHG process, which has its own independent trainable parameters. This cascading is repeated three more times, resulting in a multilayer PNN with five trainable physical layers ($\sim$ 250 physical parameters and 10 digital parameters). For the final layer, the spectrum is summed into 7 spectral bins, and the largest-sum bin gives the predicted vowel class. \textbf{C-D.} When this SHG-PNN is trained using PAT (see main text, Fig.~3), it is able to perform classification of vowels to 93\% accuracy. \textbf{C.} The confusion matrix for the PNN on a test data set, showing the labels predicted by the SHG-PNN versus the true (correct) labels. \textbf{D.} Representative examples of final-layer output for input vowels correctly classified as ``ah" and ``iy".  
    } 
    \label{fig2}
\end{figure}

To train PNNs' parameters, we use physics-aware training (PAT, Fig.~3), an algorithm that allows us to effectively perform the backpropagation algorithm for stochastic gradient descent (SGD) directly on any sequence of physical input-output transformations, such as the sequence of nonlinear optical transformations in Fig.~2. In the backpropagation algorithm, automatic differentiation efficiently determines the gradient of a loss function with respect to trainable parameters. This makes the algorithm $\sim N$-times more efficient than finite-difference methods for gradient estimation (where $N$ is the number of parameters). PAT has some similarities to quantization-aware training algorithms used to train neural networks for low-precision hardware \cite{hubara2017quantized}, and  feedback alignment \cite{lillicrap2016random}. PAT can be seen as solving a problem analogous to the `simulation-reality gap' in robotics \cite{jakobi1995noise}, which is increasingly addressed by hybrid physical-digital techniques \cite{howison2021reality}. 

\begin{figure}
    \centering
    \includegraphics[width=0.95\textwidth]{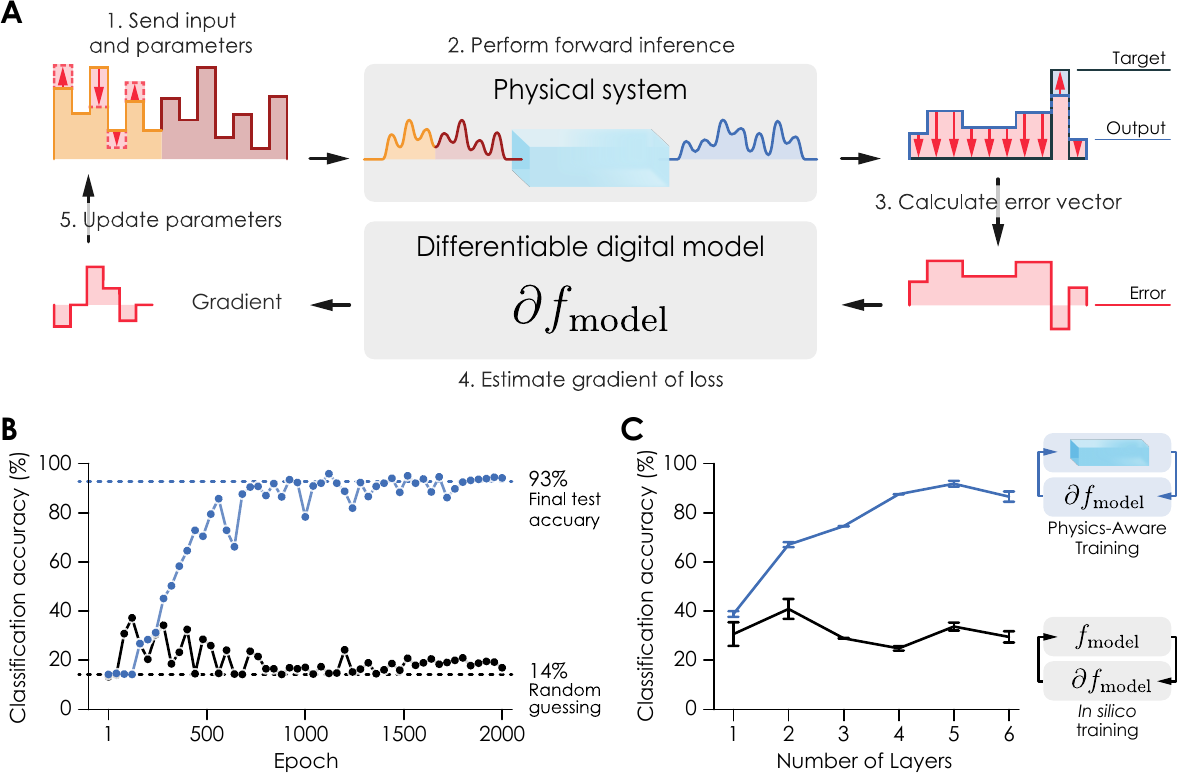}
    \caption{\textbf{Introduction to Physics-Aware Training. A.} Physics-aware training (PAT) is a hybrid physical-digital algorithm that implements backpropagation to train the controllable parameters of a sequence of dynamical evolutions of physical systems, \textit{in situ}. PAT's goal is to train the physical systems to perform machine-learning tasks accurately, even though digital models describing physics are always imperfect, and the systems may have noise and other imperfections. This motivates the hybrid nature of the algorithm, where rather than performing the training solely with the digital model (i.e., \textit{in silico}), the physical systems are directly used to compute forward passes, which keeps the training on track. For simplicity only one physical layer is depicted in \textbf{A} and we assume a mean-squared error loss function. As with conventional backpropagation, the algorithm generalizes straightforwardly to multiple layers and different loss functions, see Supplementary Section 1. First (1), training input data (e.g., an image) is input to the physical system, along with parameters. Second (2), in the forward pass, the physical system applies its transformation to produce an output. Third (3), the physical output is compared to the intended output (e.g., for an image of an `8', a predicted label of 8) to compute the error. Fourth (4), using a differentiable digital model to estimate the gradients of the physical system(s), the gradient of the loss is computed with respect to the controllable parameters. Finally, (5) the parameters are updated according to the inferred gradient. This process is repeated, iterating over training examples, to reduce the error. \textbf{B.} Comparison of the validation accuracy versus training epoch with PAT and \textit{in silico} training (i.e., training in which the digital model is used for both forward propagation and gradient estimation), for the experimental SHG-PNN depicted in Fig.~2C (5 physical layers). \textbf{C.} Final experimental test accuracy for PAT and \textit{in silico} training for SHG-PNNs with increasing numbers of physical layers. Due to the compounding of simulation-reality mismatch error through training, \textit{in silico} training results in accuracy barely distinguishable from random guessing, while PAT permits training the physical device to realize 93\% accuracy. As physical layers are added, \textit{in silico} training fails, but PAT is able to produce accurate physical neural networks. Due to the relative simplicity of the task, and the design of the SHG-PNN (which was chosen to be intuitive and involve unconventional physics, at the expense of information-processing capacity), the PNN's performance improvement saturates for depth greater than 5 layers.}
    \label{fig3}
\end{figure}

PAT’s essential advantages stems from the forward pass being executed by the actual physical hardware, rather than by a simulation. Given an accurate model, one could attempt to train by autodifferentiating simulations, then transferring the final parameters to the physical hardware (termed \textit{in silico} training in Fig.~3). Our data-driven digital model for SHG is remarkably accurate (Supplementary Figure 20) and even includes an accurate noise model (Supplementary Figures 18-19). However, as evidenced by Fig.~3B, \textit{in silico}  training still fails, reaching a maximum vowel-classification accuracy of $\sim$ 40\%. In contrast, PAT succeeds, accurately training the SHG-PNN, even when additional layers are added (Fig.~3B-C). 

An intuitive motivation for why PAT works is that the training's optimization of parameters is always grounded in the true optimization landscape by the physical forward pass. With PAT, even if gradients are estimated only approximately, the true loss function is always precisely known. Moreover, the true output from each intermediate layer is also known, so gradients of intermediate physical layers are always computed with respect to correct inputs. In contrast, in any form of \textit{in silico} training, compounding errors build up through the simulation of each physical layer, leading to a rapidly diverging simulation-reality gap as training proceeds (see Supplementary Section 1 for details). As a secondary benefit, PAT ensures learned models are inherently resilient to noise and other imperfections beyond a digital model, since the change of loss along noisy  directions in parameter space will tend to average to zero. This facilitates the learning of noise-resilient (more speculatively, noise-enhanced) models that automatically adapt to device-device variations and brain-like stochastic dynamics \cite{markovic2020physics}.

PNNs can learn to accurately perform more complex tasks, and can be realized with virtually any physical system. In Figure 4, we present three PNN classifiers for the MNIST handwritten digit classification task, based on three distinct physical systems. For each physical system, we also demonstrate a different PNN architecture, illustrating the wide variety of PNN models possible. In all cases, models were constructed and trained in PyTorch \cite{paszke2017automatic}, where each trainable physical transformation called an automated experimental apparatus with a vector of the input data, $\vec{x}$, and parameters $\vec{\theta}$, \codeword{y=run_exp(x,theta)}, which is made differentiable by using the digital model for backward operations (see Supplementary Section 1).  

In the mechanical PNN (Fig.~4A-D), a metal plate is driven by a time-varying force, which encodes both input data and trainable parameters. We find that the plate’s multimode oscillation enacts a controllable convolution on input data (Supplementary Figures 16-17).  Using the oscillating plate’s trainable transformation three times in sequence, we classify 28 by 28 (784-pixel) images input as unrolled, 784-dimensional time series. To control the physical transformations, we train element-wise rescaling of the time-series of forces sent to the plate digitally (i.e., $x_i = y_ia_i+b_i$, where $b_i$ and $a_i$ are trainable and $y_i$ is the input image or an output from a previous layer).  The PNN achieves 87\% test accuracy, close to the optimal performance of a linear classifier \cite{lecun1998gradient}. To ensure that the digital re-scaling operations are not responsible for a significant part of the physical computation, we repeat the experiment with the physical transformation replaced by an identity operation. When this model is trained, its optimal performance is comparable to random guessing (10\%). This shows that most of the PNN's functionality comes from the controlled physical transformations. 

An analog-electronic PNN is implemented with a circuit featuring a transistor (Fig.~4E-H), which results in a noisy, highly-nonlinear transient response (Supplementary Figures 12-13) that is more challenging to accurately model than the oscillating plate's response (Supplementary Figure 23). Inputs and parameters are realized through the time-varying voltage applied across the circuit and the output is taken as the voltage time-series measured across the inductor and resistor (Fig.~4F). The electronic PNN architecture is similar to the mechanical PNN’s, except that the final prediction is obtained by averaging the predictions of 7 independent, 3-layer PNNs. Since it is nonlinear, the electronic PNN outperforms the mechanical PNN, achieving 93\% test accuracy. 

Finally, using broadband SHG, we demonstrate a physical-digital hybrid PNN (Fig.~4I-L).  The hybrid architecture performs an inference as follows: First, the MNIST image is acted on by a trainable linear input layer, which produces the input to the SHG transformation. The full PNN architecture involves 4 separate channels, each with 2 physical layers, whose outputs are concatenated to produce the PNN’s prediction. The inclusion of the trainable SHG transformations boosts the performance of the digital operations from roughly 90\% accuracy to 97\%. Since the classification task’s difficulty is nonlinear with respect to accuracy, such an improvement typically requires increasing the digital operations count by at least one order of magnitude \cite{lecun1998gradient}. This illustrates how a hybrid physical-digital PNN can automatically learn to offload portions of a computation from an expensive digital processor to a fast, energy-efficient physical co-processor.

\begin{figure}
    \centering
    \includegraphics[width=0.95\textwidth]{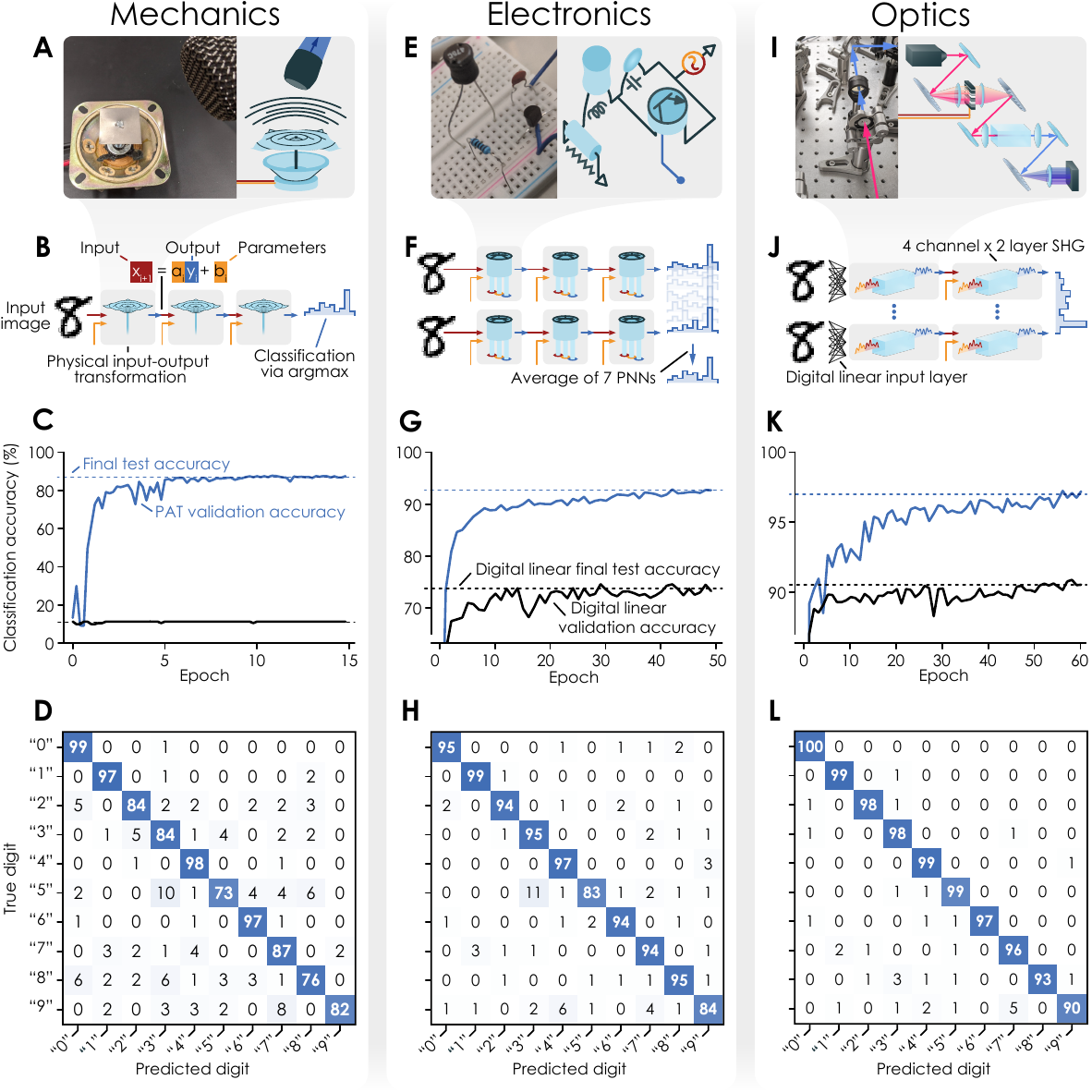}
    \caption{\textbf{Image classification with diverse physical systems.} We trained PNNs based on three distinct physical systems (mechanical, electronic, and optical) to classify images of handwritten digits. \textbf{A.} A photo and sketch of the mechanical PNN, wherein the multimode oscillations of a metal plate are driven by time-dependent forces that encode the input image data and parameters. \textbf{B.} A depiction of the mechanical PNN multi-layer architecture. To efficiently encode parameters and input data, we make use of digital element-wise rescaling of each input by trainable scaling factors and offsets. \textbf{C.} The validation classification accuracy versus training epoch for the mechanical PNN trained using PAT. Because some digital operations are used in the PNN, we also plot the same curve for a reference model where the physical transformations implemented by the speaker are replaced by identity operations. The PNN with physical transformations reaches nearly 90$\%$ accuracy, whereas the digital-only baseline experiment does not exceed random-guessing (10$\%$). \textbf{D.} Confusion matrix showing the classified digit label predicted by the mechanical PNN versus the correct result. \textbf{E-H.} The same as A-D, but for a nonlinear electrical PNN based on a transistor embedded in an RLC oscillator. \textbf{I-L.} The same as A-D, for a hybrid physical-digital PNN combining digital linear input layers (trainable matrix-vector multiplications) followed by trainable physical transformations using broadband optical second-harmonic generation. Final test accuracy is 87\%, 93\% and 97\% for mechanical, electrical, and optics-based PNNs respectively.}
    \label{fig4}
\end{figure}

\section{Discussion and Conclusion}

While the above results demonstrate that a variety of physical systems can be trained to perform machine learning, these proof-of-concept experiments leave open practical questions such as: What physical systems are good candidates for physical neural networks (PNNs), and how much can they speed up or reduce the energy consumption of machine learning? One naive approach is to analyze the costs of digitally simulating programmable physical systems relative to evolving them physically, as is done in recent work demonstrating quantum computational advantage \cite{arute2019quantum,zhong2020quantum}. Of course, such \textit{self-simulation advantages} will overestimate PNN advantages for practical tasks, but simulation analysis can still be insightful: it allows us to upper-bound and evaluate how potential advantages scale with physical parameters, to appreciate the mathematical operations that each physical system effectively approximates, and to therefore identify practical tasks each system may be best-suited for. We find that typical self-simulation advantages grow as the connectivity of the PNN's physical degrees of freedom increases, as the dimensionality of the physics increases, and as the nonlinearity of interactions increases (Supplementary Section 3). Realistic device implementations routinely achieve self-simulation advantages over $10^6$ ($10^9$) for speed (energy), with much larger values possible by size-scaling. Simulation analysis reveals that candidate PNN systems approximate standard machine learning operations like controllable matrix-vector multiplications, as in multimode wave propagation \cite{saade2016random} or coupled spintronic oscillators \cite{Romera2018spintronicPNNvowel} or electrical networks, as well as less-common operations, like the nonlinear convolutions in ultrafast optical SHG and the higher-order tensor contractions implemented by (multimode) nonlinear wave propagation in Kerr media \cite{teugin2020scalable}. 

Reaching the performance ceiling implied by self-simulation for tasks of practical interest should be possible, but will require a co-design of physics and algorithms \cite{markovic2020physics}.  On one hand, many candidate PNN systems, such as multimode optical waves, and networks of coupled transistors, lasers or nano-oscillators, exhibit dynamical evolutions that closely resemble standard neural networks and neural ordinary differential equations \cite{chen2018neural} (see Supplementary Section 3). For these systems, training PNNs with physics-aware training (PAT) will facilitate the learning of hierarchical physical computations of similar form as those in conventional deep neural networks, but overcoming the imperfect simulations, device variations, and higher-order physical processes that would pose an impenetrable challenge for approaches based on \textit{in silico} training or approximate mathematical analogies. Since their controlled dynamics are well-known to resemble conventional neural networks, it is likely that PNNs based on these systems will be able to translate self-simulation advantages of $10^6$ and higher into comparable speed-ups and energy efficiency gains on problems of practical interest. As our SHG-based machine learning shows, however, PNNs do not need not be limited to physical operations that resemble today's popular machine learning algorithms. Since they have not yet been widely studied in machine learning models, it is difficult to estimate how more exotic physical operations like nonlinear-optics-based high-order tensor contractions will translate to practical computations. Nonetheless, if we can learn how to harness them productively, PNNs incorporating these novel operations may realize even more significant computational advantages. In addition, there are clear technical strategies to design PNNs to maximize their practical benefits.  Even though the inference stage may account for as much as 90$\%$ of energy consumption due to DNNs in commercial deployments \cite{jassy2019aws,patterson2021carbon}, it will also be valuable to improve PAT to facilitate physics-enhanced or physics-based learning \cite{hughes2018training,lopez2021self}. Meanwhile, general design techniques like multidimensional layouts, in-place stored parameters, and physical data feedforward will permit scaling PNN complexity with minimal added energy cost (for a full exposition, see Supplementary Section 4). 

This work has focused so far on PNNs specifically as accelerators for machine learning, but PNNs are promising for other applications as well. PNNs can perform computations on data within its physical domain, allowing for smart sensors \cite{Zhou2020smartsensor,Martel2020smartsensor,Mennel2020smartsensor,Duarte2008singlepixelimaging} that pre-process information prior to conversion to the electronic domain (e.g., a low-power, microphone-coupled circuit tuned to recognize specific hotwords). Since the achievable sensitivity and resolution of many sensors is limited by digital electronics, PNN-sensors should have advantages over conventional approaches. More broadly, with PAT, one is simply training the complex functionality of physical systems. While machine learning and sensing are important functionalities, they are but two of many \cite{peurifoy2018nanophotonic,howison2021reality} that PAT, and the concept of PNNs, could be applied to. 

\section*{Acknowledgments}
L.G.W., T.O. and P.L.M. conceived the project and methods. T.O. and L.G.W. performed the SHG-PNN experiments. L.G.W. performed the electronic-PNN experiments. M.M.S. performed the oscillating-plate-PNN experiments. T.W., D.T.S., and Z.H.  contributed to initial parts of the work. L.G.W., T.O., M.M.S. and P.L.M. wrote the manuscript. P.L.M. supervised the project. L.G.W. and T.O. contributed equally to the work overall. The authors wish to thank NTT Research for their financial and technical support. Portions of this work were supported by the National Science Foundation (award CCF-1918549). L.G.W. and T.W. acknowledge support from Mong Fellowships from Cornell Neurotech during early parts of this work. P.L.M. acknowledges membership of the CIFAR Quantum Information Science Program as an Azrieli Global Scholar. We acknowledge helpful discussions with Danyal Ahsanullah, Vladimir Kremenetski, Edwin Ng, Sebestian Popoff, Sridhar Prabhu, Hidenori Tanaka and Ryotatsu Yanagimoto, and members of the NTT PHI Lab / NSF Expeditions research collaboration, and thank Philipp Jordan for discussions and illustrations. \textit{Competing interests:} L.G.W., T.O., M.M.S. and P.L.M. are listed as inventors on a U.S. provisional patent application (No. 63/178,318) on physical neural networks and physics-aware training.

\section*{Data and code availability}
An expandable demonstration code for applying physics-aware training to train physical neural networks is available at: \url{https://github.com/mcmahon-lab/Physics-Aware-Training}. All data generated and code used for this work is available at \url{https://doi.org/10.5281/zenodo.4719150}.

\bibliographystyle{npjqi}
\bibliography{references}
\end{document}


\title{Deep physical neural networks enabled by a backpropagation algorithm for arbitrary physical systems: supplementary material}
\maketitle

\tableofcontents
\section{Physics-aware training}
\newcommand{\ph}{\text{p}}
\newcommand{\mo}{\text{m}}
\newcommand{\pat}{\text{PAT}}
\newcommand{\pd}[2]{\frac{\partial #1}{\partial #2}}

In this section, we describe physics-aware training (PAT) and explain its effectiveness at training physical neural networks (PNNs). In section \ref{sec:pat-general}, we develop the algorithm in full mathematical detail, comparing it to the conventional backpropagation algorithm. In Sec.~\ref{sec:pat-mlpnn}, this general case is specified to a generic multi-layer physical neural network. In that section, we introduce and compare physics-aware training to the most relevant alternative for training sequences of physical systems with backpropagation, \emph{in silico} training, in which the backpropagation algorithm is applied to a simulation of the physical system(s). Finally, in Sec.~\ref{sec:pat-num}, we present a numerical example to demonstrate how PAT efficiently trains PNNs, by training a simulated feedforward PNN to perform the vowel classification task with the different training algorithms introduced in this section.  

Before delving into more detailed subsections, however, the next subsection provides a high-level intuition for why physics-aware training works, and why alternatives fail. This intuition is backed up rigorously in subsequent subsections.

\subsection{Intuition for why physics-aware training works}
The backpropagation algorithm solves the problem of efficiently training the functionality of sequences of mathematical operations with many trainable parameters to perform desired mathematical functions. Its key ingredient is the analytic differentiation of the mathematical operations to efficiently compute the gradient, and hence to determine the optimal parameter updates to improve performance. 

Physics-aware training (PAT) solves the problem of applying the backpropagation algorithms to train sequences of real physical operations to perform desired physical functions – which includes performing physically-implemented computations, i.e. physical neural networks. 

There are two natural alternatives to PAT. The backpropagation algorithm cannot be applied directly – real physical systems cannot be analytically differentiated. Instead, one could estimate derivatives by using a finite-difference estimator. But this requires $n$ uses of the physical system, where $n$ is the number of trainable parameters. Since $n$ is often $10^6$ or even $10^9$, and keeps increasing as deep neural network models are scaled up, finite-difference just isn’t a practical solution, even for very fast physical systems. The other alternative is to train the physical system using numerical simulations, which are ultimately made up of mathematical operations that one can apply the backpropagation algorithm to. The problem with this approach, however, is that a simulation is not reality – even the most finely-tuned simulation will only describe reality to within some finite precision. The existence of this inevitable simulation-reality gap leads to two problems. 

The first problem is the build-up of the simulation-reality gap through layers. Even if the simulation-reality gap is very small (i.e., the model is very accurate), it can matter when we have a process that is composed of a sequence of physical layers, where each layer after the first one takes in the output of the previous layer as an input. Basic error propagation shows us that a small simulation-reality gap (i.e., a small difference between reality and the simulation of reality) is going to build up, usually very rapidly, through layers. 

This is a very simple problem, so simple that it can be easily misunderstood. Consider a toy example. Suppose we have a function $f(x) = 2x^{1.1}$ (the `reality’) and a very good `model’ for that function $f_m(x) = 2.01x^{1.1}$. If we take a composition of $f(x)$, $f(f(f(f(f(…f(x)))))) = f^n(x)$, as occurs in a deep network or general hierarchical process, the small inaccuracies of the model blow-up. For $n=1$, the simulation-reality gap (the relative error in the model’s output compared to the reality) is about 0.5\%. For $n=5$, it has grown to 3\%. By $n=20$, the gap is 30\%. Of course, this is only a toy example: the blow-up depends, among other things, on the function, where it is evaluated, and how and where the model is inaccurate. The key point is that the model is increasingly wrong as the depth increases.

The second problem is even bigger – the build-up of the simulation-reality gap through training. Training is a sequential process that involves $10^5$, or even many more steps, each of which depends on the previous one. Each step of training results in new parameters, $\theta_{i}=T(\theta_{i-1}, f)$, where $T$ is the function(al) that performs one training step and $\theta_{i}$ is the parameter vector on the $i$th training step. The final parameters at training step $n$ are thus a composition of many applications of $T$, $\theta_{n}=T^n(\theta_{0}, f)$. Just as the simulation-reality gap grows with the number of layers, so too does it grow with training steps. Since there are usually many more training steps than layers, and because each training step is a function of the full-depth network (so it is fully affected by the first problem), this second problem is the most significant, and means that even extremely accurate models that can predict deep networks well, are still not good enough to facilitate training of real physical systems. 

Physics-aware training mitigates these problems. It solves the first problem, the depth problem, by simply using the true $f(x)$. The second problem, the training-step problem, is mitigated by PAT both because the simulation-reality gap for $f(x)$ is zero, and because the parameter updates are estimated more accurately than if backpropagation were just applied directly to the model. This second point is more subtle and will be elaborated on in rigorous detail in subsequent sections. An approximate intuition, however, is that PAT assures that when we adjust parameters to change the output of a sequence of physical operations, we are accurately estimating which \textit{direction} the output needs to change to improve its true performance. In contrast, when just performing backpropagation on a model, we may inadvertently think that the network is overshooting the desired output, when in fact, if the simulation’s parameters were applied to the true physical system, the output would be undershooting. A second aspect of this improved parameter update estimation is that, even though PAT uses a model to estimate gradients, it uses that model at the correct locations for each layer. For example, in the toy example above, if one is performing backpropagation on the model, the location at which derivatives of the model are evaluated is off by only 0.5\% for the second layer (which receives the first layer’s output as an input), but is off by 30\% for the 21st layer. With PAT, the gradients are always estimated assuming the correct inputs. The net effect is that, with PAT, the simulation-reality gap grows much more slowly than if training were performed solely with simulations. Since PAT is always grounded in the true loss, training can never become misled about how well the physical computation is performing. As a result, even if a small simulation-reality gap may still develop with PAT, it still generally finds parameters that lead to good performance, even if those parameters are sometimes different from those that would have been obtained if ideal backpropagation could be applied to the physical system. 

In sum, PAT corrects backpropagation of physical systems, compared to performing backpropagation exclusively with a model, by preventing the build-up of a simulation-reality gap in several different ways. First, it exactly calculates the output of a sequence of physical operations, as well as each intermediate output in the sequence, because it uses the actual physical system to do this. Second, it improves the estimation of parameter updates based on gradient descent, because the deviation of the true physical output from the intended output is exactly known, and because gradients are always estimated with respect to the correct inputs. 

PAT seems to do many other things that are helpful, that are mostly beyond the scope of this work, but that are worth mentioning briefly here. These other helpful things include: making sure the trained model works despite the existence of noise in the forward pass, allowing generic models to be used even when there are device-device variations, and allowing training to proceed accurately even when the model fails to account for parts of the true physics. The first occurs because PAT includes a perfect model of the physical system's noise, inherently (the physical system itself). During training, if the physical system's output randomly fluctuates, estimated parameter updates will also fluctuate. Training's progress along `noisy' dimensions (i.e., dimensions for which the loss strongly fluctuates due to noise) in parameter space will therefore be slowed depending on how noisy those dimensions are. This is essentially a generalization of how quantization-aware training \cite{hubara2017quantized} facilitates the training of deep neural networks that can tolerate low-precision weights and low-precision operations. PAT allows generic models to be used on heterogeneous devices because it does not require a perfect model for the physical system (though it does require access to the true physical system). This may be helpful when training mass-manufactured devices whose individual parameters may vary. Lastly, when the model is not accurate in particular regimes, PAT seems to see this as noise, since the estimated parameter updates will average to approximately zero along `mis-modelled' directions, due to the fluctuating disagreement between the physics and model. This is probably a generalization of the phenomenon of feedback alignment encountered when training deep neural networks with random synaptic feedback \cite{lillicrap2016random}.

\subsection{General formulation of physics-aware training}
\label{sec:pat-general}

Physics-aware training (PAT) is a gradient-based learning algorithm. 
The algorithm computes the gradients of the loss \wrt the parameters of the network. Since the loss indicates how well the network is performing at its machine learning task, the gradients of the loss are subsequently used to update the parameters of the network. The gradients are computed efficiently via the backpropagation algorithm, which we briefly review below. 

The backpropagation algorithm is commonly applied to neural networks composed of differentiable functions. It involves two key steps: a forward pass to compute the loss and a backward pass to compute the gradients with respect to the loss. The mathematical technique underpinning this algorithm is reverse-mode automatic differentiation (autodiff).
Here, each differentiable function in the network is an autodiff function which specifies how signals propagate forward through the network and how error signals propagate backward. Given the constituent autodiff functions and a prescription for how these different functions are connected to each other, i.e., the network architecture, reverse-mode autodiff is able to compute the desired gradients iteratively from the outputs towards the inputs and parameters (heuristically termed ``backward") in an efficient manner. 

For example, the output of a conventional deep neural network is given by $f(f( \ldots f( f(\bm x,  \bm \theta^{[1]}), \bm \theta^{[2]}) \ldots  , \bm \theta^{[N-1]}), \bm \theta^{[N]})$. $f$ denotes the constituent autodiff function and is given by $f(\bm x, \bm\theta) = \text{Relu}(W \bm x + \bm{b})$, where the weight matrix $\bm W$ and the bias $\bm b$ are the parameters of a given layer, and Relu is the rectifying linear unit activation function (other activation functions may be used). Given a prescription for how the forward and backward pass is performed for $f$, the autograd algorithm is able to compute the overall loss of the network. In Sec.~\ref{sec:pat-mlpnn}, the full training procedure for simple feedforward NN of this kind is described in more detail. 

\begin{figure}
    \centering
    \includegraphics[width=0.6\linewidth]{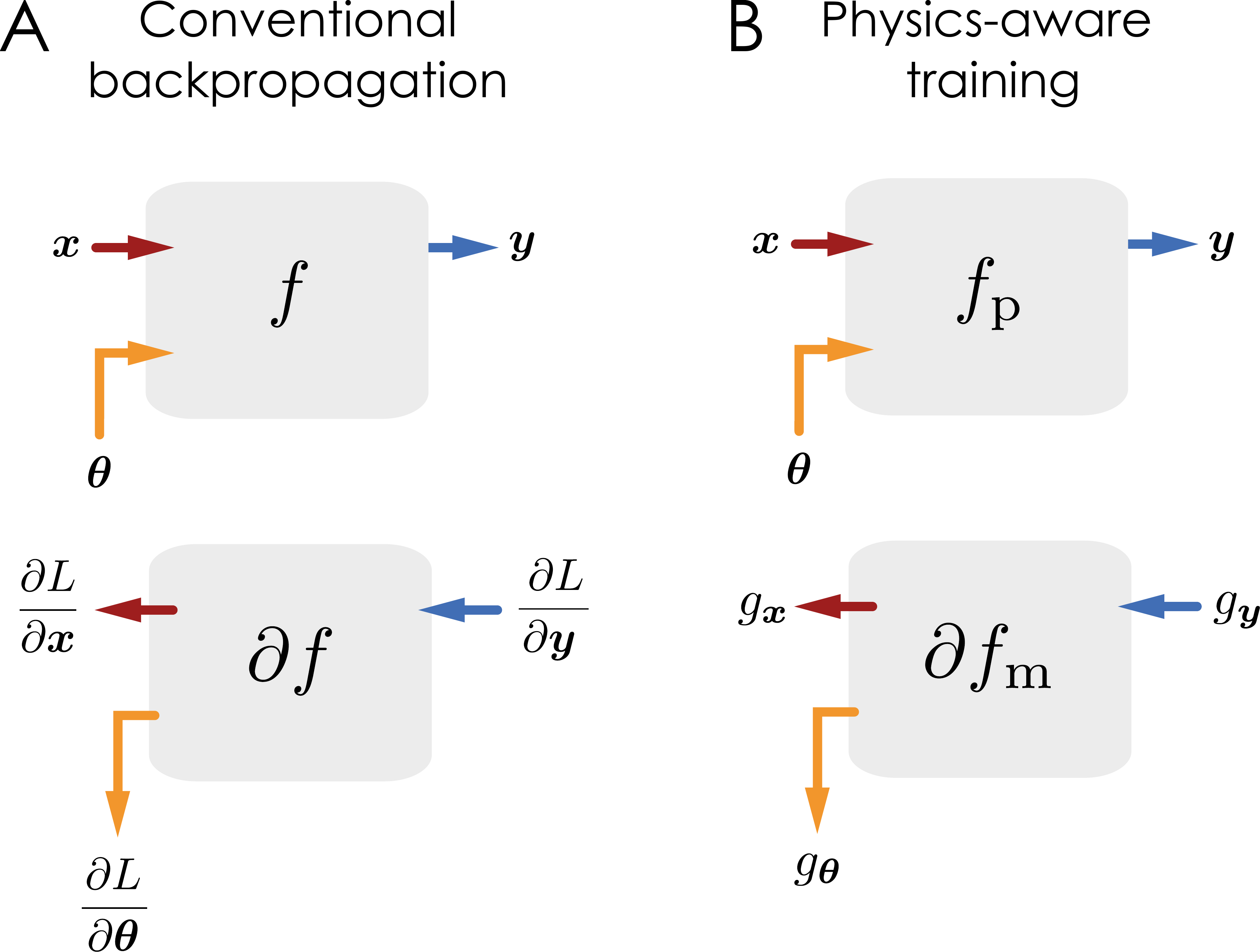}
    \caption{Schematic for the autodiff function in conventional backpropagation (A) and in physics-aware training (B). In the forward pass, input and parameters are propagated to the output. In the backward pass, gradients \wrt the output are backpropagated to gradients \wrt the input and parameters. In physics-aware training, the backward pass is estimated using a differentiable digital model, $f_\mo(\bm{x}, \bm{\theta})$, while the forward pass is implemented by the physical system,  $f_\ph(\bm{x}, \bm{\theta}$).}
    \label{fig:general-schematic}
\end{figure}

The conventional backpropagation algorithm uses the same function for the forward and backward passes (see Fig.~\ref{fig:general-schematic}). Mathematically, the forward pass, which maps the input and parameters into the output, is given by
\begin{align}
    \bm{y} = f(\bm{x}, \bm{\theta}),
    \label{eq:pat-1}
\end{align}
where $\bm{x} \in \mathbb{R}^n$ is the input, $\bm{\theta}  \in \mathbb{R}^p$ are the parameters, $\bm{y} \in \mathbb{R}^m$ is the output of the map, and $f : \mathbb{R}^n \times \mathbb{R}^p \rightarrow \mathbb{R}^m$ represents some general function that is an constituent operation which is applied in the overall neural network. 


The backward pass, which maps the gradients \wrt the output into gradients \wrt the input and parameters, is given by the following Jacobian-vector product:
\begin{align}
    \pd{L}{\bm{x}} & = \paren{\pd{f}{\bm{x}}(\bm{x}, \bm{\theta})}^\T \pd{L}{\bm{y}},
\\  \pd{L}{\bm{\theta}}  & = \paren{\pd{f}{\bm{\theta}}(\bm{x}, \bm{\theta})}^\T \pd{L}{\bm{y}},    \label{eq:pat-2}
\end{align}
where $\pd{L}{\bm{y}} \in \mathbb{R}^m, \pd{L}{\bm{x}}  \in \mathbb{R}^n, \pd{L}{\bm{\theta}}  \in \mathbb{R}^p$ are the gradients of the loss \wrt the output, input and parameters respectively. $\pd{f}{\bm{x}}(\bm{x}, \bm{\theta}) \in \mathbb{R}^{m \times n}$ denotes the Jacobian matrix of the function $f$  \wrt $\bm{x}$ evaluated at $(\bm{x}, \bm{\theta})$, i.e. $\paren{\pd{f}{\bm{x}}(\bm{x}, \bm{\theta})}_{ij} = \pd{f_i}{x_j}(\bm{x}, \bm{\theta})$. Analogously, $\paren{\pd{f}{\bm{\theta}}(\bm{x}, \bm{\theta})}_{ij} = \pd{f_i}{\theta_j}(\bm{x}, \bm{\theta})$. 

Though the backpropagation algorithm given by (\ref{eq:pat-1}-\ref{eq:pat-2}) works well for deep learning with digital computers - it cannot be naively applied to physical neural networks. This is because the analytic gradients of input-output transformations performed by physical systems (such as $\pd{f}{\bm{x}}(\bm{x}, \bm{\theta})$) cannot be computed. The derivatives could be estimated via a finite-difference approach -  requiring $n+p$ number of calls to the physical system per backward pass. The resulting algorithm would be $n+p$ times slower - this is significant for reasonably sized networks with thousands or millions of trainable parameters.

In physics-aware training, we propose an alternative implementation of the backpropagation algorithm that is suitable for arbitrary real physical input-output transformations. This variant employs autodiff functions that utilizes different functions for the forward and backward pass. As shown schematically in Fig.~\ref{fig:general-schematic}B, we use the non-differentiable transformation of the physical system $(f_\ph)$ in the forward pass and use the differentiable digital model of the physical transformation $(f_\mo)$ in the backward pass. Since physics-aware training is a backpropagation algorithm, which efficiently computes gradients in an iterative manner, it is able to train the physical system efficiently. Moreover, as it is formulated in the same paradigm of reverse-mode automatic differentiation, a user can define these custom autodiff functions in any conventional machine learning library (such as PyTorch) to design and train physical neural networks with the same workflow adopted for conventional neural networks. 

Formally, physics-aware training is described as follows. For a constituent physical transformation in the overall physical neural network, the forward pass operation of this constituent is given by 
\begin{align}
    \bm{y} = f_\ph(\bm{x}, \bm{\theta}) .
\end{align}

As a different function is used in the backward pass than in the forward pass, the autodiff function is no longer able to backpropagate the gradients at the output layer to the exact gradients at the input layer. Instead it strives to approximate the backpropagation of the exact gradients. Thus, the backward pass is given by
\begin{align}
    g_{\bm{x}} & = \paren{\pd{f_\mo}{\bm{x}}(\bm{x}, \bm{\theta})}^\T g_{\bm{y}} , 
\\  g_{\bm{\theta}} & = \paren{\pd{f_\mo}{\bm{\theta}}(\bm{x}, \bm{\theta})}^\T g_{\bm{y}} , 
\end{align}
where $g_{\bm{y}}$, $g_{\bm{x}}$, and $g_{\bm{\theta}}$ are estimators of the gradients $\pd{L}{\bm{y}}$, $\pd{L}{\bm{x}}$, and $\pd{L}{\bm{\theta}}$ respectively.

In listing 1, we show how to construct these custom autograd function for physics-aware training in PyTorch, a popular library for machine learning. With these functions at hand, the user can proceed to build physical neural network with complex architectures in PyTorch. (An implementation of physics-aware training in PyTorch is available at: \href{https://github.com/mcmahon-lab/Physics-Aware-Training}{https://github.com/mcmahon-lab/Physics-Aware-Training}.) 

\clearpage
\begin{lstlisting}[language=Python,
    label={lst:pat},
    caption=A helper function that generates the custom autograd function for physics-aware training. This code was designed to run on Python 3.7 and PyTorch 1.6 . ,
    captionpos=b,
    basicstyle=\small\sffamily,
    numbers=left,
    numberstyle=\tiny,
    frame=tb,
    columns=fullflexible,
    showstringspaces=false,
    deletekeywords={apply}]
    import torch

    def make_pat_func(forward_f, backward_f):
        """
        A function that constructs and returns the custom autograd function for physics-aware training.

        Parameters 
        ----------
        f_forward: The function that is applied in the forward pass. 
                Typically, the computation is performed on a physical system that is connected and controlled by 
                the computer that performs the training loop.
                It takes in an input PyTorch tensor x, a parameter PyTorch tensor theta and returns the output PyTorch tensor y. 
        f_backward: The function that is employed in the backward pass to propagate estimators of gradients.
                  It takes in an input PyTorch tensor x, an input PyTorch theta and returns the output PyTorch tensor y. 

        Returns
        -------
        f_pat: The custom autograd function for physics-aware training. 
              It takes in an input PyTorch tensor x, an input PyTorch theta and returns the output PyTorch tensor y. 

        """
        class func(torch.autograd.Function):
            @staticmethod
            def forward(ctx, x, theta): #here ctx is an object that stores data for the backward computation.
                ctx.save_for_backward(x, theta) #ctx is used to save the input and parameter of this function.
                return f_forward(x, theta)
            
            @staticmethod
            def backward(ctx, grad_output):
                x, theta = ctx.saved_tensors #load the input and parameters that are stored in the forward pass
                torch.set_grad_enabled(True) #autograd has to be enabled to perform jacobian computation.
                # Perform vector jacobian product of the backward function with PyTorch. 
                y = torch.autograd.functional.vjp(f_backward, (x, theta), v=grad_output)
                torch.set_grad_enabled(False) #autograd should be restored to off state after jacobian computation. 
                return y[1]
        f_pat = func.apply
        return f_pat
\end{lstlisting}

\newpage

\begin{figure}[h!]
    \centering
    \includegraphics[width=0.8\linewidth]{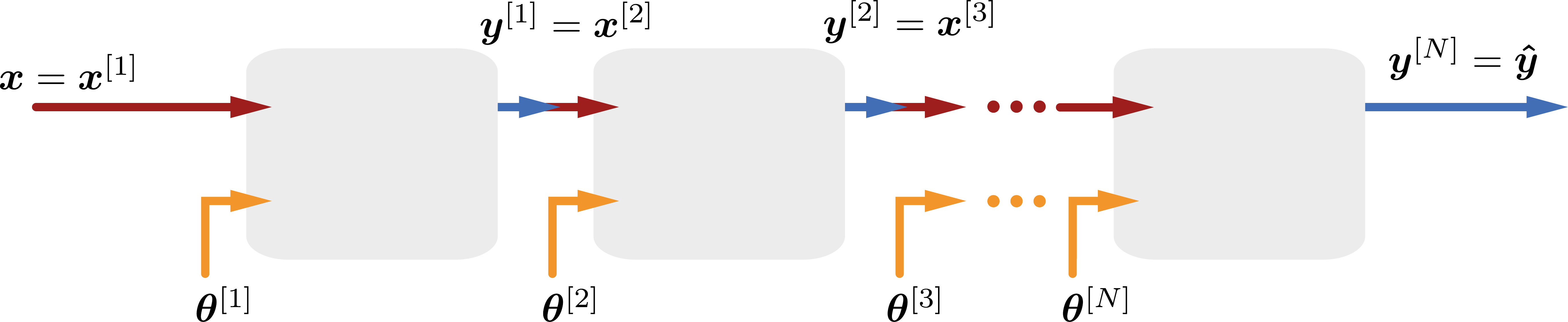}
    \caption{Schematic of feedforward physical neural network (PNN). In this architecture, the output of a physical system is passed as an input to the subsequent physical system, i.e. $\bm{x}^{[l+1]} = \bm{y}^{[l]}$. A parameter $\bm \theta^{[l]}$ is fed into each layer of the PNN to train the transformation that the PNN applies to the input $\bm x^{[l]}$.}
    \label{fig:ff_pnn}
\end{figure}

\begin{figure}[h!]
    \centering
    \includegraphics[width=0.8\linewidth]{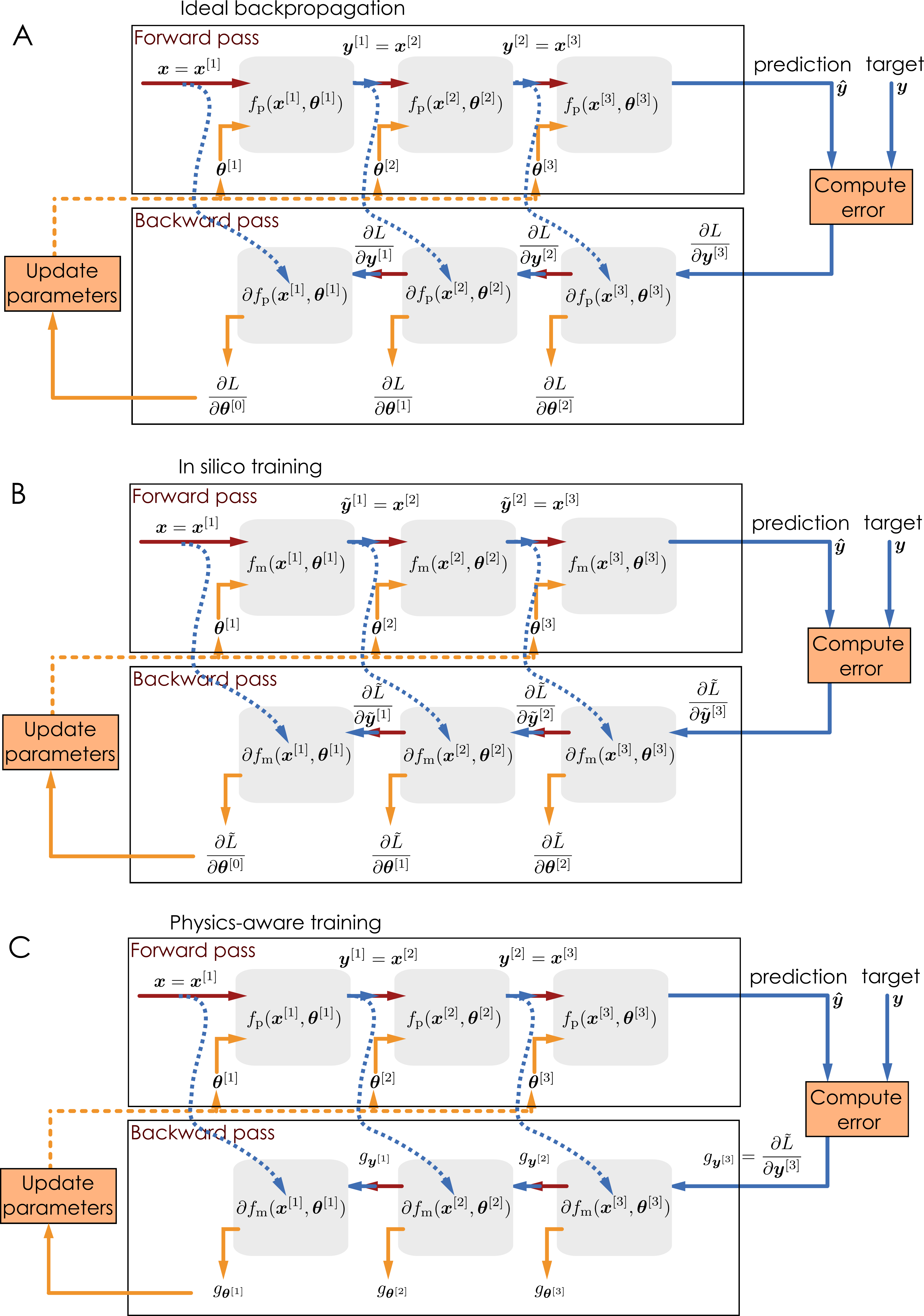}
    \caption{Schematic of the full training loop for the different training algorithms applied to feedward PNNs. All three algorithms are backpropagation algorithms and involves four key steps, namely: the forward pass, computing the error vector, the backward pass and the update of the parameters. The main difference between the algorithms is whether the physical transformation $f_\ph$ or $f_\mo$ is used for the forward and backward passes. For instance for \emph{in silico} training (panel B), $f_\mo$ is used in the forward and backward passes. Thus, the intermediate quantities in this algorithm $\tilde{\bm{x}}^{[l]}$ are different from that of the ideal backpropagation algorithm (panel A) and PAT (panel C). The blue dashed arrows indicate that physical system outputs are saved during the forward pass and the Jacobians will be evaluated at the saved locations.}
    \label{fig:concat_PNN}
\end{figure}

\FloatBarrier
\subsection{Motivating physics-aware training with multilayer feedforward architecture}
\label{sec:pat-mlpnn}

In the previous subsection, we were able to describe physics-aware training in full generality by casting it in the language of automatic differentiation. Thus, we assumed familiarity with automatic differentiation and did not provide a full prescription of the training loop that is performed during physics-aware training. In this subsection, we pedagogically walk-through all the steps involved in physics-aware training for a specific PNN architecture. Here, we specialize to the feedforward PNN architecture as it is the PNN counterpart of the canonical feedforward neural network (multilayer perceptrons) in deep learning. Similar to Sec.~\ref{sec:pat-general}, we begin by reviewing the full training loop for the conventional backpropagation algorithm. We show why it is ineffective for PNNs, before introducing the alternative training algorithms: \emph{in silico} training and physics-aware training.

From this section onwards, we call the conventional backpropagation algorithm applied to the physical transformations in PNNs the ``ideal backpropagation algorithm". This renaming is motivated by two reasons. First, as was pointed out in the last section, physical systems cannot be autodifferentiated, so it is impractical to apply this algorithm to real PNNs. Second, to avoid nomenclature confusion with \textit{in silico} training, which applies the conventional backpropagation algorithm to a model of the system. 

In supervised machine learning, a training dataset defined by the set of pairs of examples and targets $\set{(\bm{x}^{(k)} , \bm{y}^{(k)})}_{k=1}^{N_\text{data}}$ is provided. During training, the goal is to obtain parameters of the network that maps any given example $\bm{x}^{(k)}$ to an output (prediction) $\hat{\bm{y}}^{(k)}$ that is as close as possible to the desired target $\bm{y}^{(k)}$. In gradient-based learning algorithms, this training is performed by minimizing the following overall loss function $L_{\text{total}}$ via gradient descent:
\begin{align}
    L_{\text{overall}} = \frac{1}{N_\text{data}}\sum_k L^{(k)} =  \frac{1}{N_\text{data}} \sum_k \ell(\hat {\bm y}^{(k)} , \bm y^{(k)}),
    \label{eq:overall-loss}
\end{align}
where $\ell$ is the loss function of choice. For instance, in the case of mean-square error loss, $\ell(\hat {\bm y} , \bm y) = |\hat {\bm y} - \bm y|^2$. As the overall loss function is an average over the different individual examples, the gradient of the loss with respect to parameters of the networks are also averaged over individual examples. Thus, following the approach taken in Ref.~\cite{nielsen2015neural}, the following text will focus on a single example for notational simplicity; we will on occasions present expressions encompassing all examples for completeness. 

As shown in Fig.~\ref{fig:ff_pnn}, the feedforward PNN is composed of $N$ physical systems. A given example $\bm{x}$ is fed into the input of the physical system at the first layer $\bm{x}^{[1]} = \bm{x}$. Subsequent physical systems' input are taken from the output of the previous layer. Thus, the forward pass is mathematically given by the following iterative expression:
\begin{align}
\text{Perform Forward Pass} : \qquad  \bm{x}^{[l+1]} = \bm{y}^{[l]} = {f}_\ph(\bm{x}^{[l]}, \bm{\theta}^{[l]}).
\label{eq:bp-forward}
\end{align}
In this architecture, the output of the physical system at the final layer is the prediction produced by the PNN $\hat{\bm{y}} = \bm{y}^{[N]}$.

To train the parameters of the PNN via gradient descent, we require the gradient of the loss $L = \ell(\hat {\bm y}, \bm{y})$ with respect to parameters of each physical system $\pd{L}{\bm\theta^{[l]}}$. These gradients are computed via the backpropagation algorithm, which  iteratively computes the gradients from the ``back'', i.e. from the last layer towards the first layer. This iterative procedure is more efficient than if $\pd{L}{\bm\theta^{[l]}}$ is computed from scratch for each layer, which wastes computation on redundant calculations of many intermediate terms that result from the chain rule. Thus, to begin this process, we first require the error vector $(\pd{L}{\hat{\bm{y}}} = \pd{L}{\bm{y}^{[N]}})$, which is computed via
\begin{align}
\text{Compute Error Vector} : \qquad  \pd{L}{\bm{y}^{[N]}} = \pd{\ell}{{{\bm y^{[N]}}}}({{\bm y^{[N]}}} , \bm y).
\end{align}
The error vector points out the direction the output of the PNN should change in to minimize the loss. (Since the goal is to \textit{minimize} the loss, in practice one updates parameters along the opposite direction of the gradient.) In the case of the MSE loss, the error vector takes a particularly intuitive form given by the difference between the output and the target ($\pd{L}{\bm{y}^{[N]}} = 2(\hat{\bm{y}}-\bm{y})$). This error vector serves as the ``initial'' condition for the following backward pass computation:
\begin{subequations}
    \label{eq:bp-backward}
      \begin{align}[left ={\text{Perform Backward Pass} : \qquad }]
        \pd{L}{\bm{y}^{[l-1]}} & = \paren{\pd{f_\ph}{\bm{x}}(\bm{x}^{[l]}, \bm{\theta}^{[l]})}^\T \pd{L}{\bm{y}^{[l]}} , \label{eq:bp-backward1} \\
        \pd{L}{\bm{\theta}^{[l]}} & = \paren{\pd{f_\ph}{\bm{\theta}}(\bm{x}^{[l]}, \bm{\theta}^{[l]})}^\T \pd{L}{\bm{y}^{[l]}}. \label{eq:bp-backward2}
       \end{align}
\end{subequations}
To gain an intuitive understanding of Eq.~\ref{eq:bp-backward}, it is important to internalize that just like the error vector, the gradients of the loss \wrt to the outputs $\pd{L}{\bm{y}^{[l]}}$ (parameters $\pd{L}{\bm{\theta}^{[l]}}$) points out how the output (parameters) of each layer should be changed to decrease the loss. Thus, Eq.~\ref{eq:bp-backward1} iteratively prescribes how an output in the previous layer $l-1$ should change (to reduce the loss) given information about how an output in the current layer $l$ should change (to reduce the loss). Eq.~\ref{eq:bp-backward2} then takes this computed information about how outputs should be changed, to compute how the parameters in each layer should be changed to induce that desired change in the output of that layer. 

With the gradient of the loss \wrt parameters, the parameters can be updated with any gradient-descent optimizer. Though advanced optimizers such as Adam \cite{diederik2015adam} and AdaDelta \cite{zeiler2012adadelta} typically results in faster training (and are used in our work), this subsection will focus on the canonical gradient descent update rule (without momentum) for simplicity. In this framework, the parameters are updated as follows: 
\begin{align}
  \text{Update parameters} : \qquad  \bm{\theta}^{[l]} \rightarrow \bm{\theta}^{[l]} - \eta  \frac{1}{N_\text{data}}\sum_k\pd{L^{(k)}}{\bm{\theta}^{[l]}},
\label{eq:bp-update}
\end{align}
where $\eta$ is the learning rate and $\pd{L^{(k)}}{\bm{\theta}^{[l]}}$ denotes the gradient vector computed for the $k$th example $\bm{x}^{(k)}$. Consistent with \eqref{eq:overall-loss}, the parameter update involves an average over the different examples. 

Although the ideal backpropagation algorithm (\ref{eq:bp-forward} - \ref{eq:bp-update}) would in principle arrive at good parameters of the physical systems that correspond to low loss, it is not practical in practice as it cannot be performed efficiently. As mentioned in Sec.~\ref{sec:pat-general}, this is because the algorithm requires analytic gradients of the physical transformation $f_\ph$, which can only be approximated via a finite-difference approach. More specifically, each backward call requires $n+p$ number of repeated calls to the physical system, making the training prohibitively slow for PNNs with large number of parameters.  

\emph{In silico} training circumvents this issue by training the parameters of a network with a differentiable digital model $f_\mo$ of the physical transformation. As the digital model is differentiable, the analytic Jacobian required for backpropagation can be computed in roughly the same time required for the forward pass (without repetition). Hence, \emph{in silico} training could also be referred to as training performed in numerical simulations, or training performed exclusively with a numerical model. In \emph{in silico} training, the training happens solely on a digital computer, to arrive at a set of parameters for the PNNs. After training, these parameters are loaded on to physical system for evaluation. Thus, the algorithm is equivalent to performing conventional backpropagation with $f_\mo$ as the autodiff function. Thus, the full training loop is given by
\begin{alignat}{3}
& \text{Perform Forward Pass} &&: \qquad \tilde{\bm{x}}^{[l+1]} = \tilde{\bm{y}}^{[l]} = f_\mo(\tilde{\bm{x}}^{[l]}, {\bm{\theta}}^{[l]}),
\\ & \text{Compute Error Vector} &&: \qquad \pd{\tilde L}{\tilde{\bm{y}}^{[N]}} = \pd{\ell}{{\tilde {\bm y}}^{[N]}}({\tilde {\bm y}}^{[N]} , \bm y),
\\ & \text{Perform Backward Pass} &&: \qquad 
\begin{aligned}
\pd{\tilde L}{\tilde{\bm{y}}^{[l-1]}}  & = \paren{\pd{f_\mo}{\bm{x}}(\tilde{\bm{x}}^{[l]}, {\bm{\theta}}^{[l]})}^\T \pd{\tilde L}{\tilde{\bm{y}}^{[l]}}, 
\\  \pd{\tilde L}{{\bm{\theta}}^{[l]}}  & = \paren{\pd{f_\mo}{\bm{\theta}}(\tilde{\bm{x}}^{[l]}, {\bm{\theta}}^{[l]})}^\T \pd{\tilde L}{\tilde{\bm{y}}^{[l]}},
\end{aligned} \label{eq:ist-backward}
\\ & \text{Update parameters} &&: \qquad {\bm{\theta}}^{[l]}  \rightarrow {\bm{\theta}}^{[l]} - \eta \frac{1}{N_\text{data}}\sum_k \pd{\tilde L^{(k)}}{{\bm{\theta}}^{[l]}}, 
\end{alignat}
where $\tilde{\bm{x}}^{[l]}, \tilde{\bm{y}}^{[l]}$ denotes the input and output of each layer that is predicted by the differentiable digital model. With the predicted output of the PNN ($\tilde{\bm{y}}^{[N]}$), the predicted loss $\tilde L$ is computed and minimized in \emph{in silico} training via  parameter updates governed by gradients of predicted loss $\tilde L$ ($\pd{\tilde L}{{\bm{\theta}}^{[l]}}$) instead of the true loss $L$ ($\pd{L}{\bm{{\theta}}^{[l]}}$). Therefore, the angle between the gradients vectors $\measuredangle (\pd{L}{\bm{{\theta}}^{[l]}}, \pd{\tilde L}{{\bm{\theta}}^{[l]}})$ conceptually characterizes the performance of the \emph{in silico} training algorithm. When this angle is above $90^\circ$, training the PNN (via a parameter update) would decrease the predicted loss, but actually increase the true error, and \emph{degrade} the performance of the PNN.

We use the angle between gradients and the gradient obtained by ideal backpropagation as a conceptual tool here, rather than a readily computable metric of performance since the ideal backpropagation algorithm is impractical to implement in experiment. Ultimately, the best metric of training algorithms is their achieved performance. In addition, since there are usually many equivalently-performing local optima in training large neural networks, it is not necessarily an indication that a backpropagation algorithm performs poorly if it does not follow the exact same training trajectory as ideal backpropagation. This angle is only evaluated in Sec.~\ref{sec:pat-num} as we treat a hypothetical numerical example, which allows for the ``true'' gradient to be computed via autograd. In addition, the gradient vectors of the two differing algorithm must be evaluated for the same parameters and the same PNN input $\bm x$ (example of the ML dataset). Finally, in addition to the angle, the difference between the magnitude of the gradient vectors $(|\pd{L}{\bm{{\theta}}^{[l]}}|- |\pd{\tilde L}{{\bm{\theta}}^{[l]}}|)$ is also an important metric as the magnitude of the gradient vector determines how big a step is taken during training. Thus, if the magnitude of the predicted gradient is vastly off, it can lead to training instabilities.

With the angle metric in mind, it becomes apparent that \emph{in silico} training has a decisive flaw; the error vector $\pd{\tilde L}{\tilde{\bm{y}}^{[N]}}$, which is the initial condition of the backward pass, is computed via the predicted output $\tilde{\bm y}^{[N]}$ instead of the true output $\bm y^{[N]}$. Thus, without an exceptionally precise digital model, this predicted error vector would point in a significantly different direction from the true error vector. This inaccurate error vector would then be backpropagated via \eqref{eq:ist-backward}, to result in bad parameter updates. This problem is exacerbated for deep PNNs, as the difference between the predicted output of the digital model and the physical system scales exponentially \wrt to depth. In addition to the inaccurate error vector, the Jacobian matrices that are used to backpropagate the gradients in the intermediate layers are also inaccurate as they evaluated at the inaccurate points, at $\tilde{\bm x}^{[l]}$ instead of $\bm x^{[l]}$. 

Physics-aware training is specifically designed to avoid the pitfalls of both the ideal backpropagation algorithm and \emph{in silico} training. It is a hybrid algorithm involving computation in both the physical and digital domain. More specifically, the physical system is used to perform the forward pass, which alleviates the burden of having the differential digital models be exceptionally accurate (as in \emph{in silico} training). The differentiable digital model is only utilized in the backward pass to complement parts of the training loop that the physical system cannot perform. 

As described in Section.~\ref{sec:pat-general}, physics-aware training can be formalized by the use of custom constituent autodiff functions in an overall network architecture. In the case of the feedforward PNN, the autodiff algorithm with these custom functions results in and simplifies to the following training loop: 
\begin{alignat}{3}
& \text{Perform Forward Pass} &&: \qquad \bm{x}^{[l+1]} = \bm{y}^{[l]} = {f}_\ph(\bm{x}^{[l]}, \bm{\theta}^{[l]}), \label{eq:pat-forward}
\\ & \text{Compute Error Vector} &&: \qquad g_{\bm{y}^{[N]}} = \pd{L}{\bm{y}^{[N]}} = \pd{\ell}{{{\bm y^{[N]}}}}({{\bm y^{[N]}}} , \bm y) \label{eq:pat-error},
\\ & \text{Perform Backward Pass} &&: \qquad 
\begin{aligned}
    g_{\bm{y}^{[l-1]}} & = \paren{\pd{f_\mo}{\bm{x}}(\bm{x}^{[l]}, \bm{\theta}^{[l]})}^\T g_{\bm{y}^{[l]}},
\\  g_{\bm{\theta}^{[l]}} & = \paren{\pd{f_\mo}{\bm{\theta}}(\bm{x}^{[l]}, \bm{\theta}^{[l]})}^\T g_{\bm{y}^{[l]}},
\end{aligned} \label{eq:pat-backward}
\\ & \text{Update parameters} &&: \qquad \bm{\theta}^{[l]} \rightarrow \bm{\theta}^{[l]} - \eta \frac{1}{N_\text{data}}\sum_k g_{\bm{\theta}^{[l]}}^{(k)},
\label{eq:pat-ffa}
\end{alignat}
where $g_{\bm{\theta}^{[l]}}, g_{\bm{y}^{[l]}}$ are estimators of the ``exact'' gradients $\pd{L}{\bm{\theta}^{[l]}}, \pd{L}{\bm{y}^{[l]}}$ respectively. Thus, analogous to the metric defined for \emph{in silico} training, the angle between the estimated gradient vector and the exact gradient vector $\measuredangle (\pd{L}{\bm{{\theta}}^{[l]}}, g_{\bm{\theta}^{[l]}})$ conceptually characterizes the performance of physics-aware training. In physics-aware training, the error vector is exact $(g_{\bm{y}^{[N]}} = \pd{L}{\bm{y}^{[N]}})$ as the forward pass is performed by the physical system. This error vector is then backpropagated via \eqref{eq:pat-backward}, which involves Jacobian matrices of the differential digital model evaluated at the ``correct'' inputs ($\bm x^{[l]}$ instead of $\tilde{\bm x}^{[l]}$) at each layer. Thus, in addition to utilizing the output of the PNN ($\bm{y}^{[N]}$) via physical computations in the forward pass, intermediate outputs ($\bm{y}^{[l]}$) are also utilized to facilitate the computation of accurate gradients in physics-aware training.

\newpage
\FloatBarrier
\subsection{Numerical example}
\label{sec:pat-num}
In this subsection, we present a numerical example to demonstrate how PAT efficiently trains PNNs. The numerical example presented here is a simplified, simulated emulation of the experimental results presented in Figure 2 of the main manuscript. Thus, we train a PNN to perform the vowel classification task with a feedforward mult-layer architecture, where the constituent hypothetical physical transformation is an autocorrelation of the input and parameters (see \eqref{eq:pat-num-autocor}), which is a crude approximation for the physical transformation realized with broadband optical second harmonic generation. We construct a differentiable digital model for this hypothetical physical transformation, which is then employed in PAT. The performance of PAT is compared against the performance achieved by training with ideal backpropagation and \emph{in silico} training. We show the training curves and how the computation is performed in the forward and backward pass, for the different algorithms. 

In the vowel task, the input vowel formant vectors are 12-dimensional, while the output required for predictions is 7-dimensional, since we consider 7 vowel classes and use the one-hot encoding. To accommodate this, we consider the feedforward PNN introduced previously, adding some minor pre-processing and post-processesing of the initial input and final output respectively. 

More concretely, we consider a feedforward PNN with a constituent physical function $f_\ph$ that has 24-dimensional inputs and outputs. Hence, the input vowel vector is repeated twice to produce a 24-dimensional input for the first layer of the PNN and selected portions of the final output are summed to produce a 7-dimensional vector. Mathematically, the PNN is given by 
\begin{subequations}
\begin{align}
    {\bm x}^{[1]} &= [x_1, x_1, x_2, x_2, \ldots, x_{12}, x_{12}]^\T , 
    \\ \bm{x}^{[l+1]} & = \bm{y}^{[l]} = {f}_\ph(\bm{x}^{[l]}, \bm{\theta}^{[l]}) \qquad \text{for} \ l = 1 \ldots N , 
    \\ \bm o &= [y^{[N]}_{5} + y^{[N]}_{6}, y^{[N]}_{7} + y^{[N]}_{8}, \ldots, y^{[N]}_{17} + y^{[N]}_{18}]^\T. 
\end{align}
\label{eq:pat-num-pnn}
\end{subequations}
where $\bm o$ denotes the final output of the PNN \footnote{Here we denote the output by $\bm o$ instead of $\hat{\bm{y}}$ as the output of the PNN has not been made to approximate the target $\bm y$ since it has not been converted to a probability distribution via the softmax function.} and $y^{[l]}_i$ denotes the value of the neuron $i$ ($i$th index of vector $\bm y^{[l]}$) at layer $l$. As is typical in conventional NNs, the predicted label for this PNN is given by the index of the final output ${\bm{o}}$ that results in the largest value. For the constituent physical function $f_\ph$, we choose a mathematical form that roughly mirrors the transformation for the SHG experiment conducted in Figure 2 of the main manuscript. Hence, $f_\ph(\bm x, \bm\theta) = a\bm{s}/\max(\bm{s}) + c$, where $a$ and $c$ are trainable factors and offsets. Here $\bm{s}$ is an autocorrelation of the control signal, 
\begin{align}
    s_j = \sum_{i=0}^{\min(j, N-j-1)} q_{j-i} q_{j+i}
    \label{eq:pat-num-autocor}
\end{align}
where $\bm{q} = \text{clamp}([x_1, \ldots, x_N, \theta_1, \ldots, \theta_N]^T)$ represents the overall control applied to a given layer, which is clamped between 0 and 1 via the clamp operation ($\text{clamp}(v) = \max(0, \min(v, 1))$). With $f_\ph$ at hand, we train a differentiable model $f_\mo$ of the physical transformation via a data driven approach. A deep neural network (DNN) with 3 hidden layers that have 1000, 500, 300 hidden neurons respectively was trained on 2000 instances of physical input vectors (i.e., the vector containing inputs and parameters) and output pairs generated by repeated application of $f_\ph$.

The PNN is trained by minimizing the following loss function:
\begin{align}
    L = \underbrace{H(\bm y, \hat{\bm y})}_{L_{\text{cross-entropy}}} + \underbrace{\lambda \sum_l \sum_i \max(0, q_i^{[l]} - v_{\text{max}}) -\min(0, q_i^{[l]}-v_{\text{min}}) }_{L_{\text{constraint}}}
\end{align}
where $H$ is the cross-entropy function ($H(\bm p, \bm q) = -\sum_i p_i \log(q_i)$) and $\hat {\bm y} = \text{Softmax}(\bm o)$. The first term ($L_{\text{cross-entropy}}$) is the cross-entropy loss responsible for training the output of the PNN to approach the desired target, while the second term ($L_{\text{constraint}}$) is a regularization term, with regularization coefficient $\lambda$, to constrain the controls $\bm q^{[l]}$ of each layer to be between $v_{\text{min}}$ and $v_{\text{max}}$. For this numerical example $\lambda = 0.02$, $v_{\text{min}} = 0$, and $v_{\text{max}}=1$. 

\begin{figure}[h!]
    \centering
    \includegraphics[width=0.7\linewidth]{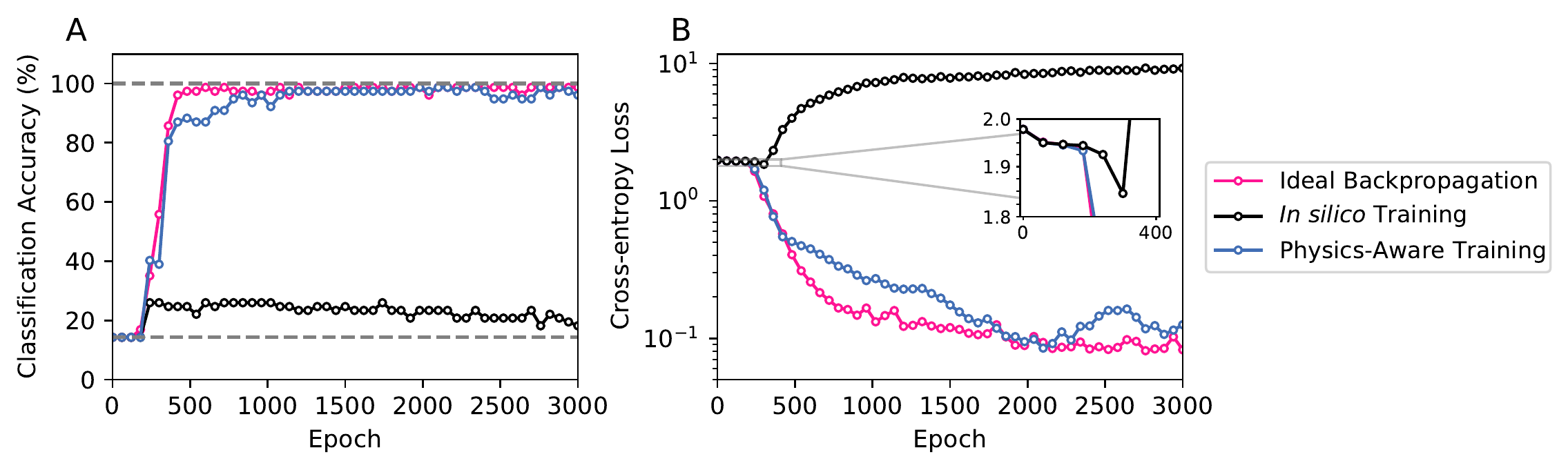}
    \caption{Comparison of the test classification accuracy (A) and test cross-entropy loss (B) versus training epoch for ideal backpropagation, \emph{in silico} training, and physics-aware training. Since this numerical example attempts to mirror the experiment performed in Figure 2 of the main manuscript, it considers the vowel classification task with a feedforward PNN, specified by \eqref{eq:pat-num-pnn}. The inset in B zooms in on the initial training phase, where the cross-entropy loss for the \emph{in silico} training briefly decreases for about 300 epoch before increasing due to compounding mismatch between the $f_\ph$ and $f_\mo$. 
    }
    \label{fig:pat-training-curve}
\end{figure}

With this loss function, the PNN was trained with the ideal backpropagation algorithm, \emph{in silico} training, and physics-aware training. In Fig.~\ref{fig:pat-training-curve}, we show the training curves for the 3 training algorithms. Initially, all 3 algorithms are able to train the parameters to reduce the cross-entropy loss (see inset of Fig.~\ref{fig:pat-training-curve}B). However, by about $\sim$350 epochs, \emph{in silico} training is no longer able to train the physical system effectively due to the increasing differences between the outputs of $f_\ph$ and $f_\mo$. PAT initially follows the training curve of the ideal backprop algorithm quite closely. By about $\sim 500$ epochs, the effective rate of learning for PAT slows down but nevertheless it is able to lower the loss consistently until the test classification accuracy is $\sim$96\%, which is close to that of ideal backprop. After $\sim$2000 epochs, the cross-entropy loss and classification accuracies exhibits fluctuations (this is also seen in Figure 2 of the main manuscript). We speculate that minor differences in the Jacobian of $f_\ph$ and $f_\mo$ causes the parameters to drift around when they are located near an optimal parameter region of the overall loss function. 

We also characterize the inner workings of the PNNs for the different algorithms. To do so, we consider two perspectives. The first looks at how the algorithms differ over just one training step. This perspective, which is presented in Fig. S5, assumes all algorithms initially have the exact same parameters, and we then take 1 training step (i.e., 1 parameter update based on 1 mini-batch of training examples). The second perspective analyzes divergence of training trajectories over many training steps. This perspective, which is shown in Figs. S6 and S7, shows how the single-step inaccuracy of \text{in silico} training compounds through training leading to increasingly worse training of the physical system, and shows how PAT avoids this divergence. 

\begin{figure}[h!]
    \centering
    \includegraphics[width=1.0\linewidth]{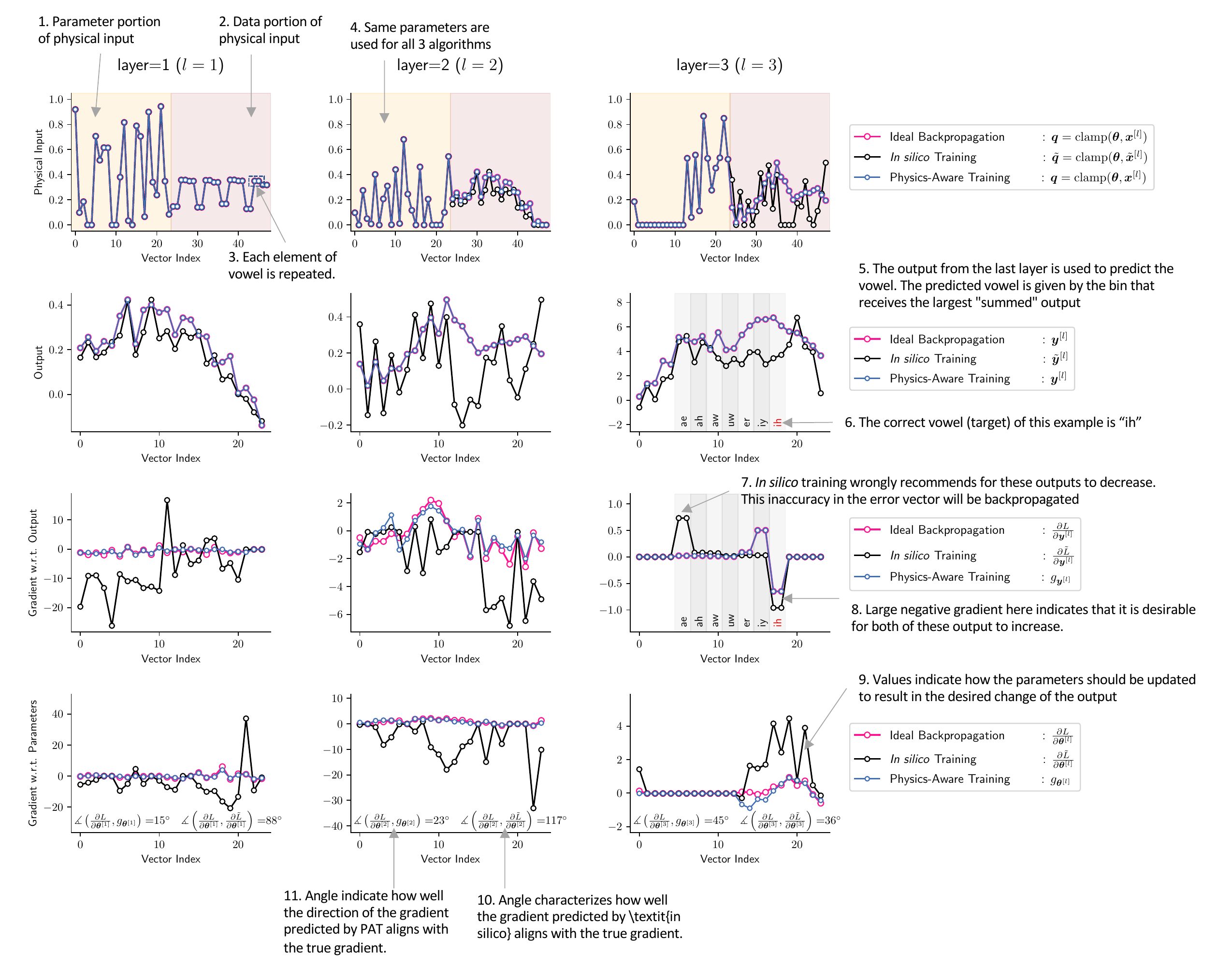}
    \caption{Comparing 3 training algorithms in a numerical example, for a single training step. More specifically, the inputs, outputs, parameters and gradients of the feedforward PNN applied to the vowel classification task are shown in each row of plots respectively, with each physical layer in the first, second and third columns of plots respectively. Here, we transfer the same parameter to all 3 algorithm, to evaluate how similar the gradient vectors are for the different algorithms.}
    \label{fig:pat-annotation}
\end{figure}

In Fig.~\ref{fig:pat-annotation}, we show how the forward and backward passes of the PNN differ across the different algorithms when the parameters are set to initially be the same. Here we observe that the prediction error between the outputs of $f_\mo$ and $f_\ph$ build up progressively through layers during the forward pass. This results in an incorrect error vector for \emph{in silico} training, that is backpropagated to result in incorrect gradients of the loss \wrt parameters. As shown in the last row of Fig.~\ref{fig:pat-annotation}, the gradients of \emph{in silico} training do not resemble the ``true'' gradient of the ideal backpropagation algorithm (which can only be obtained from a finite-difference approach in a laboratory setting), in both direction and magnitude. In contrast, PAT is able to approximate the gradients fairly accurately, as it utilizes the experimental transformation $f_\ph$ in the forward pass. This may be surprising because PAT uses the exact same model as \textit{in silico} training to estimate gradients. The key is that the gradients with PAT are estimated at the correct locations in parameter space (i.e., the inputs to each physical layer are known exactly), whereas in \textit{in silico} training, the gradients are calculated at locations that are increasingly mismatched with reality as one goes deeper into the network.

\begin{figure}[h!]
    \centering
    \includegraphics[width=1.0\linewidth]{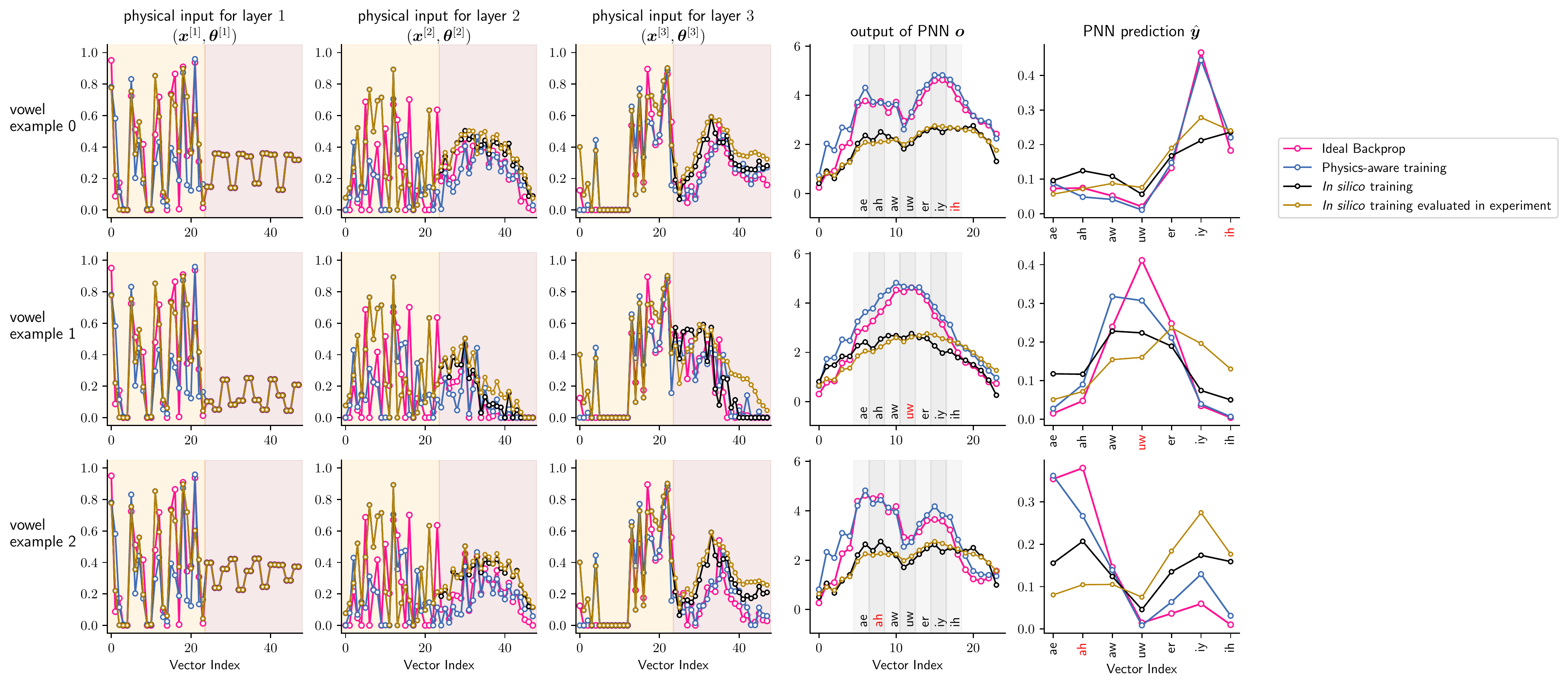}
    \caption{Comparing the training of 3 training algorithms over many training steps. Here, the training has proceeded for 300 steps. More specifically, we show the physical inputs at the different layers, the output and the final PNN prediction of PNNs for three example vowels. In contrast to Fig.~\ref{fig:pat-annotation}, each algorithms here is associated with different parameters that result from each of their own distinct training trajectories shown in Fig.~\ref{fig:pat-training-curve}. Plots associated with ``\emph{In silico training} evaluated in experiments'' refer to characterizations of the PNN when the parameters trained via \emph{in silico} training is transferred over and evaluated with the experiment (with the function $f_\ph$ instead of $f_\mo$).}
    \label{fig:pat-epoch1}
\end{figure}

\begin{figure}[h!]
    \centering 
    \includegraphics[width=1.0\linewidth]{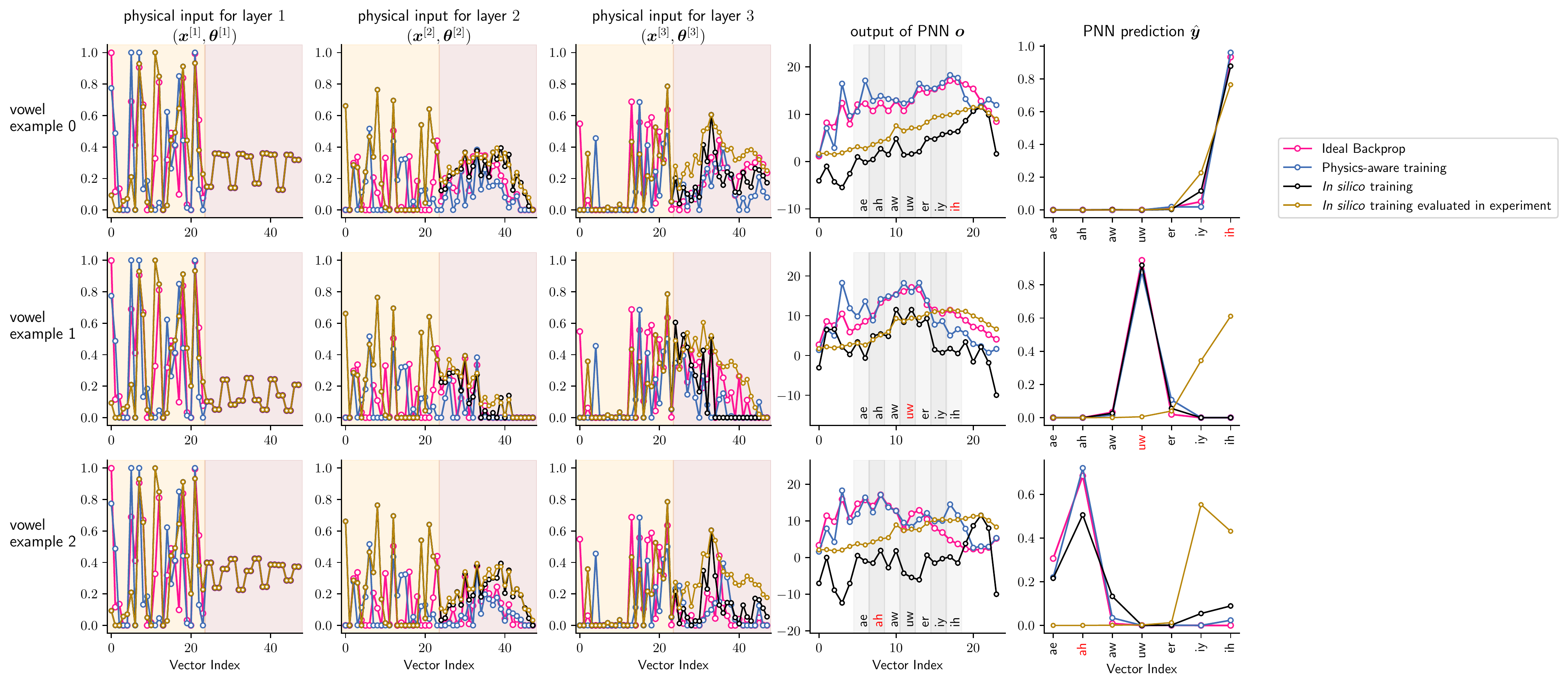}
    \caption{Comparing the training of 3 training algorithms over many training steps. Here, the training has proceeded for 800 steps. More specifically, we show the physical inputs at the different layers, the output and the final PNN prediction of PNNs for three example vowels. In contrast to Fig.~\ref{fig:pat-annotation}, each algorithms here is associated with different parameters that result from each of their own distinct training trajectories shown in Fig.~\ref{fig:pat-training-curve}. Plots associated with ``\emph{In silico training} evaluated in experiments'' refer to characterizations of the PNN when the parameters trained via \emph{in silico} training is transferred over and evaluated with the experiment (with the function $f_\ph$ instead of $f_\mo$).}
    \label{fig:pat-epoch2}
\end{figure}

In Fig.~\ref{fig:pat-epoch1} and \ref{fig:pat-epoch2}, we characterize the PNNs after training for 300 and 800 epochs respectively,  for the different training algorithms. We show how the vowel data propagates through the different layers of the PNNs to arrive at the final prediction $\hat {\bm y}$ for 3 different vowel examples. In Fig.~\ref{fig:pat-epoch1}, we see that the ideal backpropagation algorithm learns the fastest as it is already able to correct classify 2 of out the 3 vowels. The parameters learned by PAT are also reasonably similar to those obtained with ideal backpropagation, and the trained model performs similarly well. \textit{In silico} training meanwhile trains slowly, but by 300 epochs has made some small positive progress. When the PNN is executed with the parameters obtained by \text{in silico} training after 300 epochs, the PNN's output is shifted slightly towards the target vowel. Thus, consistent with Fig.~\ref{fig:pat-training-curve}, learning has occured for all 3 algorithms after 300 epochs of training. As training proceeds, however, the simulation-reality mismatch from each training step compounds, causing \textit{in silico} training to diverge rapidly away from ideal backpropagation and PAT. This can be exacerbated because \textit{in silico} training can often proceed into regions of parameter space where its predictions are especially poor, leading to an even faster divergence from ideal backpropagation and PAT. By epoch 800 (see Fig. S7), the parameters predicted by \textit{in silico} training have drifted into a region where the digital model's predictions are especially inaccurate. This is shown in the rightmost column of Fig.~\ref{fig:pat-epoch2}, where the output of the experiment with parameter trained via \emph{in silico} training classifies the vowel wrongly 2 out of 3 times, while the PAT and ideal backpropagation is able to classify all 3 vowels accurately with a reasonable high confidence, which is indicated by the value of $\hat {\bm y}$ being close to 1 for the correct vowel. 

\clearpage
\section{Supplementary methods}
In this section, we first describe the experimental setups for each of the three physical systems used to implement PNNs. We then describe our data-driven differentiable models, as well as the noise models used to improve the performance of \textit{in silico} training. Finally, the PNN architectures used in the main text for classifying vowels and MNIST digits are described and typical results are presented to illustrate how the PNNs perform the learned classification tasks.

\subsection{Femtosecond second-harmonic generation PNN}
\subsubsection{Experimental setup}

\begin{figure}[h!]
    \centering
    \includegraphics[width=0.8\linewidth]{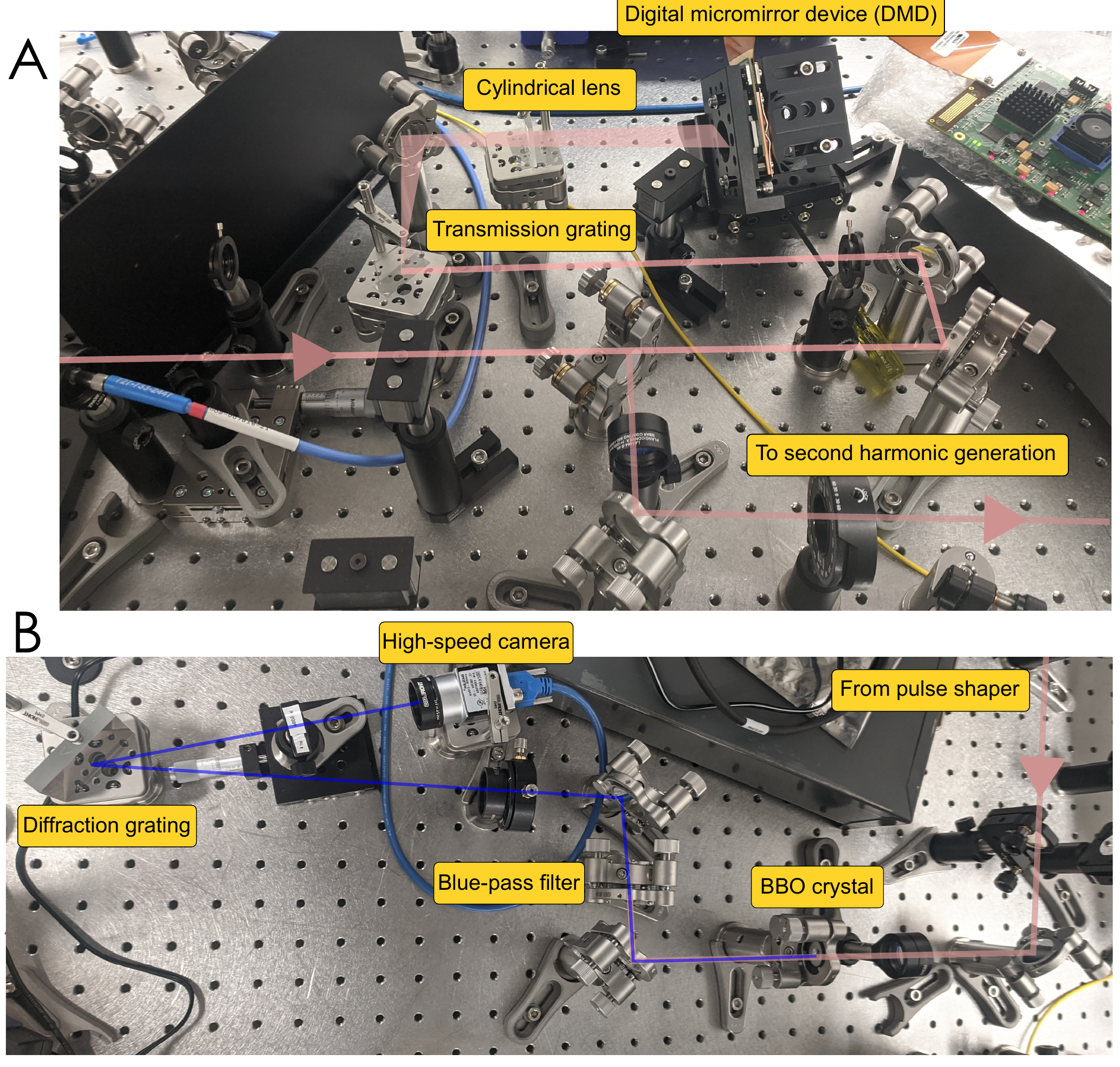}
    \caption{Labelled photos of the pulse shaper and second harmonic generation experimental setups.}
    \label{fig:SHG setup}
\end{figure}

Our femtosecond second-harmonic generation (SHG) PNN is based on a mode-locked Titanium:sapphire oscillator (Spectra Physics Tsunami) which is operated on by a digital micromirror device (DMD, Vialux V-650L)-based pulse shaper to produce complex, amplitude-modulated pulses according to input data and parameters. The spectrally-modulated pulses, centered around 780 nm, are then focused into an 0.5 mm long BBO crystal where SHG occurs. After filtering the residual 780 nm light, the resulting blue ($\sim$ 390 nm) light is diffracted by a ruled reflection grating and focused onto a fast camera (Basler ace acA800-510u) to measure the SHG spectrum. Annotated photos of the setup are shown in Supplementary Figure \ref{fig:SHG setup}.

Training PNNs requires executing the physical evolution millions of times, so it was prudent to design the setup to allow for high-throughput operation. Based on the design described above, we were able to operate the SHG-PNN at a sustained rate of 200 Hz, reliably over months (including remotely during the COVID-19 pandemic). 

The DMD-based pulse shaper achieves continuous spectral amplitude modulation in the following way. Light diffracted by a transmission grating (Ibsen Photonics PCG-1765-808-981) is focused by a cylindrical lens (150 mm, Thorlabs) onto the DMD, which is mounted at 45 degrees from the vertical so the micromirrors' rotation axis is vertical with respect to the table. The DMD is then rotated slightly so that the light reflected by the mirrors in the `0' state retroflects backward through the cylindrical lens and grating, following an identical path except with a slight downward angle so as to be separated by a pick-off mirror. By writing vertically-oriented (with respect to the table, 45 degrees to the DMD's axes) gratings of varying duty cycle to the DMD, we can achieve multi-level amplitude modulation at each spectral position (the frequency components are dispersed across the DMD in the direction parallel to the table by the grating). The design of this pulse shaper was inspired by Ref. \cite{liu2017programmable}, and to control the DMD using Python we adapted code developed for Ref.~\cite{matthes2019optical} available at Ref.~\cite{sebastien_m_popoff_2020_4076193}. 

While such a design allows continuous amplitude modulation at significantly higher speeds than most spatial light modulator-based pulse shapers, it results typically in pulses with complex spatiotemporal coupling which would be challenging to account for with traditional simulations. In this regard, our data-driven digital models were helpful in allowing the physical apparatus to be designed to optimize the reliability and speed of the experiment, rather than to ensure for the physics to easily be modelled by traditional simulations. 

\begin{figure}
    \centering
    \includegraphics[width=0.8\linewidth]{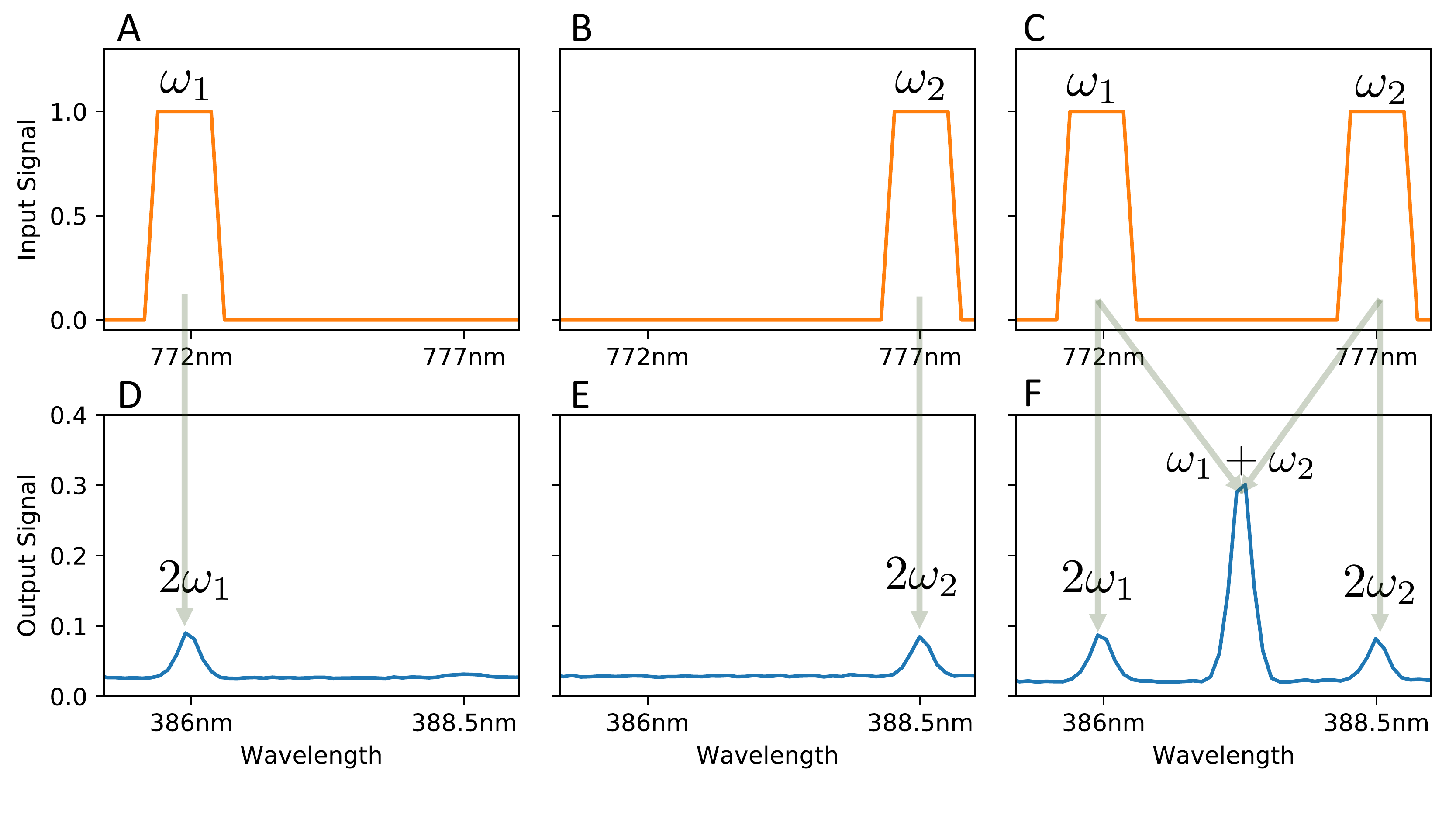}    
    \caption{Characterization of the femtosecond second harmonic generation (SHG) experimental setup with simple inputs. A-C. Sharply peaked input signals to the pulse shaper, which results in continous-wave like input waves for SHG. D-F. The resulting output spectrum of femtosecond SHG for their respective input signals. }
    \label{fig:iot-shg-nl}
\end{figure}

\begin{figure}
    \centering
    \includegraphics[width=0.8\linewidth]{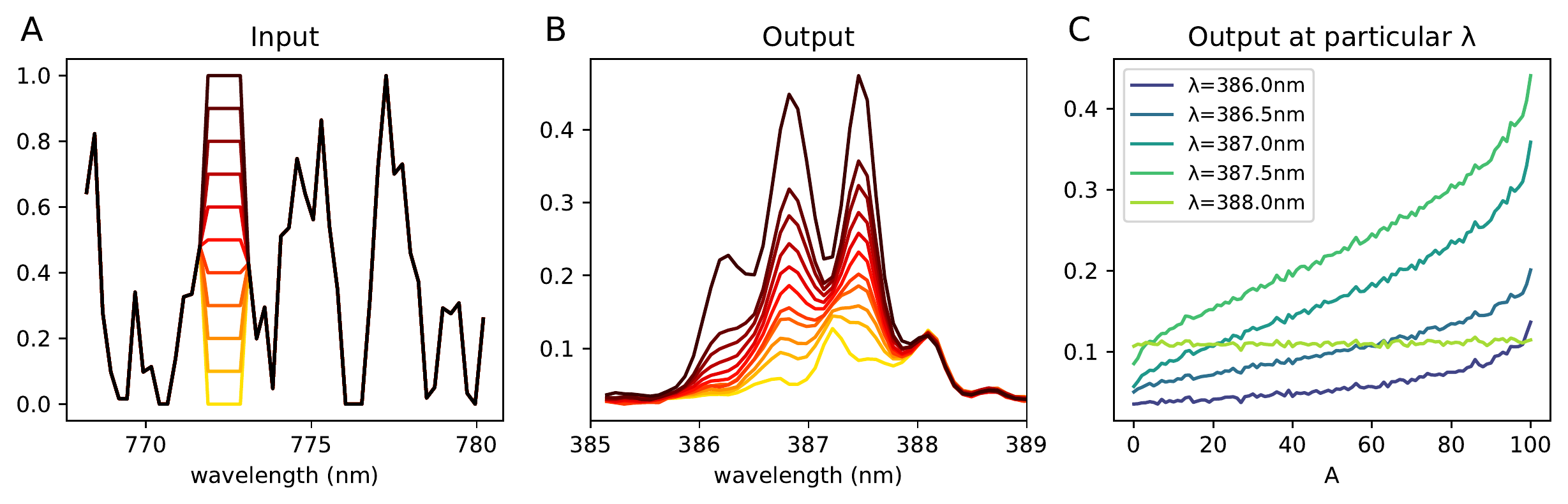}
    \caption{Response of femtosecond SHG device to a complex-waveform input whose spectrum has a small flat region with an amplitude that is varied. A. Waveform for pulse shaper to apply to optical pulse, which are the physical inputs of the SHG system. Different colors correspond to different values $(A)$ that the small flat region takes on. B. Output spectrum from SHG. Each curve of a particular color, is the output of the SHG system subject to input signals with the same color in subpanel A. C. Slices through output spectrum as a function of the input section amplitude $A$, which shows the nonlinearity of the transformation implemented by SHG.}
    \label{fig:iot-shg-com}
\end{figure}
\subsubsection{Input-output transformation characterization}
\label{sec:iot-shg}
Here, we characterize the input-output transformation of femtosecond SHG. In Fig.~\ref{fig:iot-shg-nl}, we characterize the SHG setup with simple quasi-cw inputs to elucidate the basic physics of the device. 

In Fig.~\ref{fig:iot-shg-nl} A and B, we consider continuous-wave-like inputs that are sharply peaked in wavelength. The corresponding SHG outputs in Fig.~\ref{fig:iot-shg-nl} D and E are also sharply peaked in wavelength, with half the wavelength. In panel C, we consider an input that is the sum of input A and B; the output shown in panel F, illustrates the nonlinear nature of SHG, as it is not the sum of output D and E. There is an additional strong output at a center wavelength resulting from combining both input beams at $\si{772nm}$ and $\si{777nm}$. 

Having characterized its basic behavior, we illustrate the complex controllable transformation of the SHG device in Fig.~\ref{fig:iot-shg-com}, by showing its response to a complex input waveform. Figure~\ref{fig:iot-shg-com}A shows input waveforms in which a small section has been held constant to a fixed amplitude $A$, for varying values of $A$ between 0 and 1. Figure~\ref{fig:iot-shg-com}B shows the resulting output spectrum from the SHG, which illustrates the complexity of the spectrum; The output can be significantly different across a wide portion of the spectrum despite the changed portion in the input being relatively small. Figure~\ref{fig:iot-shg-com}C plots the output spectrum at a particular wavelength for varying values of the amplitude $A$, which also shows that the nonlinearity is strongest for large amplitude $(A \approx 1)$.

\subsection{Analog transistor PNN}
\subsubsection{Experimental setup}

The electronic PNN experiments were carried out with standard bulk electrical components, a hobbyist circuit breadboard, and a high-speed DAQ device (Measurement Computing USB-1208-HS-4AO), which allowed for a single 1 MS/s analog input and output. A schematic of the setup is shown in Supplementary Figure \ref{fig:electronic setup}.

A variety of circuits were explored during the course of this work. The one used ultimately for the MNIST digit classification task was designed to be as simple as possible while still exhibiting very strongly nonlinear dynamics. The inclusion of a RLC oscillator within the circuit was inspired by the relative success of the oscillating plate (described in the next section), which we found was easily able to achieve good performance on the mostly-linear MNIST task. 

\begin{figure}[h!]
    \centering
    \includegraphics[width=0.4\linewidth]{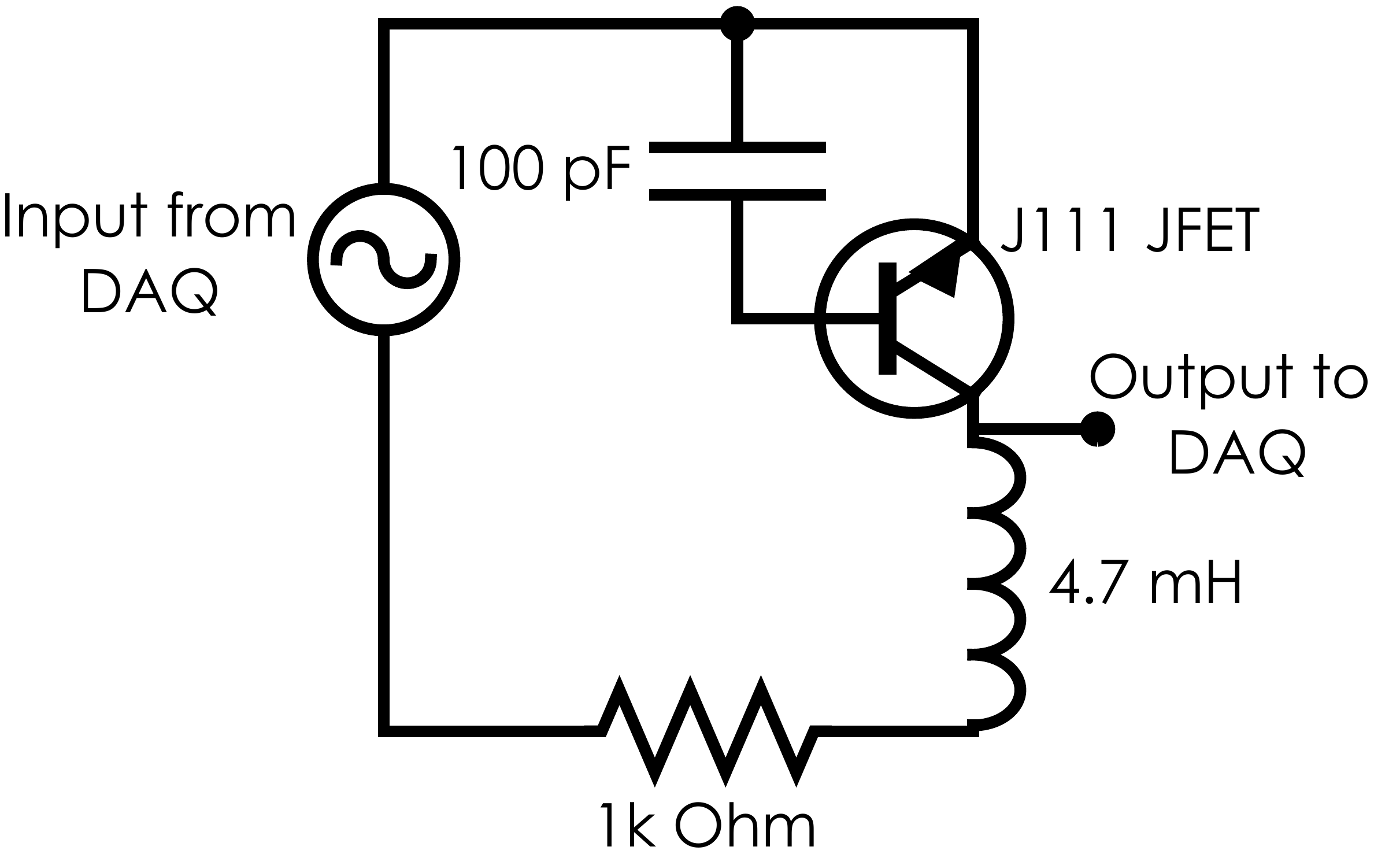}
    \caption{Schematic of the analog electronic circuit used for the electronic PNN.}
    \label{fig:electronic setup}
\end{figure}

\subsubsection{Input-output transformation characterization}
\begin{figure}[h!]
    \centering
    \includegraphics[width=\linewidth]{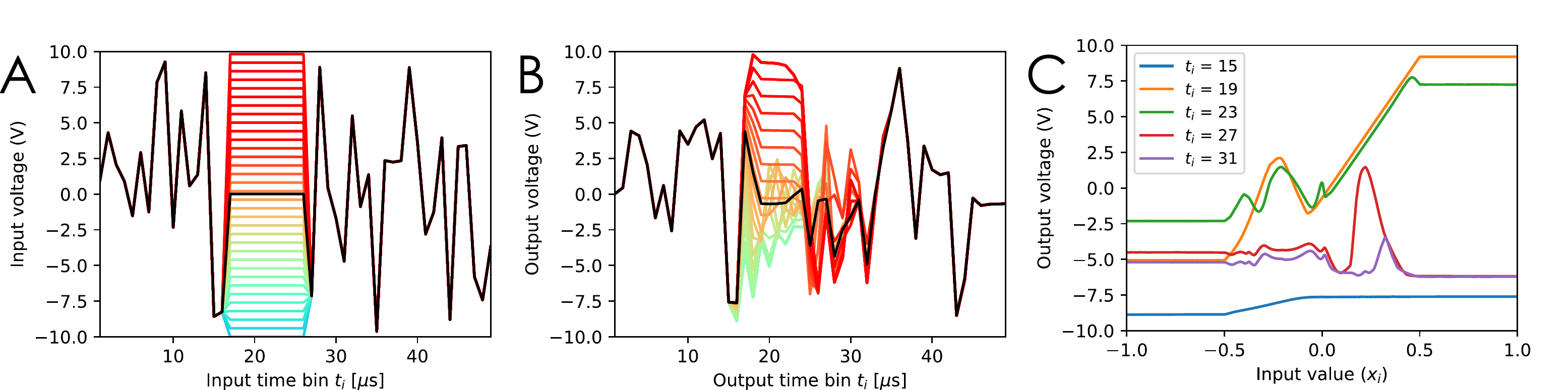}
    \caption{Response of nonlinear analog transistor circuit to a complex waveform with one varied constant section. A. Input waveforms (different colors correspond to different input waveforms, where a section in the middle has been held constant to different fixed voltages). B. Output waveforms (different colors correspond to the different input traces in A). C. Slices through output waveforms as a function of the input section amplitude show the strong nonlinearity of the circuit's transformation of the input. For the last figure, the x-axis is plotted in the calibrated, normalized units used within the PNN PyTorch code,  which approximately correspond to -10 to 10 V range from -1 to 1.}
    \label{fig:electronic nonlinearity}    

\end{figure}

\begin{figure}[]
    \centering
    \includegraphics[width=0.67\linewidth]{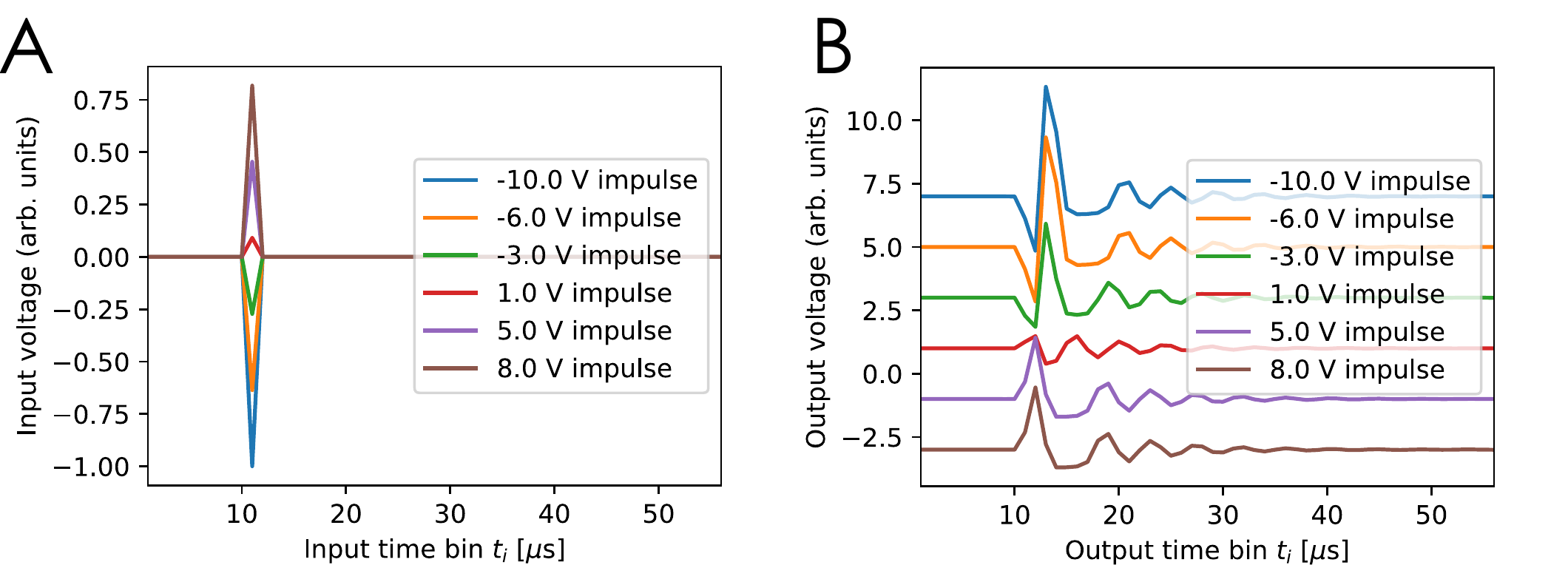}
    \caption{Impulse response of the nonlinear analog transistor circuit to impulses of varying amplitude. The circuit's response is that of a nonlinear oscillator. In B, the different curves are offset artificially to improve visual clarity.}
    \label{fig:electronic impulse}
\end{figure}

Here we briefly discuss characterization of the circuit's response. Supplementary Figure \ref{fig:electronic impulse} shows the impulse response of the circuit to an impulse with a varying amplitude. Supplementary Figure \ref{fig:electronic nonlinearity}A shows a complex input waveform in which a section has been held constant to a fixed amplitude $A$, for varying values of $A$ from -10 to 10 V. Supplementary Figure \ref{fig:electronic nonlinearity}B shows the resulting output time series, illustrating both the nonlinearity of the response as well as the complex coupling between outputs after the changed value. Finally, Supplementary Figure \ref{fig:electronic nonlinearity}C illustrates the nonlinearity of the circuit's response more clearly by plotting the output voltage of the indicated output time bins as a function of the input amplitude voltage $A$. The individual curves are highly nonlinear with respect to the input voltages. 

At the highest input amplitudes in Supplementary Figure \ref{fig:electronic nonlinearity}C, an additional nonlinearity due to the limiting of the DAQ can be observed. While this nonlinearity is not strictly a part of the electronic circuit, we opted to allow the PNN to operate in this regime to illustrate how PNNs can inherently utilize such ``undesired" nonlinearites, and because the circuit's response was already highly nonlinear, so we did not expect this additional nonlinearity to significantly change the circuit's computational capabilities (in contrast, it was important to avoid this DAQ-based nonlinearity in our oscillating plate experiments described next, since the oscillating plate's input-output response was determined to be very linear). 

\FloatBarrier
\subsection{Oscillating plate PNN}

\subsubsection{Experimental setup}

The setup of the oscillating plate PNN consists of an audio amplifier (Kinter K2020A+), a commercially available full-range speaker, a microphone (Audio-Technica ATR2100x-USB Cardioid Dynamic Microphone), and a computer controlling the setup. 

\begin{figure}[h!]
    \centering
    \includegraphics[width=0.4\linewidth]{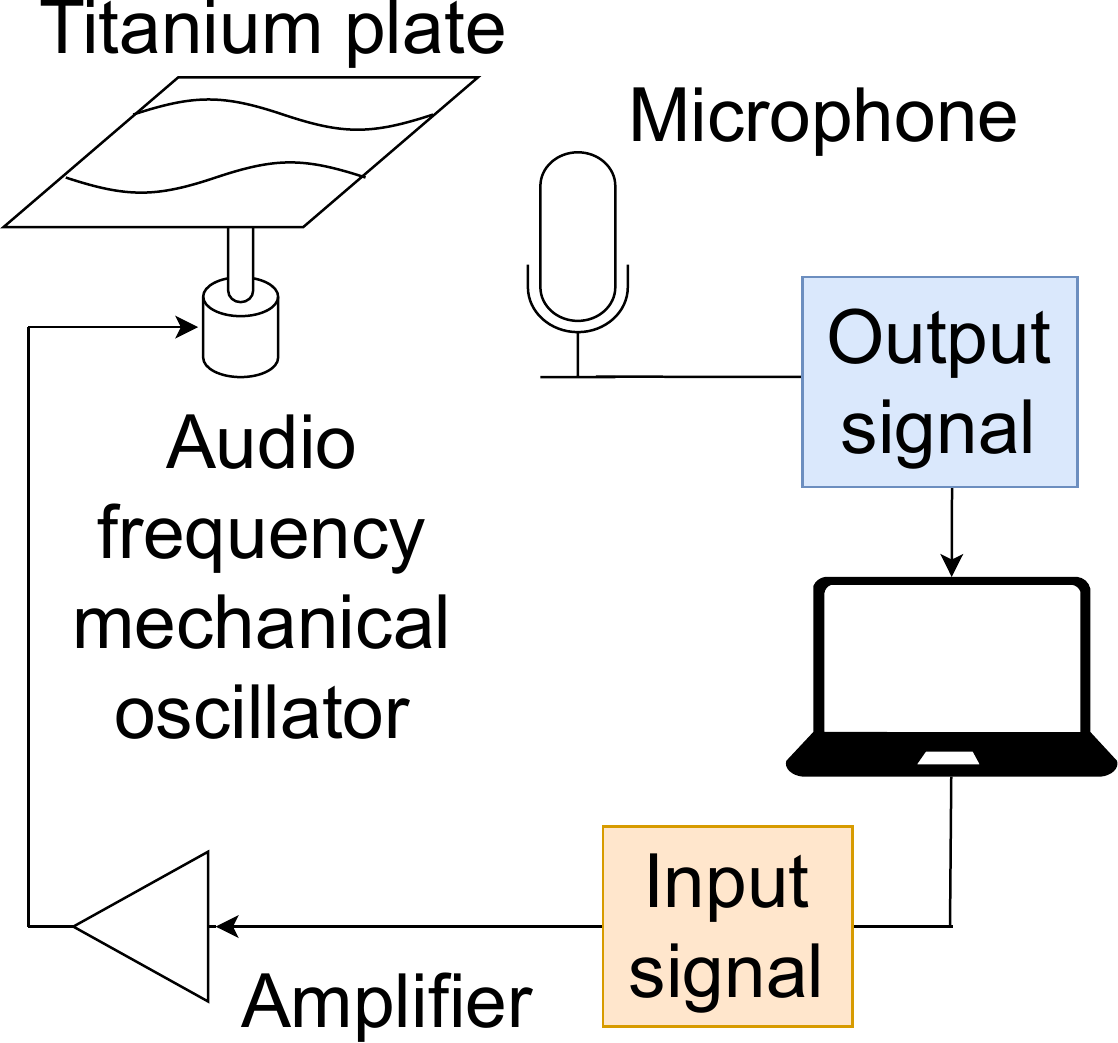}
    \caption{A schematic of the information flow in the oscillating plate experimental setup. An input and control signal from the computer is amplified and applied to the mechanical oscillator (realized by the voice coil of an acoustic speaker). A microphone records the sound produced by the oscillating plate and returns it to the computer.}
    \label{fig:speaker_setup}
\end{figure}

We use the speaker to drive mechanical oscillations of a $3.2\text{ cm} \times 3.2\text{ cm} \times 1\text{ mm}$ titanium plate that is mounted on the speaker's voicecoil.
The diaphragm has been removed from the speaker such that most of the sound produced stems from the oscillations of the titanium plate.
In Supplementary Figure \ref{fig:speaker_fabrication}, the steps taken to construct the oscillator with the mounted plate are shown.

\begin{figure}[h!]
    \centering
    \includegraphics[width=0.75\linewidth]{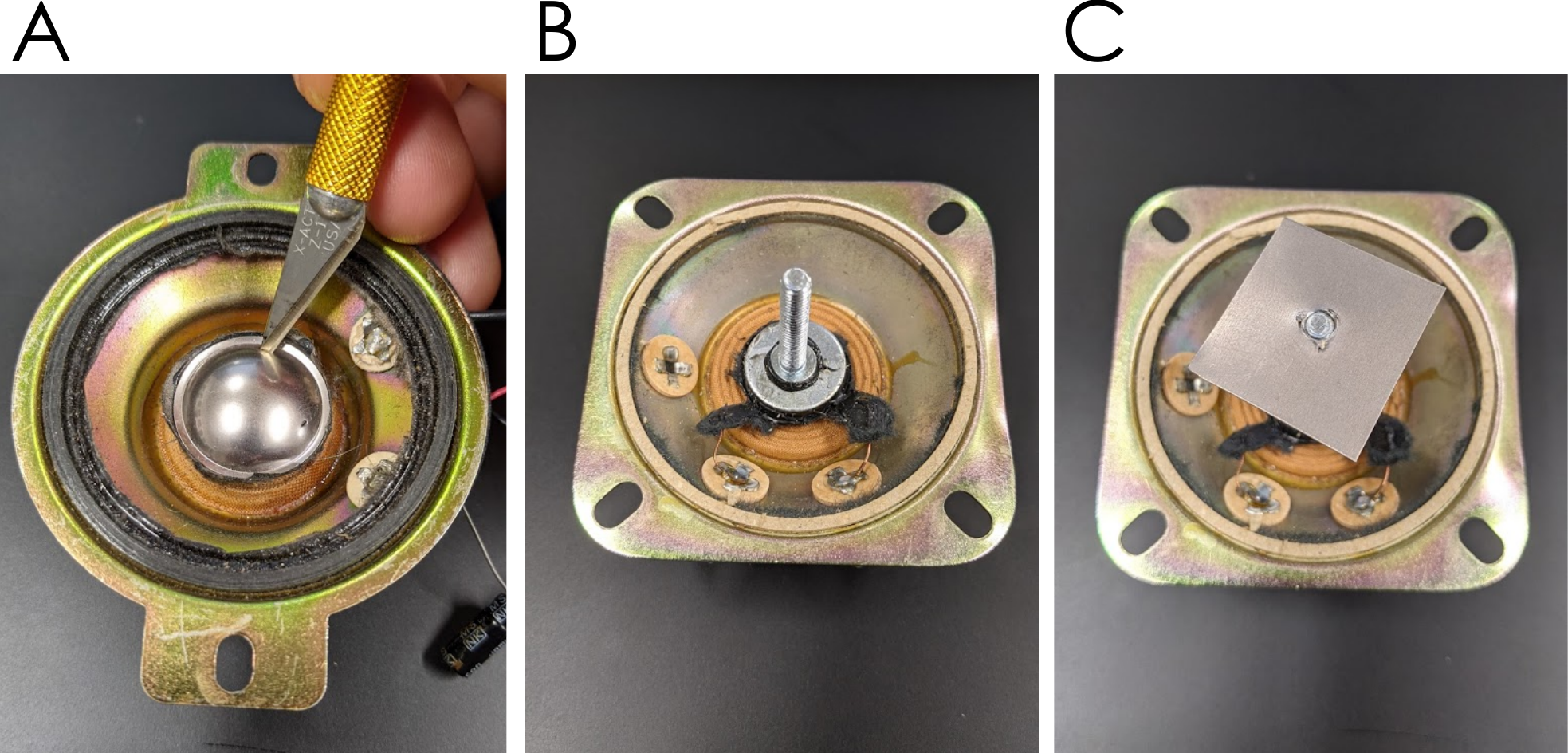}
    \caption{Photographs of steps taken to construct an audio-frequency mechanical oscillator from a commercially available speaker. A. First, we remove the diaphragm and dust cap from a speaker with a precision knife to expose the voice coil. B. Next, we glue a screw and washer to the voice coil with commercially available two component glue. C. After letting the glue dry for 24 hours, we attach the titanium plate on the screw. Finally, we securely mount the speaker on a stable surface to suppress vibrations of the whole device. Different speakers were used over the course of the experiment and not all speakers shown above are the same model.}
    \label{fig:speaker_fabrication}
\end{figure}

\subsubsection{Input and output encoding}

\begin{figure}[h!]
    \centering
    \includegraphics[width=0.75\linewidth]{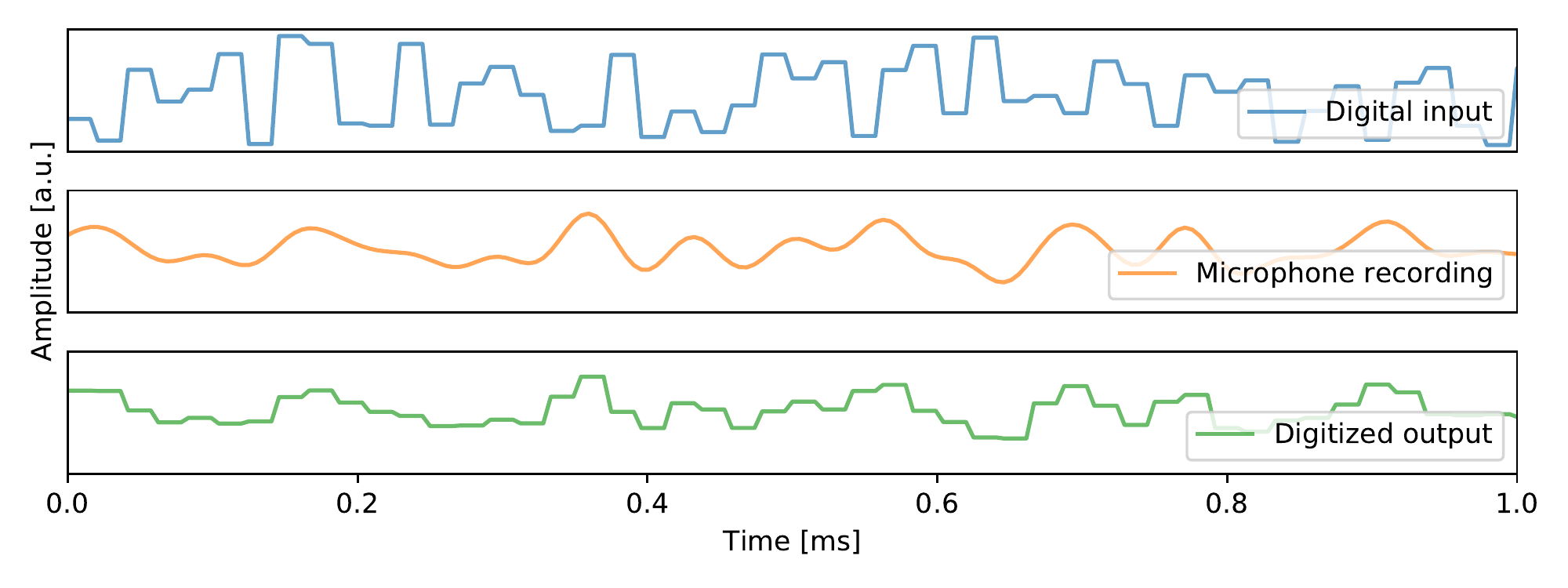}
    \caption{An example of a 48-dimensional digital input to the oscillating plate setup (blue). After driving the oscillating plate, the signal is recorded by the microphone (orange) and digitized in time to the desired output dimension, here 24 (orange). Y-axis units are arbitrary normalized amplitudes.}
    \label{fig:speaker_io_encoding}
\end{figure}

Similar to the electronical PNN, we encode the physical inputs in the time domain.
Supplementary Figure \ref{fig:speaker_io_encoding} shows the different steps of encoding and decoding signals in the oscillating plate PNN. First, inputs are encoded in a time-series of rectangular pulses that are transformed into an analog signal at 192 kS/s by the laptop's soundcard DAC. The signal is amplified by an audio amplifier and drives the voicecoil of a speaker that in turn drives the oscillating titanium plate. We ensured that both the soundcard DAC and the pre-amplifier are operated in a regime where their input-output relation is completely linear. The signal arriving at the oscillating plate was therefore a linearly amplified and slightly low-pass filtered version of the input and control signal created by the computer.
The oscillations of the plate produce soundwaves which are recorded by a microphone and converted into digital signals at 192 kS/s. The signal is further downsampled by partitioning it into a number of equally long subdivisions and averaging the signal's amplitudes over this window. The length of the window is determined by the desired output dimension of the PNN layer. 

Inputs and outputs are synchronized by a repeating trigger signal which precedes every sample that is played on the speaker. By overlapping the trigger signals we can synchronize samples to about \SI{5}{\micro\second} (1/sampling frequency).

\subsubsection{Characterization of input-output transformation}

While we initially aimed to create high-amplitude, nonlinear oscillations on the titanium plate, we realized that we could not reach such amplitudes with our setup. In Supplementary Figure \ref{fig:OPlinearity}, C, we observe that in the time-domain the output voltage's response is overwhelmingly linear with respect to the input voltage.

Neglecting noise and assuming the oscillations of the plate are completely linear, each input to the plate can be regarded as exciting a band of modes between the frequency of the input and the Nyquist frequency of the digital sampler (192 kHz). Following their excitation, the modes decay in a transient oscillation depending on the damping of the material, thereby affecting the following outputs in a way that can be expressed as a convolution: $y_k = \sum_{0<j<k}^N c_{k-j} x_{j+1}$. Here $y_k$ is the $k$th output, $x_j$ is the $j$th input, and the $c_j$ coefficients are constants determined by the medium of the oscillation and the excited modes. 

As a matrix operation, the map from inputs to outputs can be approximated by:

\begin{equation}
    \label{eq:OPmatrix}
    \begin{pmatrix}
        y_{1} \\
        y_{2} \\
        y_{3} \\
        \vdots \\
        y_{d} 
    \end{pmatrix}     
    = 
    \begin{pmatrix}
        c_{1}  & 0      & 0      & \dots  & 0 \\
        c_{2}  & c_{1}  & 0      & \dots  & 0 \\
        c_{3}  & c_{2}  & c_{1}  & \dots  & 0 \\
        \vdots & \vdots & \vdots & \ddots & \vdots \\
        0      & 0      & 0      & \dots  & c_{1}
    \end{pmatrix}     
    \begin{pmatrix}
        x_{1} \\
        x_{2} \\
        x_{3} \\ 
        \vdots \\
        x_{d} 
    \end{pmatrix}     
\end{equation}

For a typical input signal to the oscillating plate PNN, the resulting output is shown in Fig. \ref{fig:OPlinearity}, A and B.
In Fig. \ref{fig:OPlinearity}A, eleven input time-series that are identical except for 20 inputs at around $0.3 \si{ms}$ are plotted. Inputs change with a frequency of $192$ $\si{kHz}$. The transient oscillation set in motion by the inputs at $0.3 \si{ms}$ persists for about $2 \si{ms}$ (not completely shown in the figure). Hence, at this input frequency, an output of the oscillating-plate PNN is a convolution of approximately $N=384$ previous inputs. If we encoded inputs at a slower frequency, for example $5$ $\si{kHz}$, we would only convolve the previous $N=10$ inputs. Therefore, by controlling the frequency of inputs, the oscillating plate emulates the operation of a convolution with variable kernel size and fixed kernel coefficients. This feature was utilized by designing our PNN for MNIST handwritten digit classification.

\begin{figure}[h!]
    \centering
    \includegraphics[width=\linewidth]{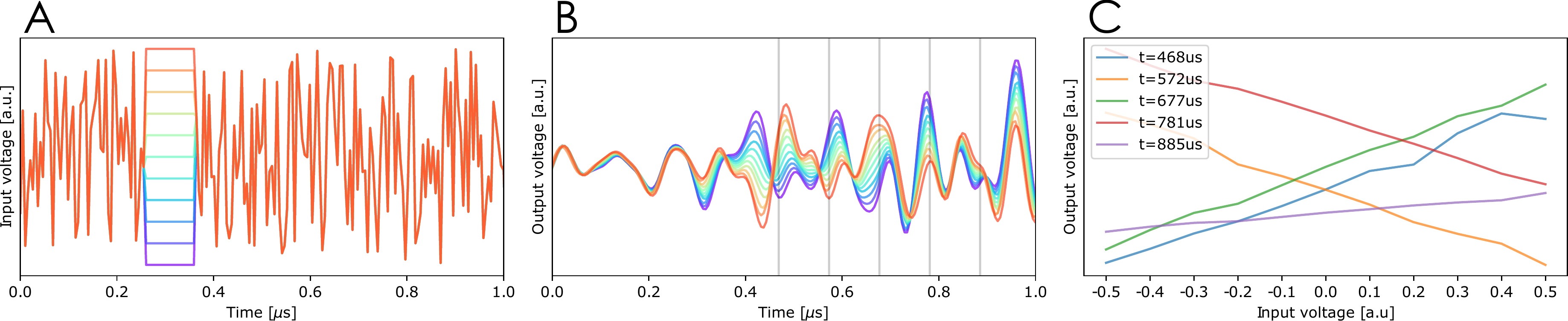}
    \caption{Response of oscillating plate to a complex waveform with one varied constant section. A. Input waveforms. B. Output waveforms. Due to causality, all outputs before \SI{50}{ms} are the same (up to noise). Afterwards, different input voltages produce different transient oscillations. C. Slices through output waveforms as a function of the input voltage (in arbitrary units) show the overwhelmingly linear response of the output to the input voltage.}
    \label{fig:OPlinearity}
\end{figure}

\subsection{Data-driven differentiable models}
\label{sec:data-driven-models}

\subsubsection{Digital model for the mean}
As explained in Section 1, to approximate the gradients of physical systems, one needs a differentiable digital model to perform autodifferentiation on in the backward pass of PAT. In principle, any differentiable model could be used, such as a traditional differential equation physics model, some other analytic model, a data-driven model like a deep neural network, or a physics-informed neural network. In this work, we found that using neural architecture search \cite{akiba2019optuna} to find optimal deep neural network architectures allowed for the most accurate digital models across the physical systems we considered, and thus this was our default approach. 

Of course, we expect that including physical insight into the differentiable digital model will usually result in better performance for any specific system, or at least allow for faster training of the digital model. Since any system used as a PNN will be ideally capable of being executed many times very quickly, we expect that it will usually be beneficial to include significant data-driven components in the digital model, since obtaining a large volume of training data will be straightforward. In some cases conceivable in the future, however, it may be more challenging to rapidly change the physical system's parameters, and the input-output dimension of the physical system used may be extremely high, such that even a large training data set significantly undersamples the space. In these and other cases, it is obvious that using physical insight to permit accurate differentiable digital models with relatively little training data will become essential. 

The accuracy of the DNN models we obtained for the SHG and nonlinear analog circuit systems will be shown and commented on in later sections. Here, we will give a summary overview of our methods and the models ultimately used. 

We first collected training data by recording the outputs of the physical system for a selection of $N$ inputs/parameter vectors, where $N$ was usually $\sim 10^4$ and the input/parameter vectors were usually drawn from a uniform distribution. We also investigated more sophisticated input distributions, but found these were generally unnecessary to produce good results (at least for the results considered in this work). This data set, consisting of $N$ input/parameter vectors and their corresponding output vectors, was split into training and validation sets and used to train neural networks whose hyperparameters were found through a neural architecture search.  

To perform our neural architecture searches, we used the Optuna package \cite{akiba2019optuna} with the following typical search space:
\begin{enumerate}
    \item Number of layers between 1 and 5
    \item Each layer width between 20 and 1500 units
    \item Initial learning rates $10^{-5}$ and $10^{-2}$
\end{enumerate}

The swish activation function was found to perform best, and we used the Adam optimizer.

For the SHG system, the optimal architecture was found to be a deep neural network with the following layer dimensions: [100, 500, 1000, 300, 50].  

For the electronic circuit, the optimal architecture was found to be a deep neural network with the following layer dimensions: [70, 217, 427, 167, 197, 70].

For the oscillating plate, we initially tried to find a deep neural network architecture. However, once we realized the system dynamics were linear, we instead adopted a fully linear model, which was implemented in our deep neural network training framework by creating a network with no hidden layers. 

\subsubsection{Noise digital model}
The digital models we used also include a model of the physical system's noise (including its dependence on the particular input/parameter vector to the system). We used these noise models for two key reasons. First, we were often interested in simulating PNN architectures and wanted to have a reasonably accurate prediction of their performance in the experiment. This was especially useful for slow physical systems, like the oscillating plate, for which MNIST experiments were much too slow to allow us to perform systematic examination of different architectures. More importantly, we wanted to ensure that we maximized the performance of \textit{in silico} training, and providing an accurate model of the system's noise is a logical, standard technique to help mitigate the simulation-reality gap. \textit{All the results for training with simulation presented in this work (including in Figure 3) were obtained with a simulation model of the physical system noise.}

We find however that while including a sophisticated model of physical system's noise in a simulation does improve the ability of \textit{in silico} training to estimate the approximate performance level of results obtained in the lab, including noise in the simulation does not allow \textit{in silico} training to perform as well as PAT for training the actual physical system. 

A total digital model incorporating both mean-field and noise predictions needs to learn the following stochastic map: $y = f_\text{mean}(x) + n_\text{noise}(x)$, where $x\in \mathrm{R}^{N_{\text{in}}}$ is the input to the PNN, and $y$ is a non-deterministic output from the PNN. This output has a deterministic component (given by the mean-field digital model $f_\text{mean}$), and a ``noise" component $n_\text{noise}(x)\in \mathrm{R}^{N_{\text{out}}}$ whose distribution must agree with the true physical system's input-dependent noise.

Note that the digital model needs to learn the noise distribution from \emph{any} given input/parameter vector $x$ (for brevity, we will refer to $x$ as just the input vector from here onward, but note that it includes both the parameters and input data to the physical system). For simplicity, we first outline how to learn the noise distribution for a given fixed input vector, denoted as  $x^\text{(fixed)}$ before addressing the more general case of predicting the noise for any input vector.

To learn the noise distribution for this fixed input, we first collect samples of outputs for the same fixed input. We execute the physical system $N_\text{repeat}$ times with the same input vector $x^\text{(fixed)}$. The output vectors will be denoted $y^{(fixed, k)}$ where $k=1,...,N_\text{repeat}$. From experimental measurements, we have ascertained that the output distribution of the physical systems considered in this work are reasonably well described by a multivariate normal distribution. Thus, their entire noise distribution is well characterized by the mean vector $\mu_i^{(fixed)} = \sum_k y^{(fixed, k)}_i$ predicted by the mean-field digital model, and a covariance matrix $\Sigma^{(fixed)}_{ij} = \frac{1}{N_\text{repeat}-1}\sum_k (y^{(fixed, k)}_i -\mu_i^{(fixed)}) (y^{(fixed, k)}_j - \mu_j^{(fixed)})$. 

How can we generate typical additive noise on top of the mean-field output vector, given the covariance matrix? The answer is simple: we need to find a matrix $A^{(fixed)}$ that satisfies $A^{(fixed)} {A^{(fixed)}}^\T = \Sigma^{(fixed)}$. Once we have the $A$ matrix, a sample of the noise $n\in \mathrm{R}^{N_{\text{out}}}$, which has the correct distribution, can be obtained from a matrix multiplication between $A^{(fixed)}$ and a vector of i.i.d normal noise $z_i \sim N(0, \sigma^2 = 1)$, i.e. $n = A^{(fixed)} z$. 

So far, the dimensions of $A$ has been left unspecified. In principle, if the noise is sufficiently simple (e.g., the covariance matrix is rank 1), the covariance matrix can be decomposed as an outer product between a single vector and itself, i.e. $A^{(fixed)} \in \mathrm{R}^{N_\text{out}\times 1}$. We can generalize this concept of compressing the representation of the noise via the following procedure. First perform the spectral decomposition on $\Sigma^{(fixed)}$ to arrive at $U D U^\T$. Then, by looking at the eigenvalue spectra, ascertain how many dominant eigenvalues are present in the covariance matrix and denote that number as $N_\lambda$. Then let $A^{(fixed)} = U \sqrt{D}[:, 1:N_\lambda]$, to arrive at a compressed $A^{(fixed)}$ matrix with dimension $N_\text{out}\times N_\lambda$. For training our noise digital models, we find it is helpful to limit the number of eigenvectors of the covariance matrix, so that the dimension of the input-output map that must be learned is as small as possible (and because higher-order eigenvectors are not usually important).

To summarize, for a given input vector $x^{(fixed)}$, we only need to learn $A^{(fixed)}\in \mathrm{R}^{N_\text{out}\times N_\lambda}$ to output samples of the noise. Thus, to learn a noise digital model for general inputs, we simply train neural network to, given many different inputs $x^{(q)}$, predict the corresponding $A^{(q)}$ for any given $x^{(q)}$. This corresponds to a neural network that outputs a vector corresponding to a flattened $A$ (a vector of dimension $N_\text{out} N_\lambda$). Training data for this network is obtained by measuring the output of the physical system many times (we found 40 repeats was usually sufficient) for a nominally-fixed $x^{(q)}$ in order to compute the corresponding $A^{(q)}$ (repeating the above procedure for each training example). When used in the forward pass of the digital model for training in simulation, the noise model NN's predicted vector $A$ is reshaped to the matrix form and then used to generate a noise sample that is added to the mean-field digital model's prediction. 

\begin{figure}[t]
    \centering
    \includegraphics[width=\linewidth]{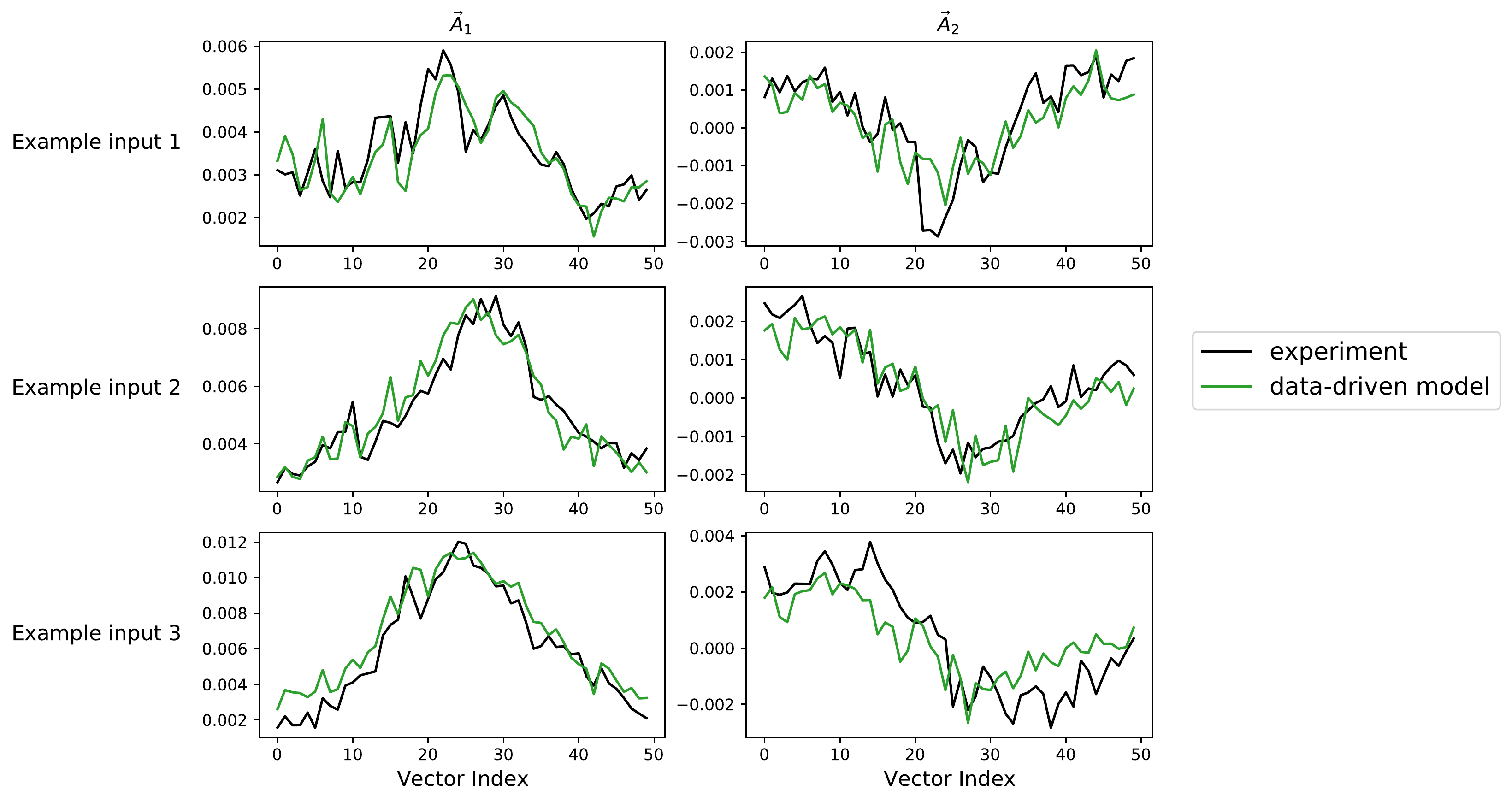}
    \caption{Characterization of the noise digital model. Here, the digital model (in black) predicts the 2 dominant noise eigenvectors $\vec A_1$ and $\vec A_2$ for three different 50-dimensional physical inputs to the SHG device. The experimental noise eigenvectors (in green) are derived from diagonalizing the sampled covariance matrix from 30 repeated measurements. Each row represents a distinct physical input.}
    \label{fig:noise_A}
\end{figure}

\begin{figure}[b]
    \centering
    \includegraphics[width=\linewidth]{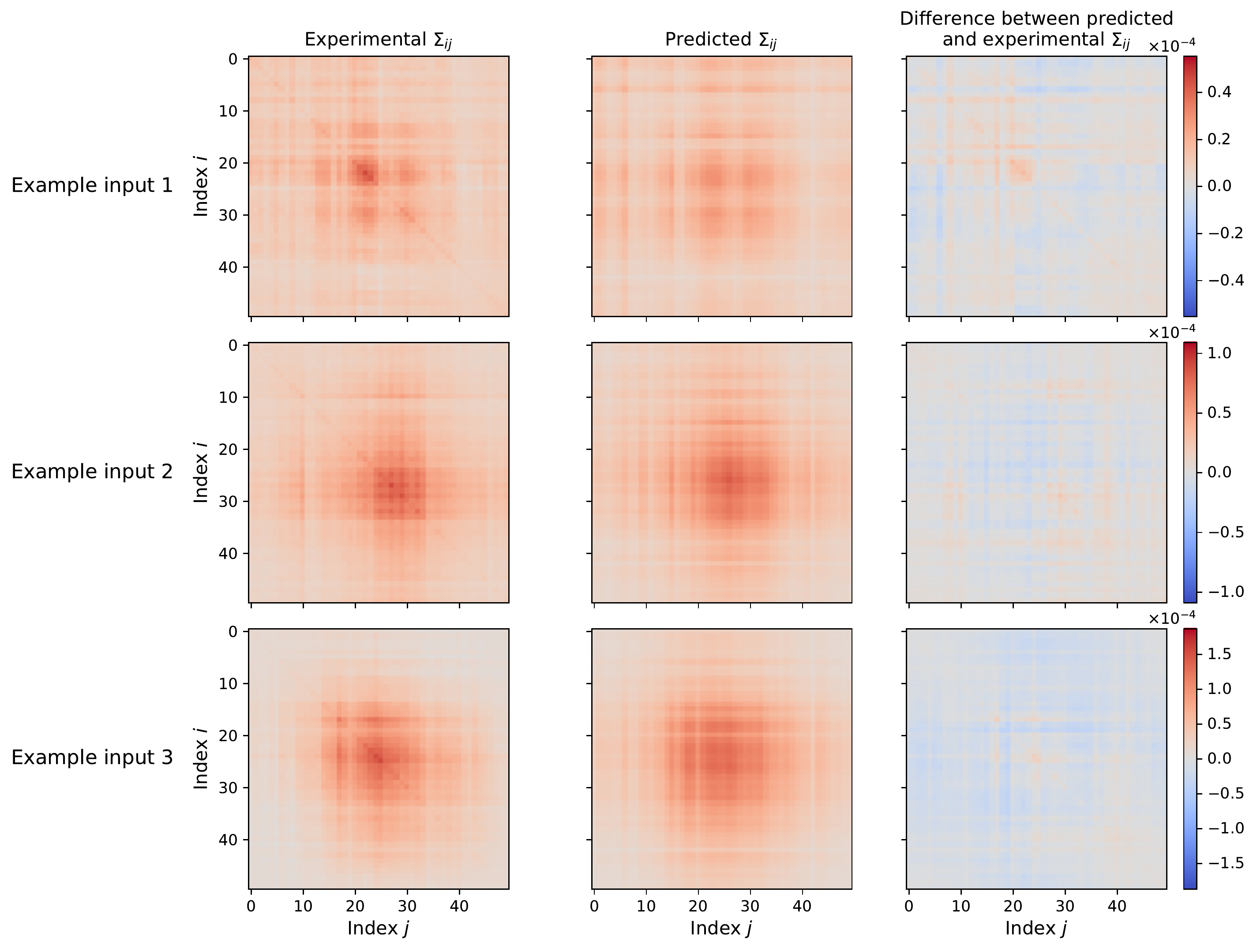}
    \caption{Characterization of the noise digital model. Left column: The experimental sampled covariance matrix from 30 repeated measurements. Middle column: The predicted covariance matrix computed with the predicted dominant noise eigenvectors $\vec A_1$ and $\vec A_2$ that are shown in Fig.~\ref{fig:noise_A}. Right column: The difference between the predicted and experimental covariance matrix. Each row represents a distinct physical input, and the covariance matrix for a given row shares a color bar that is shown on the right. The relatively low amplitudes visible in the right column show the strong agreement between the digital model's predicted noise distribution and the experimentally-measured distribution. 
    }
    \label{fig:shg_dt_noise}
\end{figure}

To provide examples of typical noise digital models, Supplementary Figures \ref{fig:noise_A} and \ref{fig:shg_dt_noise} compare, for the validation set, the performance of the noise digital model's prediction of the covariance eigenvectors and the covariance matrix to the experimental measurements of these quantities. The agreement is somewhat more approximate than what we obtain for the mean-field digital models (shown in the next subsections), but is overall excellent. 

\clearpage

\subsection{Descriptions of PNN architectures}

In this section, we describe the PNN architectures used for the different machine learning tasks implemented with each physical system. In each subsection, we also show the performance of the differentiable digital model used, and (for the PNNs not thoroughly examined in the main paper) present results illustrating how the physical transformations perform the classification task.

Before mentioning specifics for each architecture, we mention several commonalities among all PNNs we present in this work. 

First, the same basic training procedure was used in all cases. We found that the Adadelta optimizer \cite{zeiler2012adadelta} worked best, and speculate that its exclusion of momentum and parameter-dependent adaptation makes it well-suited to PNNs, whose parameters are generally not symmetric. We also usually utilized learning rate scheduling, but this was of secondary importance. 

Second, in all cases we compared trained PNNs to baseline reference models in which all physical input-output transformations were replaced by identity operations. This allowed us to confirm that, despite each PNN architecture including some digital operations, the digital operations alone were not responsible for the majority of the functionality of the trained PNN model. 

Third, for convenience, we sometimes performed some digital pooling operations to reduce the dimension of the measured physical output vectors. This is because the physical input-output transformations we considered (for convenience) had all equal input and output dimensions, and this operation allowed us to modify the dimensionality of the input-output transformations, allowing us to tackle machine learning problems with different dimensionality in the input (feature) vector and output vector (for instance, the MNIST dataset has a 196 input dimension and 10 dimensional output). Since this operation is not trained and it can only contribute a minor computational power to the PNN. Besides, any minor benefits it confers can easily be performed physically as it is equivalent to changing the sampling rate of output measurements (e.g., by using a slower digital-analog converter). Finally, we note that for the oscillating plate and electronic PNNs, downsampling operations of this kind were ultimately irrelevant, as both devices were operated at input and output sampling rates limited by the sound card and data acquisition device respectively.

Finally, in all cases it was necessary to include additional Lagrangian terms in the loss in order to ensure that physical inputs and parameters sent to physical systems were within the allowed ranges. This additional terms are identical to the Lagrangian terms that have been discussed earlier in Section \ref{sec:pat-num}.

\subsubsection{Figure 2: Femtosecond SHG-vowels PNN}
The femtosecond SHG-vowel PNN was chosen to be the first example PNN of the main manuscript for two reasons. First, the vowel classification task is simple enough to facilitate a relatively simple PNN, and second, the SHG physical transformation is relatively unusual, being a fully-nonlinear transformation with nearly all-to-all coupling of inputs. Although the second feature does not make SHG well-suited to the vowel-classification task (which is linear), we felt it served as a clear, small-scale tutorial demonstration of a multilayer PNN on a relatively high-dimensional task in which the physical transformation used is highly nonlinear. 

To make the SHG-vowel PNN intuitive and illustrative for the concept of PNNs in general, we made several design choices. We used a portion of the pulse shaper's spectral modulations for trainable parameters, and a second, distinct portion for encoding input data. In total, each SHG input-output transformation maps 100 physical inputs, which are split between the input data and trainable parameters, to a 50-dimensional output SHG spectrum. 

\begin{figure}[h!]
    \centering
    \includegraphics[width=0.9\linewidth]{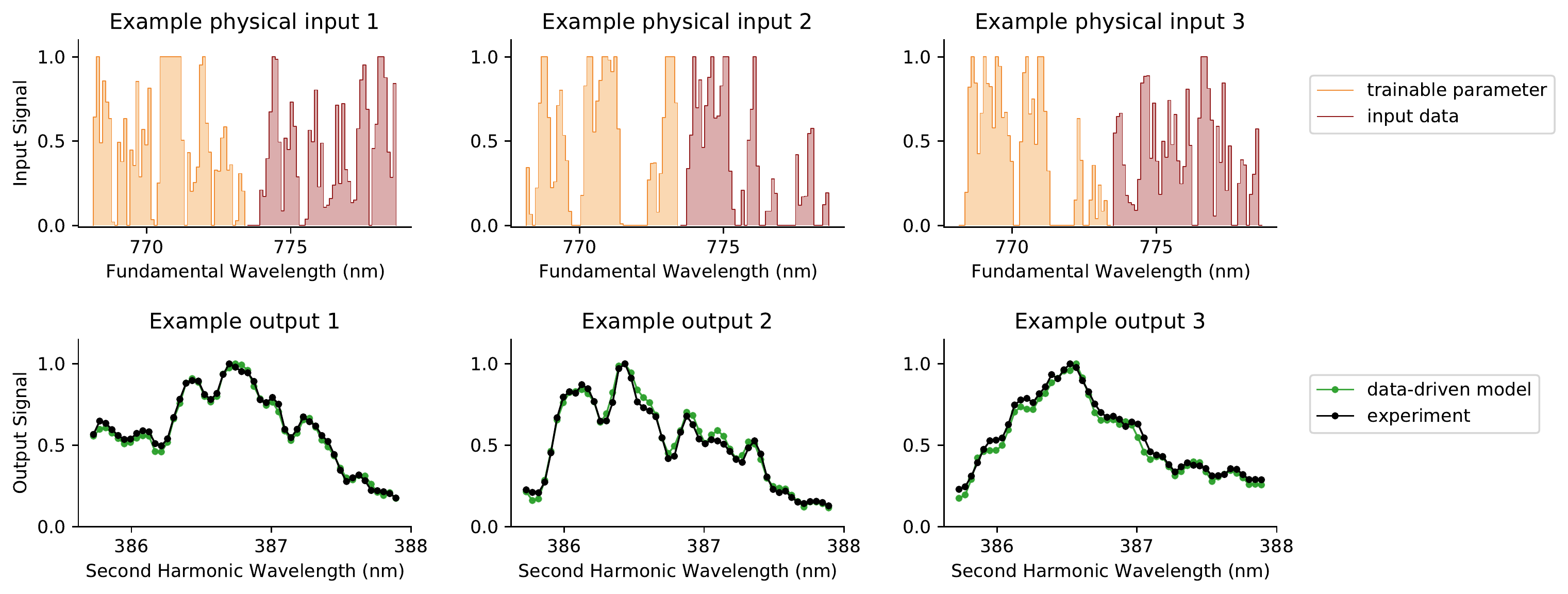}
    \caption{Inputs and outputs of the physical transformation performed by the femtosecond SHG device. Here, 100 dimensional physical inputs applied to the pulse shaper are equally split between the input (machine learning data) and parameters. The outputs of the SHG is 50 dimensional. As shown by the good agreement between the black and green curves, the data-driven models emulate the physical system well. Note that the physical inputs considered here is part of the ``test'' set and has not been used to train the data-driven model.}
    \label{fig:shg_dt}
\end{figure}

In Fig.~\ref{fig:shg_dt}, we show some representative input and outputs of this transformation, where physical inputs are split equally between the input and parameters (50 dimensional input and parameters each). Consistent with the analysis performed in Sec.~\ref{sec:iot-shg}, the transformation is complex, as evidenced by the widely-varying outputs. Nevertheless, this transformation can be effectively modeled via a data-driven model. By performing the neural architecture prescribed in Sec.~\ref{sec:data-driven-models}, we found a deep neural network that is able to learn the transformation effectively. This digital model's excellent agreement with the experimental input-output transformation is shown in Fig. S18.

Having characterized the performance of the differentiable digital model, we now show how the SHG system performs vowel classification. To begin, we first describe the vowel classification task that we perform. We use the vowel dataset from Ref.~\cite{Torrejon2017GrollierspintronicNN}, which can be downloaded at this \url{https://static-content.springer.com/esm/art\%3A10.1038\%2Fs41586-018-0632-y/MediaObjects/41586_2018_632_MOESM2_ESM.docx}, with an additional (untrainable) renormalization to account for the fact that the SHG system takes in physical inputs between 0 and 1. For completeness, we will paraphrase the description of the dataset that is in the supplementary information of Ref.~\cite{Torrejon2017GrollierspintronicNN}, before describing the renormalization we apply to the dataset. 

The vowel data considered is a subset of the Hillenbrand vowel database, which can be found at \url{https://homepages.wmich.edu/~hillenbr/voweldata.html}. In this dataset, each spoken vowel is characterized by their formant frequencies, which refers to the dominant frequencies (in units of $\si{Hz}$) present in an audio signal. More specifically, each vowel is characterized by a 12-dimensional vector, which contains the first three formants sampled at the steady state (whole audio signal) and three temporal sections, centered at 20\%, 50\% and 80\% of the total vowel duration respectively. The subset of vowels we consider are the vowels spoken by 37 female speakers which have complete formant information (some vowels in the full dataset have formants that could not be measured). Since each speaker utters 7 different vowels, the vowel dataset contains a total of 259 vowels ($N_\text{data} = 257$). 

In order to accommodate the finite input range of the SHG experiment, we normalize them via the element-wise min-max feature scaling, where each dimension of the input vector is renormalized by their corresponding maximum and minimum values (over all 259 vowels). Mathematically, if the original dataset is given by $\set{(\bm{x}^{(k)} , \bm{y}^{(k)})}_{k=1}^{N_\text{data}}$, the examples of the transformed dataset $\set{({\bm{x}'}^{(k)} , \bm{y}^{(k)})}_{k=1}^{N_\text{data}}$ is given by $ {x'}^{(k)}_i = \paren{x^{(k)}_i - x_{\text{min}, i}}/\paren{x_{\text{max}, i} - x_{\text{min}, i}}$, where the element-wise minimum and maximum values are given by $x_{\text{min}, i } = \min\limits_k x^{(k)}_i$ and $x_{\text{max}, i } = \max\limits_k x^{(k)}_i$. 

\begin{figure}[h!]
    \centering
    \includegraphics[width=\linewidth]{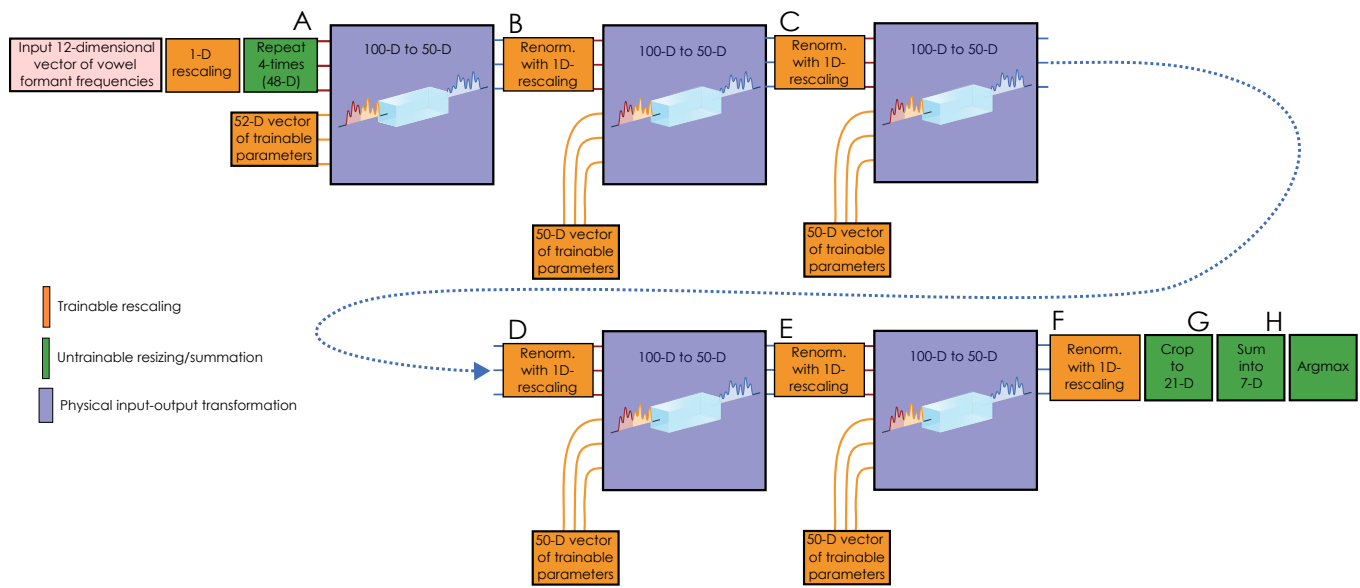}
    \caption{The full PNN architecture used for the second harmonic generation vowel classifier (for the case of 5-layers). Annotation letters (A-H) refer to the plots in Fig \ref{fig:vowel_dataflow}. }
    \label{fig:vowel architecture}
\end{figure}

\begin{figure}[h!]
    \centering
    \includegraphics[width=\linewidth]{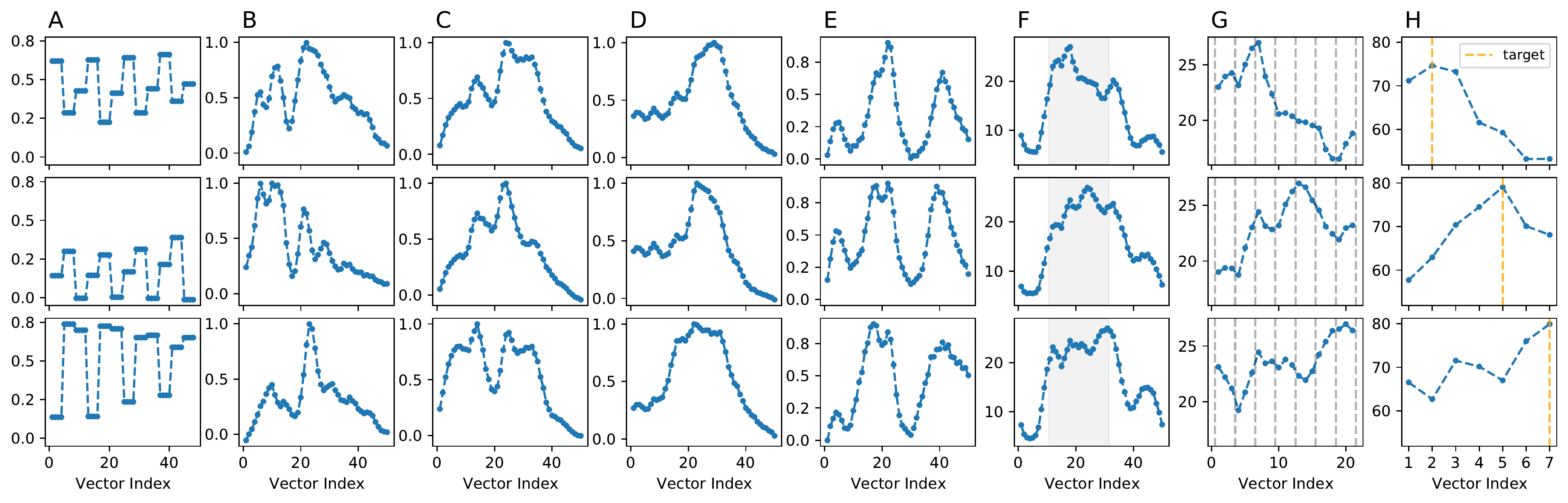}
    \caption{The sequence of operations that make up the trained SHG PNN vowel model. The lettered labels corresponds to intermediate quantities that are labelled in Fig.~\ref{fig:vowel architecture}.
    }
    \label{fig:vowel_dataflow}
\end{figure}

The full architecture for performing the vowel classification with the SHG system is schematically shown in Fig.~\ref{fig:vowel architecture}. In the following, we outline the steps involved in performing an inference with the PNN. To show how the intermediate quantities evolve in the PNN, we plot quantities that are labelled by letter labels in Fig.~\ref{fig:vowel architecture} in subpanels of Fig.~\ref{fig:vowel_dataflow}. 
\begin{enumerate}
    \item The 12-D normalized vowel formant frequencies are overall rescaled, i.e. $x_i \rightarrow ax_i + b$, where $a$ and $b$ are trainable parameters. 
    \item The 12-D vector is then repeated 4 times. The 48-D output is the input to the first SHG device, i.e. $\bm{x}^{[1]} = [x_1, x_1, x_1, x_1, \dots, x_{12}, x_{12}, x_{12}, x_{12}]^\T$ to arrive at a 48-D input for the first SHG system. This 48-D input is shown in A of Fig.~\ref{fig:vowel architecture} and Fig.~\ref{fig:vowel_dataflow}.
    \item For the first SHG system, this input is concatenated with a 52-D parameter and the total 100-D physical input $\bm q^{[1]} = [\bm x^{[1]}, \bm \theta^{[1]}]^\T$ is sent to the first SHG.
    \item The SHG performs the physical computation $\bm s^{[1]} = f_\text{SHG}(\bm q^{[1]})$. The output of the first SHG $\bm s^{[1]}$ is shown in B of Fig.~\ref{fig:vowel architecture} and Fig.~\ref{fig:vowel_dataflow}. 
    \item The 50-D output of the first SHG $\bm s^{[1]}$ is renormalized and an overall trainable rescaling is applied, i.e. $\bm y^{[1]} = a^{[1]} \bm s^{[1]} / \max(\bm s^{[1]}) + b^{[1]}$, where $a^{[1]}$ and $b^{[1]}$ are trainable scalars. 
    \item The 50-D output is then concatenated with a 50-D parameter and sent to the next SHG device. 
    \item Step 4 and 5 are repeated until the output of the fifth SHG device has been renormalized and rescaled. Mathematically, they are given by, 
    \begin{align}
        \bm x^{[l]} & =  y^{[l-1]}
    \\  \bm q^{[l]} & = [\bm x^{[l]}, \bm \theta^{[l]}]^\T
    \\  \bm s^{[l]} & = f_\text{SHG}(\bm q^{[l]})
    \\  \bm y^{[l]} & = a^{[l]} \bm s^{[1]} / \max(\bm s^{[l]}) + b^{[l]}
    \end{align}
    for $l=2, 3, \ldots 5$. The output of each SHG device is shown in C-E of Fig.~\ref{fig:vowel architecture} and Fig.~\ref{fig:vowel_dataflow}.
    \item From the 50-D output, a 21-D section is cropped as follows $\bm y^{[\text{c}]} = [y^{[5]}_{12}, y^{[5]}_{13}, \ldots y^{[5]}_{32}]^\T$. The 21-D output is shown in F of Fig.~\ref{fig:vowel architecture} and Fig.~\ref{fig:vowel_dataflow}.
    \item Each 3-D section of this new 21-D output is summed to arrive at the final output of the PNN, i.e. 
    $\bm o = [y_1^{[\text{c}]}+y_2^{[\text{c}]}+y_3^{[\text{c}]}, y_4^{[\text{c}]}+y_5^{[\text{c}]}+y_6^{[\text{c}]}, \ldots y_{19}^{[\text{c}]} + y_{20}^{[\text{c}]} + y_{21}^{[\text{c}]}]^\T$. This final output is shown in G of Fig.~\ref{fig:vowel architecture} and Fig.~\ref{fig:vowel_dataflow}.
\end{enumerate}

Analogous to Sec.~\ref{sec:pat-num}, the PNN is trained by minimizing the following loss function:
\begin{align}
    L = H(\bm y, \text{Softmax}(\bm o)) + \lambda_\mathrm{\theta}\sum_l L_{\text{bound}}(\bm \theta^{[l]}, v_{\text{min}},  v_{\text{max}}) +  \lambda_\mathrm{x}\sum_l L_{\text{bound}}(\bm x^{[l]},  v_{\text{min}}, v_{\text{max}})
\end{align}
where $H$ is the cross-entropy function and $L_{\text{bound}}(\bm z, v_{\text{min}}, v_{\text{max}})= \sum_l \frac{1}{N_{\bm z}}\sum_i^{N_{\bm z}} \max(0, z_i - v_{\text{max}}) -\min(0, z_i-v_{\text{min}})$. The last two terms are regularization terms to constraint the inputs and parameters of each SHG device, whose parameters are $\lambda_\mathrm{\theta}=0.5, \lambda_\mathrm{x}=2, v_{\text{min}} = 0, v_{\text{max}}=1$. 

Instead of vanilla stochastic gradient descent (as in \eqref{eq:pat-ffa}), we perform PAT with the Adadelta optimizer. This is because the adaptive learning rate is helpful for dealing with the different nature of parameters that exist in the model. The initial learning rate is $1.0$ and it was reduced by half every 700 epochs. 

\FloatBarrier
\subsubsection{Figure 4: Analog transistor-MNIST PNN}

As described in the main article, the electrical circuit chosen to serve as an electronic PNN demonstration was selected to exhibit highly nonlinear transient dynamics by embedding a transistor within an RLC oscillator. 

This circuit was, unexpectedly, particularly challenging to produce an accurate differentiable digital model for. While we expect that incorporating knowledge-based components or other physical insight could improve the results, in order to keep our procedure general, we found that an effective workaround was to simply reduce the dimensionality of the input-output transformation. Empirically, we found that 70-dimensions (70 input samples, and 70 output samples) was small enough to achieve reasonable results. 

The digital model was found by performing a neural architecture search. The best model found obtained a validation loss of 0.09 (with data normalized to -1 to 1). It is a 4-layer neural network with swish activation functions, and 217, 427, 167, and 197 hidden units. The Adam optimizer was used with a learning rate of 4.56$\times 10^{-4}$.  

A related challenge with the analog electrical PNN was that our implementation exhibits high noise, mainly due to lost samples during streaming over the USB connection, and to jitter related to triggering at the limit of the device's temporal resolution. Even after implementing post-selection to account for traces corrupted or lost during the fast read-write operation of the DAQ, the mean deviation amplitude between the output voltage time series for two identical inputs was over 5\% of the maximum absolute output amplitude, and nearly 20\% of the absolute mean output amplitude. 

\begin{figure}[h!]
    \centering
    \includegraphics[width=\linewidth]{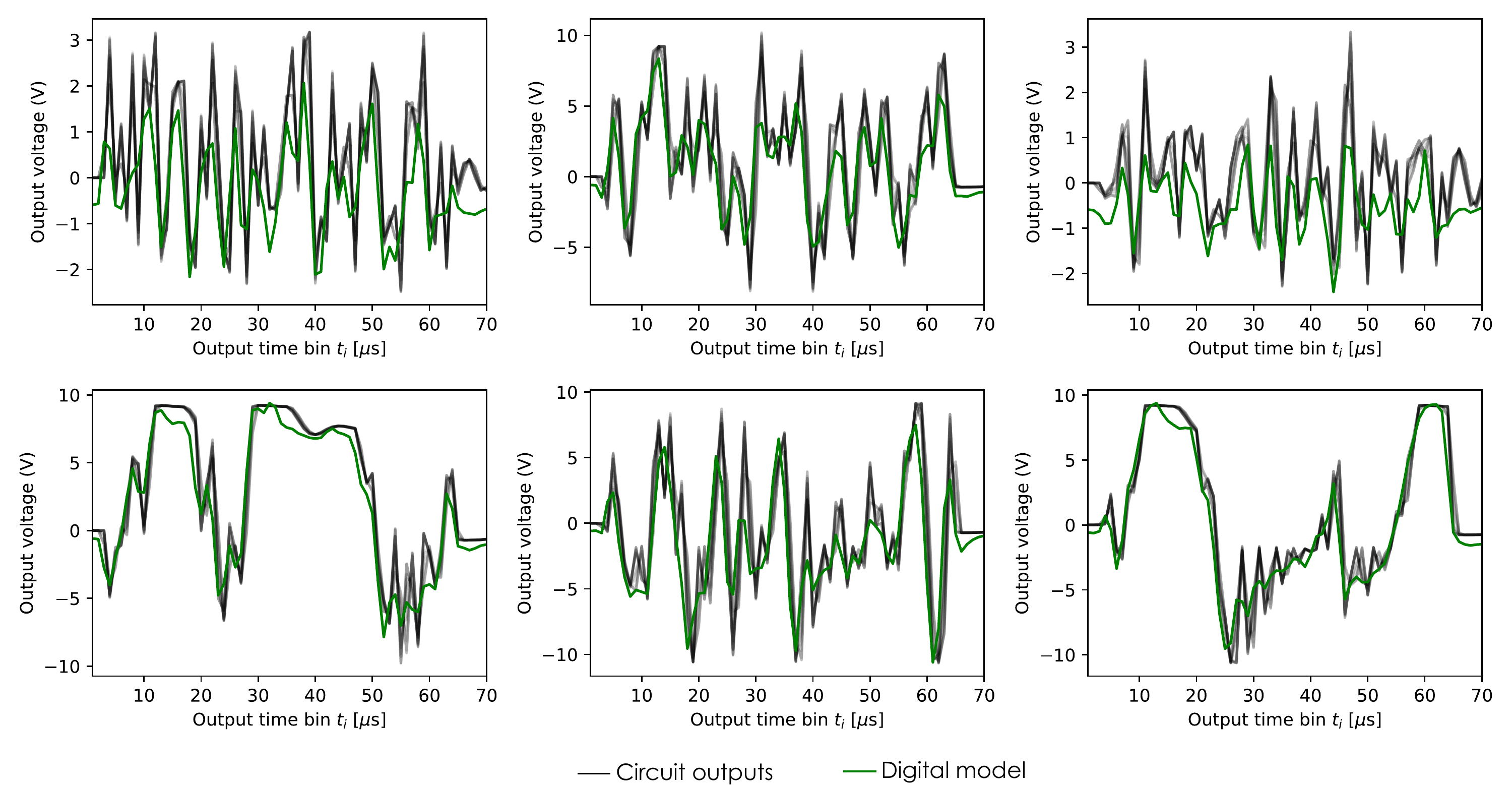}
    \caption{The full PNN architecture used for the analog transistor circuit MNIST digit classifier. For details see main text.}
    \label{fig:ePNN digital twin}
\end{figure}

Supplementary Figure \ref{fig:ePNN digital twin} shows the differentiable digital model's predictions compared to the circuit's actual output. As a reference point, the figure shows 40 output traces from the circuit, each taken  for the same inputs. The figures illustrates that a main limiting factor in the digital model's accuracy is in fact the large, complex noise in the circuit's input-output transformation. 

These two considerations, the low dimensionality of the input-output transformation, and the significant noise of the circuit's operation, influenced our PNN architecture design, shown in Supplementary Figure \ref{fig:ePNN architecture}. 

To make the 196-D MNIST images compatible with the 70-D input-output transformations of the electronic circuit, we chose to sum blocks of the MNIST image into 14-dimensional vectors (i.e., perform a pooling operation on the 196-dimensional vector in blocks of 14), then repeat this 14-dimensional vector 5 times as the input to the first layer of the PNN architecture. Of course, many other options are possible, but we found this strategy to be effective. 

\begin{figure}[h!]
    \centering
    \includegraphics[width=\linewidth]{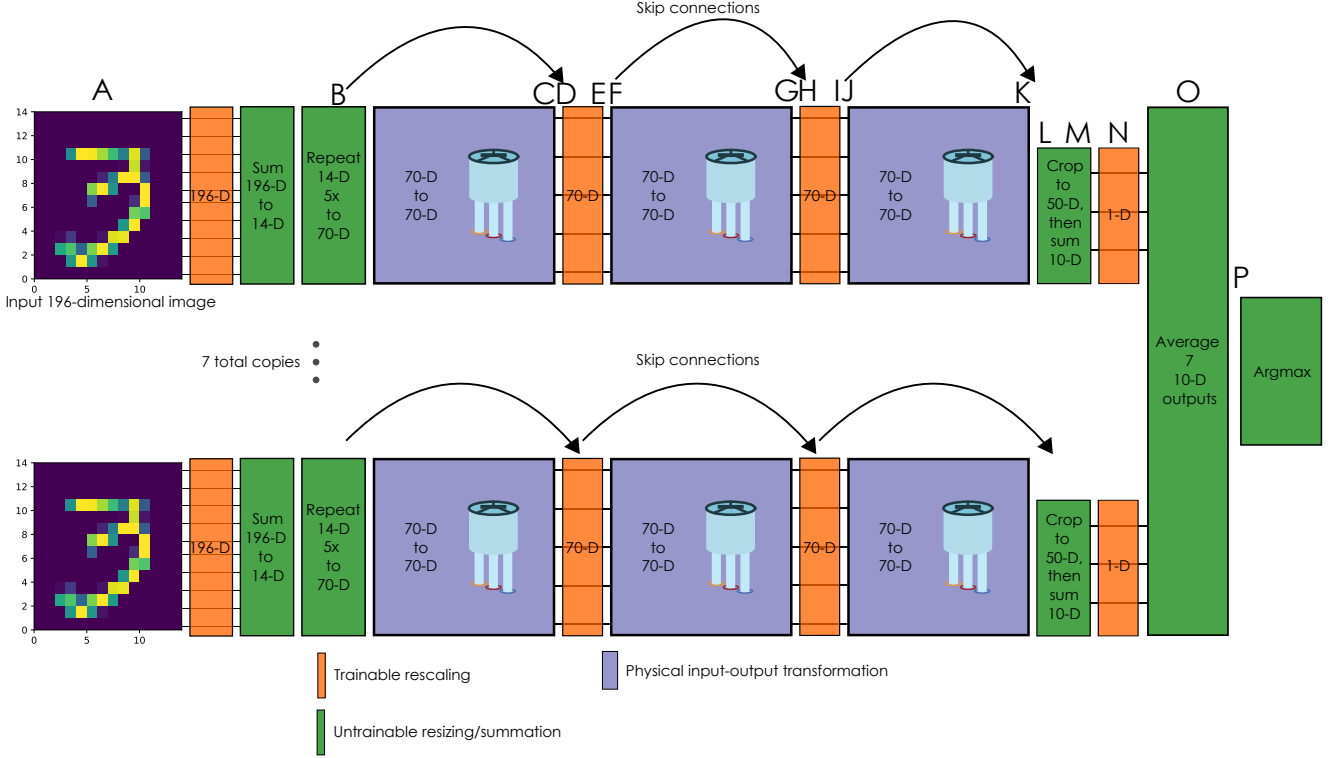}
    \caption{The full PNN architecture used for the analog transistor circuit MNIST digit classifier. Annotation letters (A,B,C,D,...) refer to the plots in Fig \ref{fig:ePNN MNIST} and the description in the main text.}
    \label{fig:ePNN architecture}
\end{figure}

\begin{figure}[h!]
    \centering
    \includegraphics[width=\linewidth]{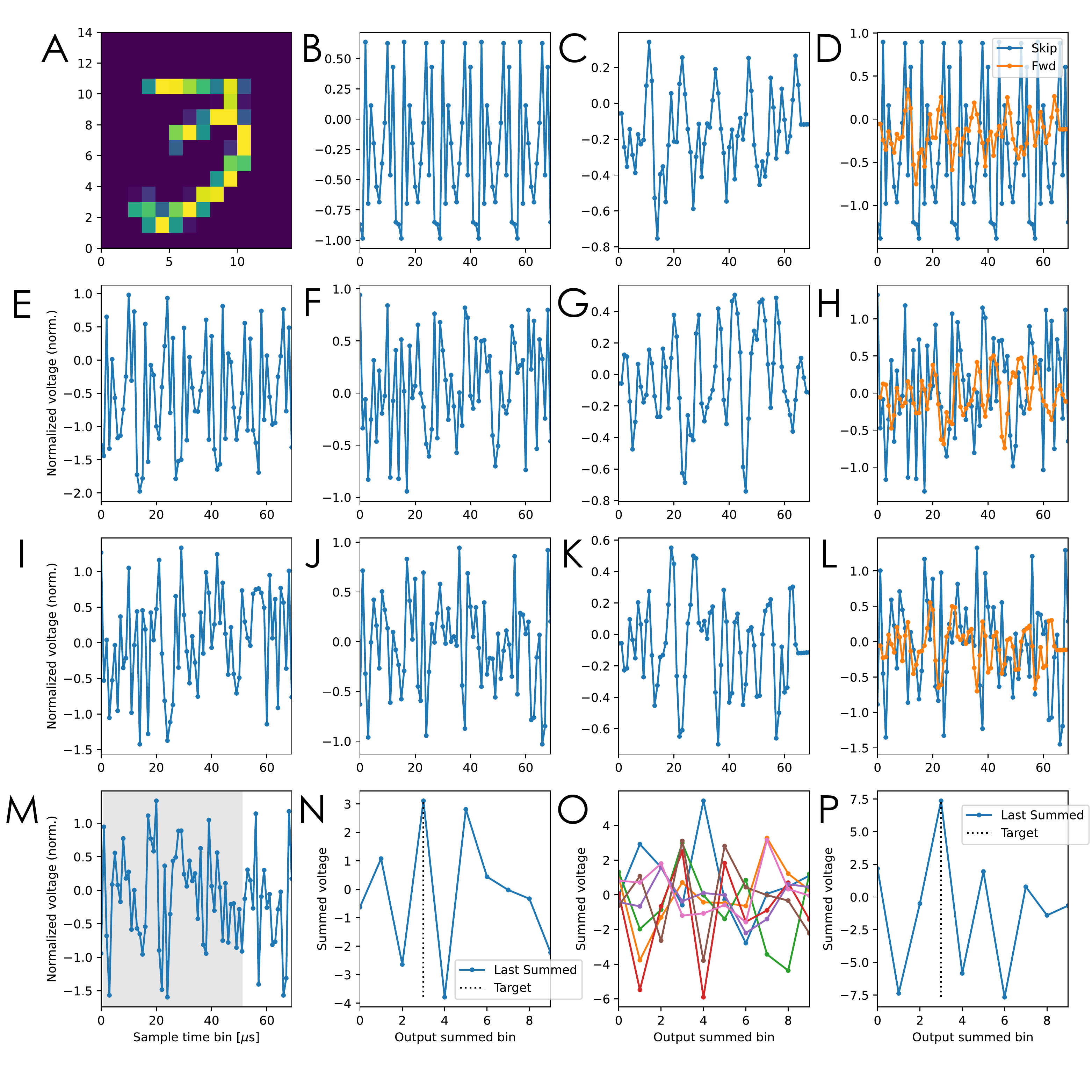}
    \caption{The sequence of operations that make up the trained electronic PNN MNIST model. For details of each section, see main text. In part O, the different-colored curves correspond to the output from the 7 different 3-layer PNNs.}
    \label{fig:ePNN MNIST}
\end{figure}

To obtain robust performance despite the circuit's noise and the digital model's relatively poor accuracy, we chose to implement two additional design components. First, we operate the PNN as a `committee': 7, 3-layer PNNs each produce a prediction, and the full PNN produces its prediction by taking a weighted average of these. Second, we added trainable skip connections between layers of the PNN. These connections, inspired by residual neural networks, were added to allow the network to act further as an ensemble of sub-networks \cite{veit2016residual}, thus adding to the redundancy of the architectures (and, in principle, helping it to achieve good classification performance even when any one sub-part stochastically fails) \cite{veit2016residual}. 

Each inference of the full architecture involves the following steps. By letter labels (A,B,C, ...), we refer each step to its corresponding location in Supplementary Figure \ref{fig:ePNN architecture} and to the subpanels of Supplementary Figure \ref{fig:ePNN MNIST}. 
\begin{enumerate}
    \item 196-D MNIST images (A in Supplementary Figures \ref{fig:ePNN architecture} and \ref{fig:ePNN MNIST}, downsampled from 784-D) are element-wise rescaled, i.e. $x_i \rightarrow x_ia_i+b_i$, where $a_i$ and $b_i$ are trainable parameters  .
    \item Pool the 196-D vector from 196 to 14-D (i.e., sum each region of 14).
    \item Concatenate this 14-D vector 5 times to produce a 70-D input for the first physical transformation (B in Supplementary Figures \ref{fig:ePNN architecture} and \ref{fig:ePNN MNIST})
    \item Run the physical transformation once, producing an output 70-D vector $\vec{y}$  (B-C in Supplementary Figures \ref{fig:ePNN architecture} and \ref{fig:ePNN MNIST}).
    \item Add the output $\vec{y}$ to the initial input via trained skip connection, i.e. $\vec{y} \rightarrow \vec{y} + a \vec{x}$, where $a$ is a trainable skip weight and $\vec{x}$ was the input to the physical system (D-E in Supplementary Figures \ref{fig:ePNN architecture} and \ref{fig:ePNN MNIST}). 
    \item Apply trainable, element-wise rescaling to produce the input for the second physical transformation, i.e., $x_i = y_ia_i + b_i$ (E-F in Supplementary Figures \ref{fig:ePNN architecture} and \ref{fig:ePNN MNIST}).
    \item Run the physical transformation once, producing an output 70-D vector $\vec{y}$ (F-G in Supplementary Figures \ref{fig:ePNN architecture} and \ref{fig:ePNN MNIST}).
    \item Add the output $\vec{y}$ to the initial input via a trained skip connection, i.e. $\vec{y} \rightarrow \vec{y} + a \vec{x}$, where $a$ is a trainable skip weight and $\vec{x}$ was the input to the physical system (H-I in Supplementary Figures \ref{fig:ePNN architecture} and \ref{fig:ePNN MNIST}). 
    \item Apply trainable, element-wise rescaling to produce the input for the second physical transformation, i.e., $x_i = y_ia_i + b_i$ (I-J in Supplementary Figures \ref{fig:ePNN architecture} and \ref{fig:ePNN MNIST}).
    \item Run the physical transformation once, producing an output 70-D vector $\vec{y}$ (J-K) in Supplementary Figures \ref{fig:ePNN architecture} and \ref{fig:ePNN MNIST}).   \item Add the output $\vec{y}$ to the initial input via trained skip connection, i.e. $\vec{y} \rightarrow \vec{y} + a \vec{x}$, where $a$ is a trainable skip weight and $\vec{x}$ was the input to the physical system (L-M in Supplementary Figures \ref{fig:ePNN architecture} and \ref{fig:ePNN MNIST}). 
    \item Sum 50 elements of $\vec{y}$ in sections of 5 to produce a 10-dimensional vector (M-N in Supplementary Figures \ref{fig:ePNN architecture} and \ref{fig:ePNN MNIST}).
    \item Repeat the above steps for 6 other sub-PNNs, each with its own independent trainable parameters, producing its own independent output 10-dimensional vector (Outputs of each independent sub-PNN shown in O in Supplementary Figure \ref{fig:ePNN MNIST}).
    \item Take a weighted sum of the 7, 10-dimensional vectors, $\vec{y}_{\text{out}} = \sum_{i=1}^7 w_i \vec{y}_i + b_i$ (P in Supplementary Figures \ref{fig:ePNN architecture} and \ref{fig:ePNN MNIST}).
\end{enumerate}

To train the PNN, we used PAT with the Adadelta optimizer with learning rate of 0.75, which was reduced to 0.375 after 30 epochs, then to 0.28 after 45 epochs. We found Adadelta to be particularly effective compared to other optimizers, and hypothesize that momentum may inhibit the suppression of gradient errors in PAT, at least in these very small-scale proof-of-concept PNNs. Although we found it resulted in only moderate improvements in small-scale tests, we retrained the digital model for the electronic circuit every two training epochs, using a combination of random 70-dimensional inputs and local data (i.e., the inputs and outputs obtained from the current training data). This was done in part to help compensate for variations in the circuit's response over time, and also to nominally help improve the digital model's accuracy within the local region of parameter space.

\subsubsection{Figure 4: Oscillating plate MNIST PNN}

\begin{figure}[h!]
    \centering
    \includegraphics[width=\linewidth]{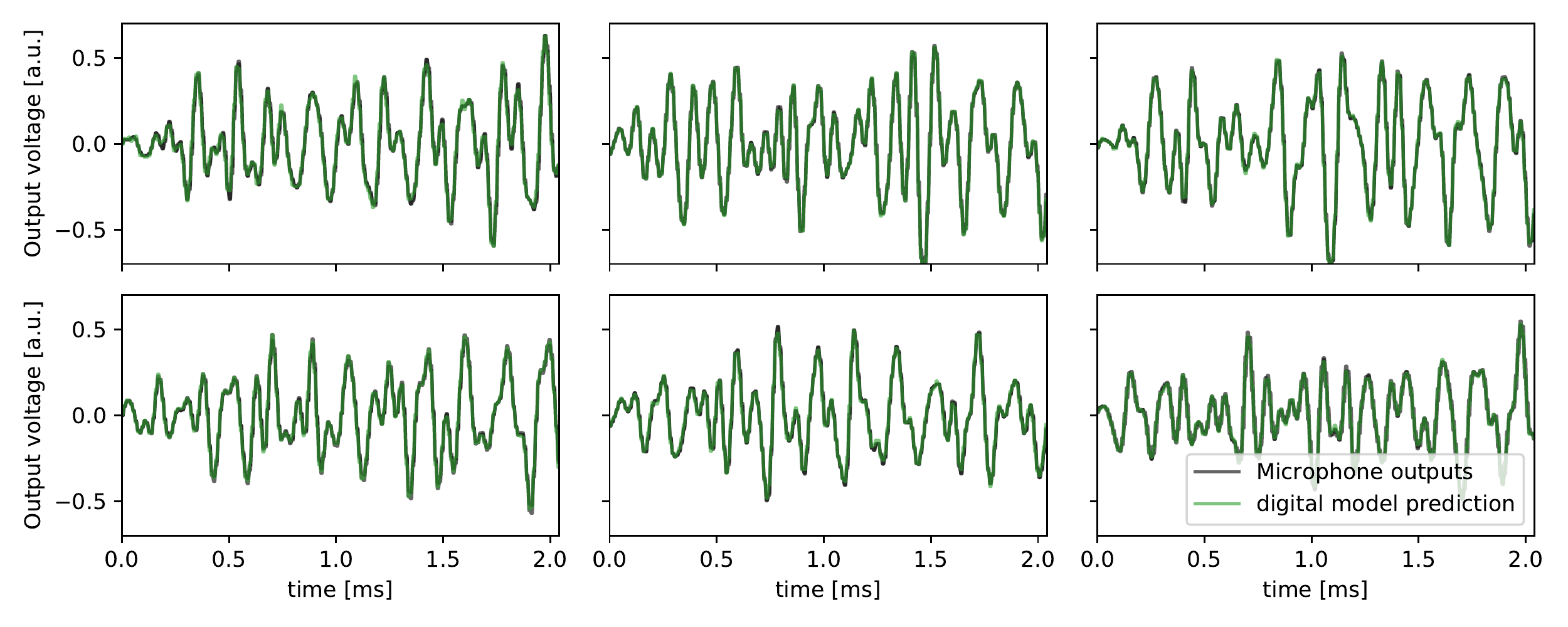}
    \caption{Multiple microphone recordings for 196 uniformly distributed input voltages at an input frequency of $192$kHz and the digital models' predictions. The linearity of the input-output transformation allowed excellent agreement between digital model and experiment as evident from the almost indistinguishable traces.}
    \label{fig:OPmodelvsexp}
\end{figure}

\begin{figure}[h!]
    \centering
    \includegraphics[width=\linewidth]{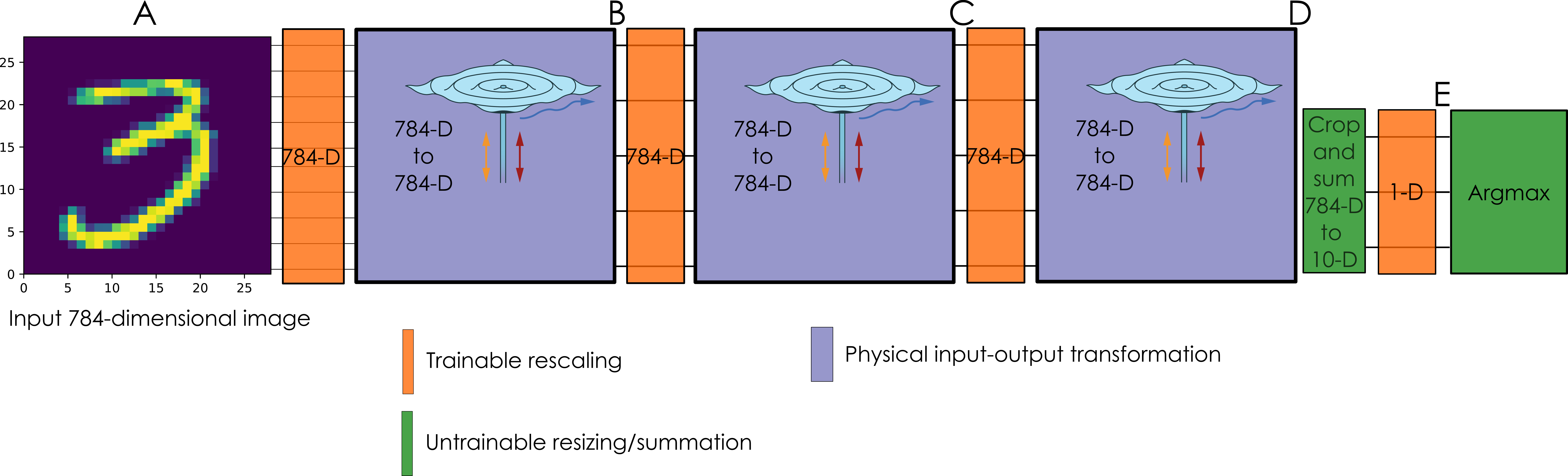}
    \caption{The full PNN architecture used for the oscillating plate MNIST digit classifier. Annotation letters (A,B,C,D,E) refer to the plots in Fig \ref{fig:OP MNIST} and the description in the main text.}
    \label{fig:OP architecture}
\end{figure}

\begin{figure}[h!]
    \centering
    \includegraphics[width=\linewidth]{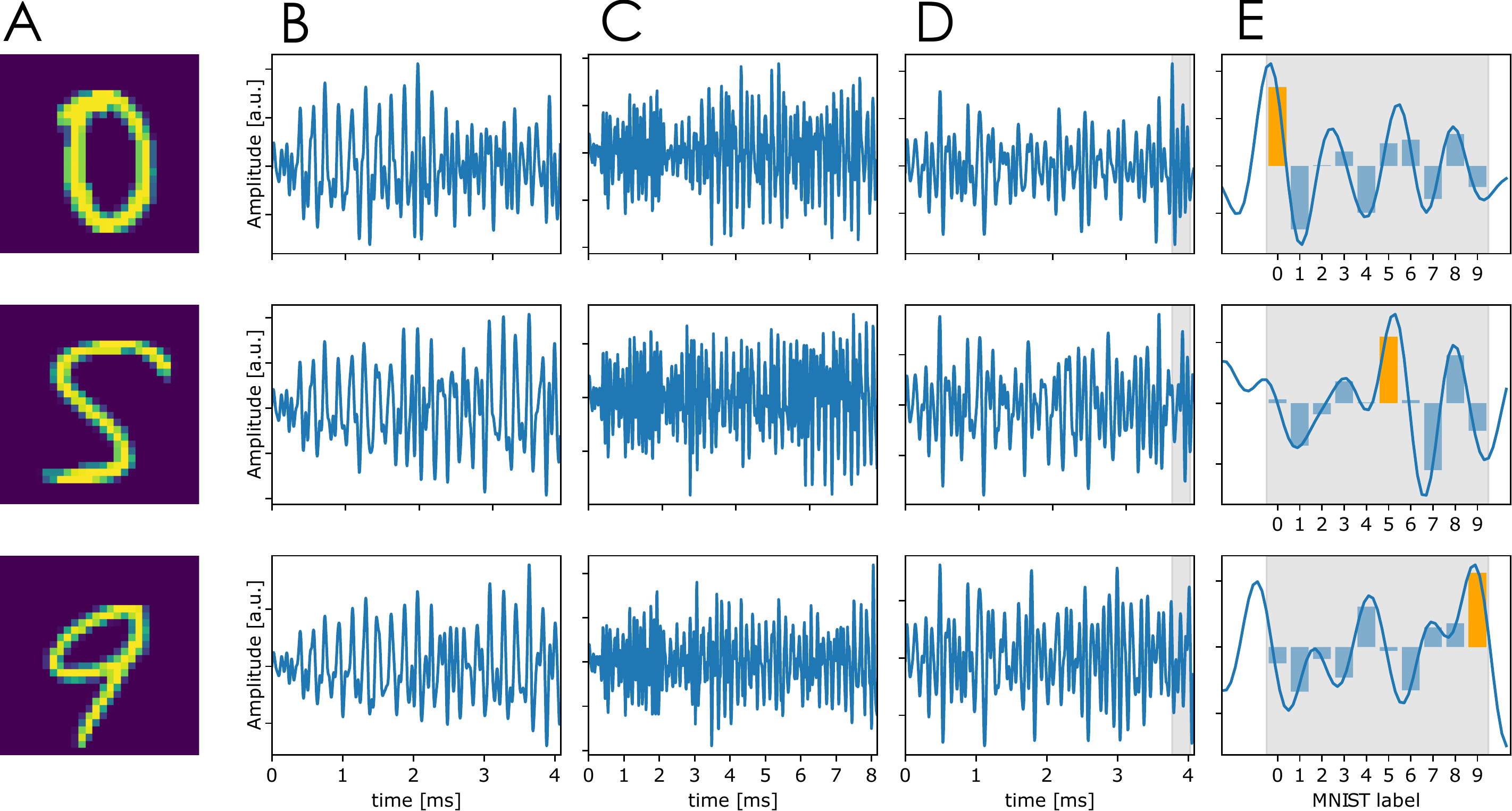}
    \caption{The sequence of operations that make up the oscillating plate PNN MNIST model. For details of each section, see main text.}
    \label{fig:OP MNIST}
\end{figure}

The linearity of the oscillating plate’s input-output transformation allowed us to accurately learn even very high-dimensional input-output transformations.
There was no need to downsample the original 784-D MNIST images as we could learn a 78-4D to 784-D dimensional input-output map. 
Instead, we could unroll and encode a whole image into a time-multiplexed voltage signal. 
Outputs of the same dimension were read out from the voltage over time signal of the microphone.

The strong performance of a linear single-layer perceptron on the MNIST dataset and the realization that the oscillating plate PNN performs a matrix multiplication as in Eq. \ref{eq:OPmatrix} inspired us to use the linear oscillating plate in a similar fashion. 
By cascading a few layers of element-wise-rescaling and passing the signal through the speaker we roughly emulate the operation of a single-layer perceptron.

The final layer of the oscillating plate consists of reading out a short time window from the microphone output voltage. The exact position and length of the time window was a trainable hyperparameter of our network architecture, however, its position was generally closer to the end of the output as information from the input can only interact with later inputs due to causality.
The window is divided into 10 bins. 
MNIST digits were classified according to which bin had the strongest average signal.

Each inference of the full architecture involves the following steps:
\begin{enumerate}
\item 784-D MNIST images are element-wise rescaled, i.e. $x_i \rightarrow x_i a_i+b_i$, where $a_i$ and $b_i$ are trainable parameters.
\item Run the physical transformation once, producing an output 784-D vector $\vec{y}$.
\item Apply trainable, element-wise rescaling to produce the input for the second 784-D physical transformation, i.e. $x_i \rightarrow x_i a_i+b_i$. 
\item Run the physical transformation once, producing an 784-D vector $\vec{y}$.
\item Apply trainable, element-wise rescaling to produce the input for the third 784-D physical transformation, i.e. $x_i \rightarrow x_i a_i+b_i$. 
\item Run the physical transformation once, producing an 784-D vector $\vec{y}$.
\item Take outputs 724 to 774 and average 5 consecutive points to produce a 10-D vector $\vec{y_\text{out}}$.
\end{enumerate}

Due to the simple PNN architecture and the very accurate linear digital twins, we could achieve high accuracy on MNIST by only using \emph{in silico} training and transferring the trained parameters onto the experiment. Using PAT still resulted in performance gains, although not as dramatic as for the nonlinear physical systems. Usually, \emph{in silico} performance in experiment only fell a few percentage points short of PAT, and after just a few epochs (~3) of transfer learning with PAT, the full performance could be retrieved. We expect therefore that, at least for very shallow networks, PAT is less crucial for linear physical systems, and we expect that \textit{in silico} training can be generally applied to greater effect in such systems. While this may be useful in some contexts, we note that sequences of linear systems cannot realize the computations performed by deep neural networks, since they ultimately amount to a single matrix-vector multiplication.

\subsubsection{Figure 4: Femtosecond SHG-MNIST PNN}

\begin{figure}[h!]
    \centering
    \includegraphics[width=\linewidth]{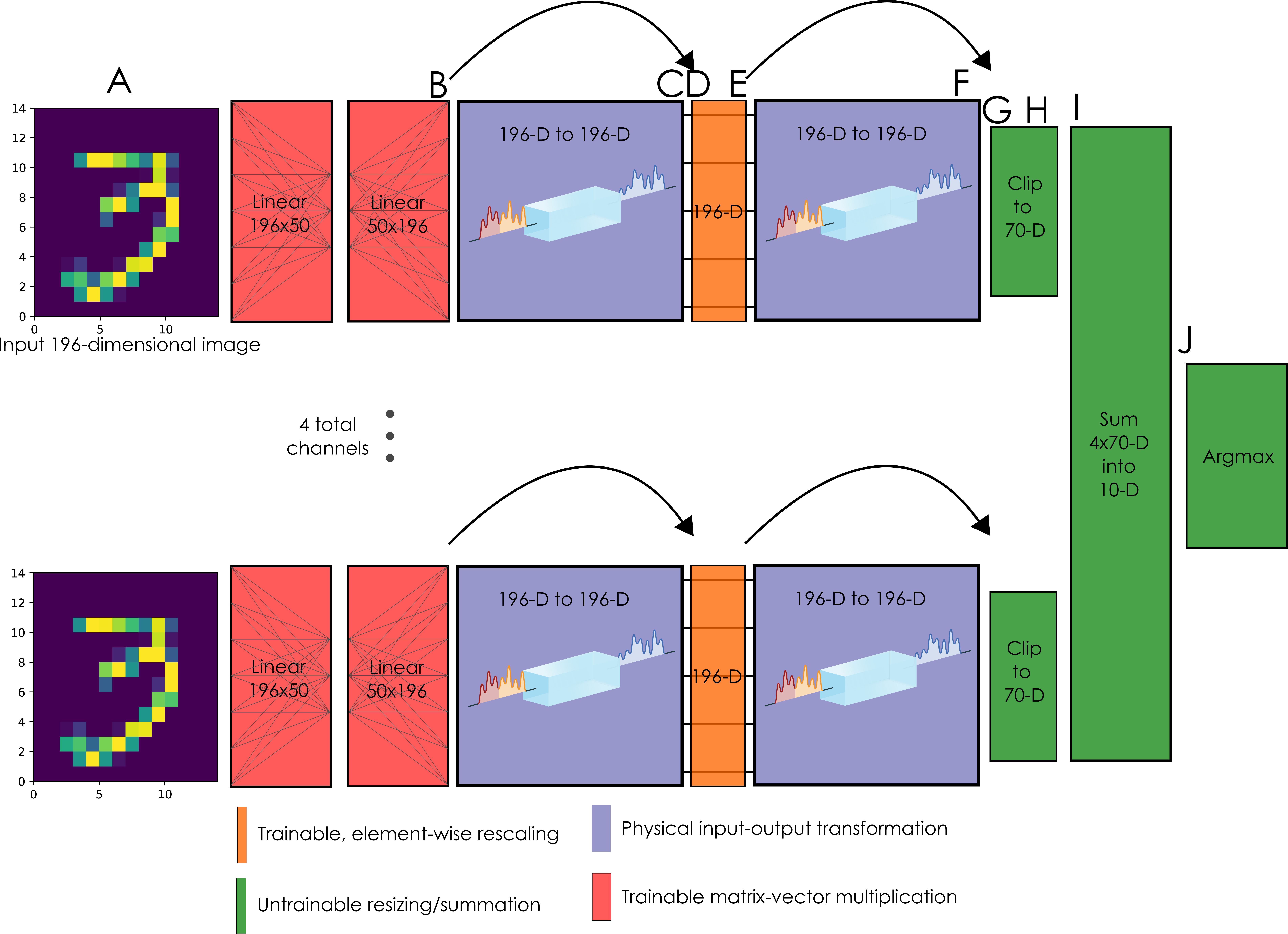}
    \caption{The full PNN architecture used for the SHG MNIST digit classifier. Annotation letters (A,B,C,D,E) refer to the plots in Fig \ref{fig:SHG MNIST1 dat} and the description in the main text.}
    \label{fig:SHGMNIST1 architecture}
\end{figure}

\begin{figure}[h!]
    \centering
    \includegraphics[width=\linewidth]{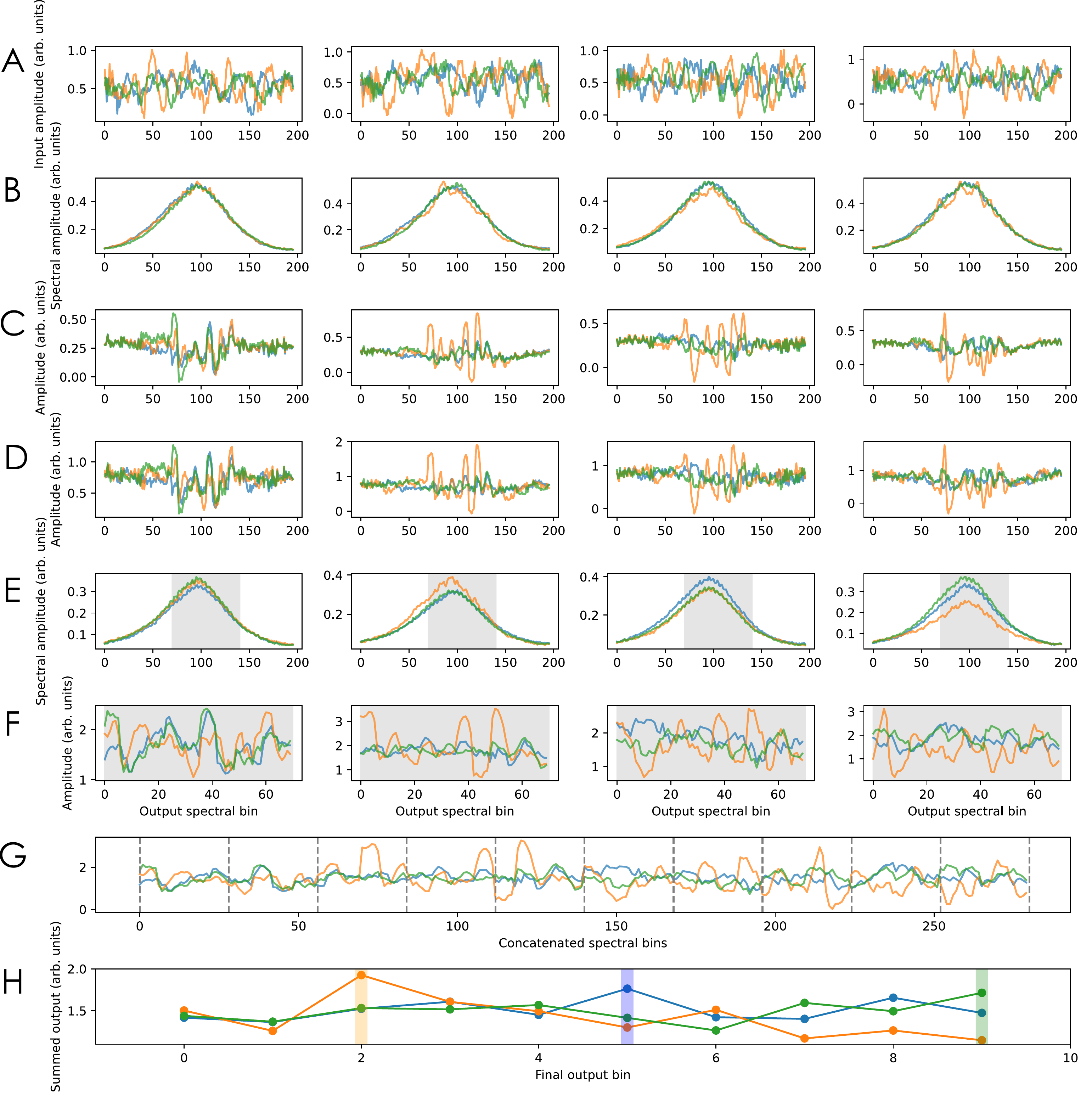}
    \caption{The sequence of operations that make up the SHG PNN MNIST model. For details of each section, see main text. In parts E and F, the grey section shows the region clipped. In part H, the partially-transparent bars show the location of the correct classification. The different color curves (blue, orange, green) correspond to 3 different input images.}
    \label{fig:SHG MNIST1 dat}
\end{figure}

For the hybrid physical-digital MNIST PNN we present in Fig. 4, we chose to use linear input layers to pre-process MNIST images prior to SHG transformations. The structure of the input layer is in two segments, such that the input dimension is reduced to 50, then transformed back to 196, which is the input-output dimension of the SHG transformation. This seemingly unusual choice was made because we observed that the SHG tended to produce very little change in its output spectrum for high-dimensional random input vectors. Thus, we view the two-step input layer as projecting the input MNIST image into a low-dimensional space of SHG `input modes', which ideally produces highly variable output spectra. We found that, in order to maximize the sensitivity of the SHG's output spectra to the shape of the input modulation, it was optimal to use an effective input dimension of about 10-30 ``input modes" (i.e., to choose the input layer to be comprised of two segments, with a middle dimension of 10-30). 

In order to minimize the size of the SHG-MNIST model considered in the main article however, we ultimately increased this input layer dimension to a relatively large value (50) so that good performance could be achieved with relatively few uses of the SHG process. Our desire to keep the model size small (in terms of number of physical system uses) was driven by the relatively slow (200 Hz) speed of the SHG transformation - if the SHG transformation were used many times, it would have been difficult for us to test different configurations. The result of these choices is, however, that the SHG plays less of a role in the hybrid network's calculations than it can with a more optimal choice of input dimension. 

Despite this deviations from a more optimal model, the hybrid model presented in Fig. 4 nonetheless leverages the SHG's nonlinear transformations to effectively gain $\sim$ 7\% improvement on the MNIST digit classification task compared to when the SHG transformations are replaced by identity transformations. While this seems like a small improvement, we note that the difficulty of the MNIST digit classification task is highly nonlinear in accuracy: a model reaching nearly 90\% needs only $\sim 10^3$ operations, while one reaching 96\% or higher may require 100-1000 times as many operations. 

Similar to the electronic PNN, we found it was helpful to include skip connections in the network. This is especially helpful because the SHG transformation does not support an identity operation, and is therefore prone to losing information. 

The PNN architecture used for MNIST digit classification with the SHG physical system in Fig. 4 of the main paper is depicted schematically in Fig. \ref{fig:SHGMNIST1 architecture}. The flow of inputs through the network is shown in Fig, \ref{fig:SHG MNIST1 dat}, where 3 different inputs are shown by different colored-curves. In all three cases, the digits are classified correctly.

Each inference of the full architecture involves the following steps:
\begin{enumerate}
\item 196-D MNIST images are operated on by a trained 196x50 matrix, producing a 50-dimensional vector. Without any nonlinear activation, a second trained 50x196 matrix is applied to produce the input to the first SHG transformation (Fig. \ref{fig:SHG MNIST1 dat}A). This is performed for 4 separate, independent channels (shown in different columns of Fig. \ref{fig:SHG MNIST1 dat}). 
\item The SHG apparatus is executed with the 196-dimensional input, producing a 196-dimensional output vector, which is depicted in Fig. \ref{fig:SHG MNIST1 dat}B.
\item The output of the SHG transformation is operated on by a trained element-wise rescaling, i.e. $x_i \rightarrow x_i a_i+b_i$, resulting in Fig. \ref{fig:SHG MNIST1 dat}C.  
\item To this is added the input to the SHG transformation, multiplied by a trained (1-D) skip weight. This results in Fig. \ref{fig:SHG MNIST1 dat}D.  
\item The SHG apparatus is executed with the 196-dimensional input, producing a 196-dimensional output vector, which is depicted in Fig. \ref{fig:SHG MNIST1 dat}E.
\item The SHG transformation's output is clipped from index 70 to 140, and added to the SHG input, which is multiplied by a trained skip weight. This produces a 70-dimensional vector depicted in Fig. \ref{fig:SHG MNIST1 dat}F.
\item The 4 independent channels' output 70-dimensional vectors are concatenated into a 280-dimensional vector, shown in Fig. \ref{fig:SHG MNIST1 dat}G.
\item The 280-dimensional vector is summed in bins of length 28 to produce a 10-dimensional vector $y_\text{out}$.
\item Perform softmax with a trainable temperature $T$ on $y_\text{out}$ (for training). Perform argmax to obtain the predicted digit (for inference).
\end{enumerate}

\clearpage
\section{Simulation analysis of candidate PNN systems}

In this section, we analyze the simulations of several candidate PNN systems. The purpose of this section is three-fold:
\begin{enumerate}
    \item To identify, based on the form of the numerical simulations, what types of mathematical operations candidate physical systems approximate, 
    \item To determine plausible self-simulation advantages, which set an upper-bound on the computational benefits each PNN system could provide, and 
    \item To analyze the scaling of the latter with respect to physical parameters.
\end{enumerate}

Physical processes closely resemble the structure of DNNs that operate on natural data: real physical processes typically feature low-order nonlinearities, hierarchical structure, local correlations, sparsity, and symmetry \cite{lin2017does}. DNNs that perform well on natural data either exploit these features in their design (as in convolutional DNNs, which intrinsically account for translational symmetry and locality, or in networks trained with data augmentation to improve robustness), or they acquire it as the result of extended training with natural data. Although DNNs are usually nominally deterministic algorithms, noise like that encountered in real physical systems used for PNNs is often added to improve their performance. DNNs are often trained with data that is distorted or has had noise added to improve the generalization of their models or to reduce sensitivity to adversarial examples. During training, adding noise (such as drop-out) is also common for these purposes. In other applications, such as generative models, noise can play a more fundamental role. Noise within PNNs may be useful for similar reasons, provided it is not too large. In sum, all the reasons DNNs work well for computations involving natural data and natural processes are correspondingly strong rationale for why PNNs should be able to perform many of the same tasks modern DNNs perform. 

Physical systems that are good candidates for PNNs implement operations that are well-suited to DNNs, even though they may not always be capable of universal computation. Such candidate PNN processes implement controllable matrix-vector operations and convolutions, as well as nonlinearities and tensor contractions. In the vast majority of physical systems, arbitrary control over these physically-implemented computations will not be easy. Instead, the physical processes will provide these operations approximately and controllable within some subspace of possible operations (typically subject to sparsity, locality and other symmetries) rather than fully-arbitrary mathematical operations. As a result, although these physical systems may provide significant computational benefits via operations similar to those implemented by modern DNNs, they are unlikely to be useful in isolation, and will sometimes require cooperation from a universal digital computer. In this case, it becomes essential to keep track of how many operations the PNN is saving the digital computer - this and related considerations are addressed in the last section of this document.

We find that, generally, self-simulation advantages are higher when the physical system is intuitively more complex. Concretely, this corresponds to it being designed such that: 
\begin{enumerate}
\item Nonlinearity is present, especially higher-order nonlinearities and those that couple multiple degrees of freedom.
\item All degrees of freedom are coupled in heterogeneous ways to many other degrees of freedom.
\item The dimensionality is high (e.g. 3-D versus 2-D). However, this will often correspond to a reduced nonlinearity. 
\end{enumerate}

We find that self-simulation advantages significantly over $10^6$ are common in candidate systems. This is comparable to the increase in speed of supercomputers from 1995 to the present (from roughly 10$^{11}$ to 10$^{17.5}$ GFLOPs \cite{historysupercomputing}. Of course, the actual potential of PNNs depends both on how well the self-simulation advantage estimates computational advantages for other, more useful tasks, how well PNN models can be designed to leverage advantageous physical computations, and how the physical devices used as PNNs might advance in the coming years. Nonetheless, such large self-simulation advantages (SSAs) confirm that, if these non-trivial steps can be completed successfully, non-traditional physical systems do have enormous potential for advancing real-world machine learning applications within the next decades. 

\subsection{General case}

The findings summarized above can be appreciated intuitively by considering a coupled nonlinear differential equation model of a physical dynamical system with $N$ distinct degrees-of-freedom (modes) $x_i$, which interact with up to $n$th-order nonlinearities, and are driven by $N$ driving forces $f_i(t)$

\begin{equation} \label{generalDE}
    \dot{x}_i(t) = \sum_{j=1}^N a_{ij}(t) x_j(t) + \sum_{j=1}^N\sum_{k=1}^N b_{ijk}(t) x_j(t) x_k(t) +  \sum_{j=1}^N\sum_{k=1}^N\sum_{l=1}^N c_{ijkl}(t) x_j(t) x_k(t) x_l(t) + ... + f_i(t).
\end{equation} 

For such a system, if the coupling of the $N$ degrees of freedom is heterogeneous and non-negligible over $M$ neighbours for all orders of nonlinearity, then the simulation complexity is dominated by the highest-order nonlinear term as $\Delta T \Delta f NM^n$, where $\Delta f$ is the bandwidth of the system (the inverse of the necessary time step) and $\Delta T$ is the simulated time. Note that, because sparsity, locality and other symmetries can be exploited to make each term faster to numerically evaluate, it is important to consider the coupling range $M$ of heterogeneous, non-negligible coupling. In most systems, the coupling range depends on the order of the nonlinearity, so a more general expectation is 

\begin{equation}
    \text{Simulation Complexity} \sim \Delta T \Delta f N Q_{\text{max}},
\end{equation}
\nobreak where $Q_{\text{max}} = \text{max}[M_i^i]$ is the maximum coupling complexity over the different terms, where $M_i$ is the coupling range of heterogeneous coupling for the $i$th order nonlinear term.

The power required to sustain such a physical system will usually be linear in the number of modes $N$, and the time taken is of course $\Delta T$, so the self-simulation advantage will typically scale with $Q_{\text{max}}$. This however assumes that $n$ and $M$ are independent of the power supplied to the physical system, which will generally not be true: in most systems, as the energy localized in the system is increased, more nonlinearity and more complex, long-range coupling are necessary to describe the physics. As the dimension of the dynamics increases, the ability for localized degrees of freedom radiating a fixed amount of energy to couple with neighbours increases exponentially. This is intuitive: higher-dimensional physics is harder to simulate. 

Equation \ref{generalDE} also identifies, very generically, some of the physical data transformations realized by physical systems. If the physical parameters of the system that determine $a_{ij}$, $b_{ijk}$, $c_{ijkl}$ can be adjusted and trained, then the evolution of the $x_i(t)$ implement matrix-vector operations (first term) and higher-order tensor contractions (subsequent terms). Depending on the locality and symmetry of $a_{ij}$, $b_{ijk}$, $c_{ijkl}$, these operations will often be similar to convolutions. In general, these coefficients could also be controlled in time, $a_{ij}(t)$, $b_{ijk}(t)$, $c_{ijkl}(t)$. While this naturally opens up significantly more complex sequences of operations in a single physical system, it may not be compatible with in-place parameters (parameters which are stored in-place in the physical device, and therefore do not need to be retrieved from a digital memory each time an inference is performed). For this reason (see the last section of this document), we expect the most useful, scalable parameterization of most physical systems to be the use of trainable, but time-independent physical parameters $a_{ij}$, $b_{ijk}$, $c_{ijkl}$.

Alternatively, if some of the $x_i(t)$ are controllable parameters (i.e., are treated as forced source terms, rather than terms obeying Eqn, \ref{generalDE}), then one may recognize the nonlinear terms as realizing tensor operations of order $n$, where $n$ is the order of nonlinearity (for example, the quadratic term $\sum_{j=1}^N\sum_{k=1}^N b_{ijk} x_j x_k$ could result (in part) in a controlled matrix-vector operation if some of the $x_k(t)$ are controlled parameters). If the drive terms $f_i(t)$ are used as controllable parameters, the outcome will probably be similar to controlling $x_i(t)$, depending on the memory of the system's dynamics, and the strength of the drive. For the same reasons mentioned in the last paragraph, such time-dependent physical parameters may pose scaling challenges if they require fast reading from a digital memory, and digital-to-analog conversion. 

This general model also suggests that the greatest physical advantages can be gained with operations that are typically not used in modern deep neural networks, namely high-dimensional tensor operations, since these are associated with higher-order nonlinear coupling. Of course, this depends on utilizing these operations efficiently to perform machine learning, which has not been widely studied. Understanding how to use operations of this kind efficiently for machine learning could thus help to advance the computational benefits obtained with PNNs.  

In the rest of this section as follows, we will analyze the physical costs (energy and time) to evolve a selection of realistic PNN modules based on different physical systems, then compare these physical costs to those incurred by performing a simulation with a state-of-the-art digital electronic device. For these comparisons, we will consider the Nvidia DGX Superpod supercomputer \cite{Green500Nov2020} which achieves 39 pJ per floating point operation and 2.8$\times 10^{15}$ operations/s, and the Nvidia Titan RTX GPU \cite{NvidiaTitanRTX}, which achieves 9.8 pJ/operation and 32$\times 10^{12}$ operations/s for half-precision operations. 

\subsection{Single-mode nonlinear optical pulse propagation: scaling}
For ultrafast pulse propagation in nonlinear dispersive media, a standard simulation algorithm is the split-step Fourier method \cite{agrawalNLFO}. We will use the computational complexity of this algorithm as our basis for assessing self-simulation advantage. 

For an optical propagation length $L$, the split-step algorithm divides the propagation into small steps whose length must be much shorter than the rate at which nonlinear effects occur, $L_{\text{nl}}$. The pulse is described by a vector of $N_{\omega}$ values which contain the complex amplitudes of the electric field envelope. The number of spectral points is determined by the maximum frequency range of the pulse and any new frequency content the propagation creates, and by the time window needed to describe the pulse throughout the propagation, $N_\omega = T\Delta \omega$, where $\Delta \omega$ is the bandwidth of the simulation and $T$ is the required time window of the simulation.

In the split-step algorithm, linear effects like dispersion are efficiently applied in the Fourier domain. The complexity of the split-step algorithm is dominated by the Fourier transform operations required, so:
\begin{equation}
N_{\text{ops}} \sim N_z N_\omega \log(N_\omega),
\end{equation}
\nobreak where $N_z$ is the number of longitudinal steps required. The minimum step size is proportional to the rate of nonlinear effects, $N_z = L\times \gamma f(P)$, where $f(P)$ is the dependence of the nonlinear rate on the peak power $P$ of the pulse and $\gamma$ is the nonlinear constant, which depends on the material, the type of nonlinearity, and the transverse confinement of the single spatial mode the pulse propagates in. For Kerr (cubic) nonlinearity, $f(P)=P$, whereas for quadratic nonlinearity $f(P)=\sqrt{P}$, generally $f(P) = P^{(n-1)/2}$ where $n$ is the order of the nonlinearity. 

The peak power $P$ is related to the minimum temporal feature of the pulse, $\sim 1/(\Delta \omega)$, where $\Delta \omega$ is the bandwidth of the simulation. $N_\omega = T\Delta \omega$, where $T$ is the required time window of the simulation. In other words, $N_\omega$ is the maximum time-bandwidth product that occurs during the propagation, which generally increases for more complex, dispersive and nonlinear pulse propagation. For the sake of simplifying the discussion, we will let $P_0 = E_p N_\omega$, though we note that this will often overestimate the nonlinearity. 

The maximum rate at which physical computations can be performed is determined by the longest of the following: duration between input light pulses, duration between electro-optic modulations, and the maximum duration of the optical pulses during propagation, with the last required to ensure that each physical computation is independent (which is usually, but not always required for classification tasks). We will assume this corresponds to about 33 ps (so a rate of 30 GHz). Pulsed sources of this repetition rate are possible \cite{carlson2018ultrafast}, and electro-optic modulators with bandwidths in excess of 40 GHz are standard. Provided the length of the nonlinear optical medium, and the dispersion and pulse spectral broadening are chosen so that the maximum pulse duration never exceeds 33 ps, 30 GHz operation should therefore be feasible.  

The physical advantage for computation (PAC) in time is then:
\begin{equation}
\text{PAC}_t = \frac{\text{Time for simulation}}{\text{Time for physical evolution}} = \frac{t_c N_{\text{ops}}}{t_p}, 
\end{equation}
\nobreak where $t_c$ is the time per operation for the reference conventional computer, and $t_p$ is the time the physical evolution takes (being limited by one of the above considerations). 

Similarly, the physical advantage for computation in energy is:

\begin{equation}
\text{PAC}_e = \frac{\text{Energy for simulation}}{\text{Energy for physical evolution}}=\frac{e_c N_{\text{ops}}}{E_{\text{phys}}},  
\end{equation}
\nobreak where $e_c$ is the energy per operation of the reference computer and $E_{\text{phys}}$ is the total energy required to instantiate and measure the pulse propagation. Instantiating the pulse propagation and measuring its result requires significantly more energy than just that needed for the optical pulse, as an example will show.  

\subsubsection{Quantitative example}
In this section, we calculate the energy and speed self-simulation advantage of a realistic, scaled device based on ultrafast nonlinear pulse propagation. Our intention here is not to design a device that would maximize the self-simulation advantage within the most optimistic conditions, but to consider a realistic (though not trivial) experimental demonstration possible with existing technology. Throughout, we will note room for improvement in near-future devices, and at the end analyze both the realistic present-case and near-future potential.

We consider a device as follows. An ultrashort pulse at 1550-nm is split into 20 wavelength bins, where it can be modulated by electro-optic amplitude modulators and then recombined. Through these electro-optic modulators, input vectors can be applied at $\sim$ 10-100 GHz via spectral modulations. For physical parameters, the pulse is then shaped by a phase pulse shaper based on a spatial light modulator or similar device. This device allows $\sim$ 10$^3$ or more independent controllable parameters for the subsequent nonlinear pulse propagation, which can be stored in-place over many uses of the PNN without incurring energy costs due to memory reads or digital-analog conversion. The modulated pulse then propagates in a nanophotonic waveguide based on periodically-poled thin film lithium niobate. To maximize the complexity of the dynamics, we assume the waveguide is poled to create an artificial Kerr effect.

In Ref.~\cite{jankowski2020ultrabroadband}, 1 pJ pulses were sufficient to observe significant supercontinuum pulse propagation in a cm-scale waveguide. However, the pulses used were simple, transform-limited pulses. To ensure very complex pulse propagation with the structured, data-encoded pulses used for computation, we assume that roughly 50 times as much pulse energy should be used (spread across a longer, more complex pulse). To account for various optical losses in a realistic device (transmission around ~1\%), and the wall-plug efficiency of the initial mode-locked laser (around 10\%), we assume a total of 50 nJ of energy is required per optical pulse. This energy could be reduced by several orders of magnitude in a fully-integrated device, since the optical losses would be drastically reduced. Assuming a 0.5 cm waveguide, the pulse duration should remain below 33 ps throughout the waveguide (for a dispersion of -15 fs$^2$/mm, such a waveguide would correspond to at most 10-20 dispersion lengths, so a maximum duration of 10-20 ps is reasonable to expect given the expected pulse bandwidths described below). 

To inject data to the physical system, it must be converted from a digital computer's digital format to a format (typically analog) suitable for the physical system. The most energy-efficient ADCs are capable of approximately 1 pJ/conversion \cite{xu201623mw} at 24 GHz. For the sake of this order-of-magnitude estimate, we will assume ADC and DAC operations each require 1 pJ/conversion. Integrated electro-optic modulators that operate with $\sim$ pJ/modulation are typical \cite{sun2015single,atabaki2018integrating,wang2018integrated}, while modulators capable of few-bit-depth modulation appear possible \cite{wang2018integrated,koeber2015femtojoule,srinivasan201556}. For our purposes here, we will assume each fast optical modulation requires an additional 1 pJ on top of the DAC cost. Although on-chip photodetectors with ultralow capacitance may enable amplifier-free detection of light pulses as weak as 1 fJ \cite{nozaki2016photonic,tang2008nanometre}, we will assume that detectors add an additional 1 pJ on top of the ADC cost for a modern experiment. Finally, the spatial light modulator requires roughly 10 W to maintain, resulting in 330 pJ per 33 ps inference. Of course, in future devices this would be significantly reduced, presumably to significantly below 1 W.

Adding these physical cost per inference is: 
\begin{equation}
    330 \text{ pJ} + 20 \times 1 \text{ pJ} + 20 \times 1 \text{ pJ} + 20 \times 1 \text{ pJ} + 50 \text{ nJ} = 50.39 \text{ nJ} 
\end{equation}
\nobreak for the present-day experiment, and
\begin{equation}
    33 \text{ pJ} + 20 \times 0.1 \text{ pJ} + 20 \times 0.1 \text{ pJ} + 20 \times 10 \text{ fJ} + 0.5 \text{ nJ} \approx 537 \text{ pJ}
\end{equation}
\nobreak for an improved, near-future device. For both devices, the time-per-computation would be limited by the propagation time, 33 ps, since fast electro optic modulators and DAC/ADC would permit input/output rates exceeding 1/(33 ps) = 30 GHz.

We now consider the energy and time costs for simulating this system with state-of-the-art conventional hardware. For the 50 pJ pulses we have assumed, if the waveguide is designed to support weak anomalous dispersion, peak powers of roughly 1 kW ($\sim$ 10 pJ localized into a 10 fs region) are probable, corresponding to a nonlinear length scale of \SI{10}{\micro\metre} for the device in Ref. \cite{jankowski2020ultrabroadband} which has an effective self-phase modulation coefficient of 100/Wm. To adequately simulate this would require a longitudinal step size of around \SI{1}{\micro\metre} or smaller. To accommodate spectral broadening up to around 400 THz, a spectral range $\Delta \omega$ of 800 THz is required for accurate simulations. To accommodate the complex shaped pulses and dispersion of the broadband light, a temporal window size of 10-50 ps would be expected, since the initial pulse duration would be roughly 1-5 ps depending on the spectral modulations applied (100-1000 phase degrees of freedom across a $\sim$ 50 THz input bandwidth), and the 5 mm waveguide corresponds to roughly 5-8 characteristic dispersion lengths for the broadened pulse (assuming a dispersion of -15 fs$^2$/mm).  

Thus, using the Fourier split-step method, the number of operations for simulations would be:

\begin{equation}
    N_z N_\omega \log_2(N_\omega) = (5 \text{ mm} / \SI{1}{\micro\metre}) \times 800 \text{ THz} \times 25 \text{ ps} \times \log_2 (800 \text{ THz} \times 25 \text{ ps}) \approx 5\times 10^8 \text{operations.}
\end{equation}

Using the Nvidia Titan RTX, this would cost 5 mJ and take \SI{16}{\micro\second}.  This corresponds to roughly 10$^5$ times more energy and 10$^5$ times longer than the present-day experiment, and roughly 10$^7$ more energy and 10$^5$ times longer than the near-term device. 

Using the Nvidia DGX SuperPOD, the simulation could be performed more quickly, 179 ns, but with greater energy cost, 20 mJ.  This corresponds to roughly 10$^5$ times more energy and 10$^4$ times longer than the present-day experiment, and roughly 10$^8$ more energy and 10$^4$ times longer than the near-term device. 

Although the self-simulation advantage possible with nanophotonic nonlinear optics is large, the simulation analysis presented here highlights some important caveats. First, the scaling of the self-simulation advantage for single-mode nonlinear optical pulse propagation is in general only marginally superlinear with respect to the optical degrees of freedom, $N_\omega$, since the logarithmic factor is relatively weak. Since larger $N_\omega$ can give rise to shorter pulses, in some regimes $N_z$ could scale optimistically as $N_\omega$, suggesting a best-case scaling of $N_\omega^2\log_2(N_\omega)$. It is not obvious if pulse propagation that supports this complexity scaling will also be useful for practical computations, however. A more crucial caveat is that the operations performed by single-mode nonlinear optical pulse propagation are, by default, not well-matched to the standard operations performed in modern deep neural networks, rather they are high-order tensor contractions. Furthermore, because of energy and momentum conservation in these nonlinear interactions, it will rarely be possible to realize fully arbitrary high-order tensor contractions or convolutions. While PAT and PNNs can facilitate training physical computations based on these exotic operations (this was one scientific motivation for using nonlinear optics as a first demonstration of PNNs here), it is not yet clear how those PNNs should be designed to be useful, since machine learning based on their underlying operations is not well-studied. In summary, while single-mode, nanophotonic nonlinear optics presents many promising features for PNNs, we believe that significant effort will be required to understand how to utilize the exotic nonlinear optical operations in practical tasks. In contrast, other systems (such as multimode waves, coupled oscillators or electrical networks) discussed subsequently have a more direct connection to standard machine learning operations, so we expect it will be easier in the short term to apply these systems as PNNs, using standard machine learning techniques, to practical problems.  

\subsection{Self-simulation analysis of multimode wave propagation}
Multimode wave propagation is an appealing platform for physical neural networks for similar reasons that motivate its use in physical reservoir computing \cite{rafayelyan2020large,teugin2020scalable}. In the linear regime, multimode wave propagation permits very high-dimensional random matrix-vector multiplications with potentially very low-energy \cite{rafayelyan2020large}. If the medium light propagates through can be controlled, then these matrix-vector operations will likewise be controllable, making for a very direct connection to standard deep neural network operations. In the nonlinear regime, the complexity of pulse propagation can grow significantly, allowing for PNNs with much greater self-simulation advantages. This logic has motivated using multimode nonlinear wave propagation for reservoir computing \cite{teugin2020scalable}.

\subsubsection{Linear propagation}

All linear multimode wave propagation can be modelled as a matrix-vector operation, where a transmission matrix $T$ operates on an input vector of complex mode amplitudes $\vec{x}$ \cite{popoff2010measuring}. Thus, the number of digital operations required to simulate a linear multimode wave system scales as $MM'$, where $M$ is the number input modes and $M'$ the number of measured output modes. Since the energy required to implement the wave propagation scales linearly with $M$ (assuming $M'=M$), and the energy required to perform the matrix-vector multiplication scales as $MR$, where $R$ is the rank of the transmission matrix, the self-simulation advantage scales linearly with the number of modes. 

As an example, multimode fibers can support $\sim 5\times10^6$ modes, and fibers of $\sim 10$ cm length can exhibit all-to-all random coupling \cite{xiong2017principal}. An experiment involving a 1 mW beam, a 30 kHz digital micromirror device with 5 million pixels, and a high-speed 1 Mpixel camera would consume approximately 10 mW + 1 W + 5 W $\approx$ 6 W and performs, in effect, $(5\times10^6)^2 \times 30 \times 10^3 \text{Hz} \sim 8\times 10^{17}$ operations per second, corresponding to about 10 aJ/operation. In comparison, the GTX GPU (DGX Superpod) requires $10^6$ ($5\times 10^6$) times more energy, and $10^4$ (300) times more time. Even if such an operation has a limited number of trainable parameters, large random matrix operations appear to be useful for a variety of applications \cite{ohana2020kernel,rafayelyan2020large}. As components within PNNs, they could be integrated within conventional neural architectures. Their parameters could be trained using gradient descent alongside others using PAT. Even in implementations without trainable parameters, PAT would also allow backpropagation through their operations, to train preceding or subsequent trainable operations as part of hybrid digital-physical architectures. Nonetheless, it is obvious that having controllable physical parameters to permit arbitrary, or nearly-arbitrary matrix-vector operations would give rise to a stronger connection with the format of standard deep neural networks, and therefore probably to a faster route to good performance on practical tasks.

\subsubsection{Nonlinear propagation}

In the nonlinear regime, multimode wave propagation is considerably more complex, and the self-simulation advantages can be very high. While such large self-simulation advantages recommend complex nonlinear wave systems (and many-body nonlinear dynamical systems in general), we caution that these large numbers should be taken with significant skepticism with respect to their translation to general-purpose advantages across a wide range of practical problems. While the performance of multimode nonlinear wave propagation-based reservoir computers suggests that some practical advantages may be obtained \cite{teugin2020scalable}, it is still challenging to finely control multimode nonlinear waves and the main operations that contribute to the large self-simulation advantages are exotic to modern neural networks. Overall, much more work is necessary.

Additionally, systems with extremely large self-simulations advantages should be evaluated with caution because it may be possible to dramatically accelerate the simulations of these systems not only by more efficient simulation algorithms, but also by data-driven simulations, such as in Ref.~\cite{kasim2020up}. This will be especially likely if the simulations need only to consider a relatively constrained set of possible inputs, rather than a fully general simulation, or when interactions are subject to symmetry constraints. 

With these caveats noted, we now consider multimode nonlinear optical wave propagation. The standard model for multimode Kerr media scales with the number of guided modes $M$ as $M^4$ \cite{poletti2008description}. Although this scaling can be significantly alleviated if symmetry assumptions can be applied \cite{laegsgaard2017efficient}, in fibers with strong disorder (which implements all-to-all linear coupling as considered above) these assumptions will be broken and the $M^4$ scaling will hold. In highly-multimode fibers, more efficient simulations may be performed by simulating light propagation using a 3+1-dimensional split-step Fourier algorithm, similar to the one considered in the previous section \cite{wright2017multimode}. However, these simulations still scale at least as fast as $M^2N_\omega/E_pL_b$ in the general case, where $M$ is the number of guided modes, $E_p$ is the pulse energy, $N_\omega$ is the number of frequency points, and $L_b$ is the beat length for interference between involved modes. Since $L_b \sim 1/\Delta_n$ grows with the number of modes in the fiber, the effective scaling is roughly $M^\alpha N_\omega/E_p$, where $\alpha$ is a scaling parameter at least larger than $2$.

In the former experiment, assuming a 1 W pulsed laser source with 31 kHz repetition rate, \SI{32}{\micro\joule} pulses would produce complex supercontinuum in even highly-multimode fibers such as the $5\times 10^6$ one considered above. To simulate such pulse propagation, $N_x > 3000$ spatial grid points ($\approx \sqrt{5\times 10^6}$, the best-case number for a fiber with $5\times 10^6$ modes) and $N_\omega >10^4$ ($\approx$ 100 THz $\times$ 100 ps, the expected spectral bandwidth times a minimal estimate of the temporal window required). Assuming an core-cladding index difference of 0.01 and a 532 nm center wavelength, the beat length of $\Delta_n/\lambda$ is \SI{53}{\micro\metre}, suggesting a best-case step-size of $\sim$ \SI{25}{\micro\metre}. 
Then, the number of operations required to simulate a 100-cm fiber evolution would be approximately:

\begin{equation}
    [N_x\log_2(N_x)]^2N_\omega \log_2 N_\omega L/L_{\text{step}} = 6.4\times 10^{18}\text{ operations.}
\end{equation}

Since these operations are, in effect,  physically executed at a 31 kHz rate (limited by the DMD update), the effective operation rate is $2\times 10^{23}$ operations/s. Assuming the same components as described above and a 1\% laser wall-plug efficiency, this corresponds to \SI{106}{W}/$2\times 10^{23} = 5\times 10^{-22}$ J/operation. This number is astoundingly small: the physics is $10^{10}$ ($10^{11}$) more energy efficient than a digital simulation using the RTX GPU (DGX Superpod), and executes 10$^5$ (2000) times faster. 

\subsection{Multimode nonlinear oscillators/oscillator networks}

Networks of coupled oscillators can be simulated using a set of differential equations like \ref{generalDE}, although for bulk multimode oscillators in which the modes overlap significantly in space it will usually be more efficient to instead simulate partial differential equations for the time-evolution of the bulk field in $(x,y,z)$. As an example of a locally-connected network of localized nonlinear oscillators, we consider nanoscale spintronic oscillators, such as those used for reservoir computing in Ref. \cite{Torrejon2017GrollierspintronicNN}. In that work, they claim that the power consumption per oscillator could be as low as \SI{1}{\micro W}, with resonance frequencies into the 10s of GHz \cite{sato2014properties}. 

If we assume that nonlinearity in the oscillators is strictly local, the number of operations required to simulate a network of coupled spintronic oscillators is then:

\begin{equation}
N_{\text{ops}}\approx \Delta T \Delta f [NM + N]
\end{equation}
\nobreak where $\Delta T$ is the simulated time, $\Delta f$ is the relevant frequency bandwidth, $N$ is the number of oscillators, $M$ is the number of nearest neighbours each oscillator is (linearly) coupled to. 

Given these assumptions, the coupled oscillator network follows an equation that closely resembles a typical deep residual neural network \cite{he2016deep}: 

\begin{equation} \label{oscNN1}
    \dot{x}_i(t) = \sum_{j=1}^N a_{ij}(t) x_j(t) + f_i(t) + \sigma(x_i(t)),
\end{equation} 
\nobreak where $\sigma$ is the local nonlinearity of each oscillator. As is noted in neural ordinary differential equations \cite{chen2018neural}, integrating the above equation using, for example, an Euler method, leads to an update equation:

\begin{equation} \label{oscNN2}
x_i[t+1] = x_i[t] + \Delta t (f_i[t] +  \sum_{j=1}^N a_{ij}[t] x_j[t] + \sigma(x_i[t])),
\end{equation} 
\nobreak which is just the layer-by-layer propagation through a deep residual neural network. 

Let us assume a 10 GHz resonance, such that the maximum relevant frequency in a simulation is 20 GHz. We'll consider $N=100$ oscillators with a coupling range of $M=100$ and assume that nonlinearity can be implemented by 1 operation per oscillator per step. 

For a time of 1 ns, the number of operations is then:

\begin{equation}
N_{\text{ops}}\approx \Delta T \Delta f [NM + N] = 1 \text{ ns} \times 20 \text{ GHz} \times [100\times 100 + 100] \approx 2 \times 10^5
\end{equation}

This costs \SI{2}{\micro\joule} (\SI{7.9}{\micro\joule}) and takes 6.3 ns (72 ps) on the GTX GPU (DGX Superpod) respectively. 

If we assume that each oscillator requires 2.5 times its nominal power requirements to allow it to efficiently couple with 100 neighbours, each oscillator requires \SI{2.5}{\micro W} and the total energy required physically is 250 fJ. However, the input and output voltages signals will in general require conversion from/to digital format. As discussed in the previous sections, a cost of 1 pJ/sample for each DAC and ADC operation at 20 GHz is plausible \cite{xu201623mw}. For 100 inputs and 100 outputs with 20 GHz $\times$ 1 ns = 20 samples each, the DAC and ADC costs are each 100 $\times$ 20 $\times$ 1 pJ = 2 nJ. Thus, the total energy cost to evolve the physical system is about 4 nJ. 

Thus, the self simulation advantage of the network is about $10^3$ for energy, and between about 0.1 to 10 for speed. While this seems relatively small compared to previous subsections, it is straightforward to increase. 

In this example, the energy required for the oscillators themselves was negligible compared to the DAC and ADC costs. Thus, we can increase the complexity of operations the device performs by increasing the `hidden' oscillators. Let us now consider a network of $N=10^5$ coupled oscillators, each coupled to $M=100$ neighbours, and of which only $N_{\text{in}}=100$ and $N_{\text{out}}=100$ are used for input and output. The rest of the oscillators we imagine to be used to cause and control the physical transformation. Their injection currents might be trainable parameters which can be changed at a much slower rate during training, and otherwise maintained in-placed during inference (and therefore not incurring DAC/ADC costs each inference). If we assume that \SI{10}{\micro W} is required to maintain each of these ``parameter" oscillators, then the total energy for oscillators over 1 ns is about 1 nJ. This is still less than the energy required for DAC and ADC.

For the larger oscillator network, the number of operations is however much larger, $2\times 10^8$. Consequently, the calculation costs $10^5$ times more than the first example: it costs 2 mJ (8 mJ) and takes \SI{6.3}{\micro\second} (71 ns) on the GTX GPU (DGX Superpod) respectively. And, as a result, the larger oscillator network's SSAs are around $10^6$ to $10^7$ for energy and 70 to 6000 for speed. 

We note that greater speed advantages are possible if one increases $N$ while holding $N \Delta T$ constant. When this is done, the SSA increases linearly with $N$. Of course, there will be a minimum useful $\Delta T$, and as the number of oscillators increases, it will usually be necessary to increase this time so that distant ones can meaningfully interact. Thus, how much spatial parallelism/scaling can be exploited is unclear and will depend on the system and the required transformations. 

Finally, we note that this section has neglected energy costs associated with the controllable linear coupling. We have inherently assumed a technique that requires no additional power to maintain, such as bi- or multi-stable circuit elements. Even though these devices may require power to adjust, their influence on the cost of performing inference could be, ideally, close to zero. We will expand on this point at the end of the next sub-section, but obviously if connections require active power to maintain, even very small amounts, it could rapidly become non-negligible due to the enormous number of connections involved.

\subsection{Nonlinear electrical circuit dynamics/analog transistor networks}

To describe the transient analog nonlinear dynamics of a network of transistors, one may use a nonlinear circuit simulation strategy, solving a system of nonlinear differential equations of at least size  $N_TN_d$, where $N_T$ is the number of transistors, and $N_d$ is the number of dimensions ($\sim$nodes) in the equivalent circuit model of the transistor. For accurate large-signal analog models such as the EKV model, $N_d \sim 10$. 
To solve for the dynamics of a nonlinear network of transistors, a common technique is to numerically integrate the differential equations that describe the network by solving at each time-step the nonlinear system of equations using a nonlinear solver such as Newton-Raphson. Alternatively, if we can approximate all nonlinearities by low-order Taylor-series approximations, one may instead integrate a system of coupled nonlinear differential equations directly. Since the nonlinearities in electrical circuits are typically strong, including strong saturation and exponential dependencies, the latter approach is not often used.  

The standard approach scales in complexity as $\mathcal{O}(n^{1.3 -1.8}$) \cite{gunther2005modelling,rewienski2011perspective,benk2017holistic}, with the most complex step due to the linear system solving required for each iteration of the Newton-Raphson method. For simulating very large circuits, there are methods to improve this scaling depending on the topology and details of the circuit \cite{rewienski2011perspective,benk2017holistic}. However, these methods sacrifice accuracy, and depend on the specific circuit topology. 

Even though it is less accurate, integrating a system of equations similar to \ref{generalDE} allows more straightforward insight into the circuit simulation scaling and the operations it effectively performs. Assuming integration of the nonlinear system of differential equations, the number of operations to simulate the transistor network is nominally:
\begin{equation}
N_{\text{ops}}\approx \Delta T \Delta f [N_TN_dM_1 + N_TN_d^2M_2^2 + ... + N_TN_d^nM_n^n],
\end{equation}
\nobreak where $\Delta T$ = time of simulation, $\Delta f$ = bandwidth of simulation (the inverse of the time-step, roughly the  maximum frequency expected to be encountered in the dynamics), $N_d$ = number of degrees of freedom needed to describe each transistor $\sim$ 10-20, $n$ = highest polynomial order of the nonlinear coupling needed to describe the operating regime, and $M_i$ = range of transistor-transistor coupling (i.e., the number of other transistors each transistor is non-negligibly coupled to) for each order of nonlinearity. 

Although it may not be true when nano-sized transistors are operated near one another at radio frequencies, we will assume that the nonlinear terms are local to each transistor, so $M_i=1$ for $i$ larger than 1. In this case, 

\begin{equation}
N_{\text{ops}}\approx \Delta T \Delta f [N_TN_dM_1 + N_TN_dC_{\text{nl}}],
\end{equation}
\nobreak where we have assumed that a more efficient method is used to calculate the nonlinearity at each transistor, such that the scaling is a constant factor larger than the number of degrees of freedom required to simulate each transistor. We have done this because there are obviously much more efficient methods for the transistor nonlinearity than the full ODE expansion in Eqn. \ref{generalDE}, and we would thus grossly overestimate the simulation cost. If we let $C_{\text{nl}} = N_d$, it is in some sense equivalent to assuming that the calculation of the nonlinearity is about as efficient as if the nonlinearity were approximated by terms up to quadratic. Provided $C_{\text{nl}}$ is not much larger than $M_1$, its choice is not significant in our final estimate, since the nonlinear part of the simulation will be comparable to or smaller than the linear coupling part. 

Before considering a concrete example, it is worthwhile to emphasize that, again, with these assumptions, and assuming that the internal complexities of the transistor are abstracted away, the physical system follows an equation for which each numerical step resembles propagation through a residual layer: 

\begin{equation} \label{transNN}
x_i[t+1] = x_i[t] + \Delta t (f_i[t] +  \sum_{j=1}^N a_{ij}[t] x_j[t] + \sigma(x_i[t])).
\end{equation} 

Although the transistor network differs in some details from the oscillator networks just considered, they both ultimately resemble typical neural networks provided their linear coupling is controllable, and their nonlinearity is relatively local. Since simulating the equivalent circuit for a transistor in the nonlinear, RF-regime is significantly harder than what we assumed for the nonlinear oscillators, the transistor networks may have a modest constant-factor advantage in self-simulation over nonlinear oscillators (around 10-20 times). While PNNs could make use of such physics beyond Eqn. \ref{transNN}, it is difficult to estimate how much computational benefit such higher-order physics would confer. The main takeaway is just that, up to some approximations and abstractions, networks of coupled transistors (or diodes) perform operations as they evolve that are very similar to those in conventional neural networks. Therefore, like the oscillator networks in the previous section, we feel relatively more confident in asserting that PNNs based on these systems may ultimately be able to approach their self-simulation advantage for tasks of practical interest (i.e., that computational advantages for practical tasks could be as high as $\sim$ 1$\%$ of the self-simulation advantages). 

As an example, let us assume an electronic network of typical modern transistors. Modern CPUs/GPUs require ~10 W per billion transistors ($\sim$ 100 nW per transistor, on average) \cite{Sun2019}, but in part this power requirement relies on efficient use of transistors such that all are not on all the time, so the power consumption per transistor during its operation is much higher. Clock rates of up to few GHz are typical. In typical digital usage, these transistors are operated in constrained safe-operation regimes. However, as analog devices, these same transistors are more complex, especially if driven outside these typical regimes, for high-frequency analog signals beyond the small-signal regime \cite{enz1995analytical,enz2002mos}. We will therefore assume the transistors dissipate roughly 1000 nW each, and can be operated at digital clock cycles of a few GHz, such that their nonlinear analog dynamics can be as fast as 20 GHz or so. In addition, we will typically want to use digital-analog conversion for inputs. As is discussed in the last section, it is ideal if the majority of the circuit's parameters are stored in-place. We will therefore assume there are $N_T$ simultaneous, time-dependent inputs to and outputs from the circuit, requiring $N_T$ DAC and ADCs. 

Let us assume $N_T$ = 100 transistors described by a $N_d = 10$ equivalent circuit, an evolution time of 10 ns, 5 GHz bandwidth inputs (so that a maximum frequency of 20 GHz occurs in the simulation) and a coupling range of $M=100$ neighbouring transistors. This leads to a physical energy cost of:
\begin{equation}
    N_T \times P_\text{{transistor}} \times \Delta T = 100 \times 100 \text{ nW} \times 10 \text{ ns} = 1 \text{ pJ}
\end{equation}
\nobreak for the transistors themselves, and

\begin{equation}
    N_T \times 2 \times \Delta T f_{\text{ s}} = 100 \times 10 \text{ ns} \times 1 \text{ pJ} \times 5 \text{ GHz}  = 10 \text{ nJ}
\end{equation}
\nobreak for the DAC and ADC, where $f_{\text{s}}$ is the input and output sampling frequency. Arguably, the transistor power requirement can be neglected, both because it is much smaller than any DAC/ADC cost, but also because the signals from the DAC provide the power required to drive the transistors. We will however not do this, since we are interested in order-of-magnitude estimates that are preferably conservative. 

Meanwhile, simulations require:

\begin{equation}
    N_{\text{ops}}\approx \Delta T \Delta f [N_TN_dM_1 + N_TN_dC_{\text{nl}}] = 10 \text{ ns} \times 20 \text{ GHz} \times [100\times10\times100 + 100\times10\times10] \approx 10^7
\end{equation}

As a result, the digital simulation, which consists effectively of locally-structured matrix-vector multiplications and 3rd-order tensor contractions, takes 0.1 mJ (0.4 mJ) and 310 ns (3.6 ns) on the GTX GPU (DGX Superpod). 

From this, we see that a small ($N_T=100$) analog transistor network has a self-simulation advantage of around $10^4$ to $10^5$ for energy, and between 0.4 and 30 for speed. To increase this advantage (and thus, speculatively, the computational capabilities of a PNN based on the analog transistor network), it is clear that establishing longer-range coupling $M$ would be desirable.

Much like the previous section, we could increase the speed advantage simply by increasing the number of transistors. For example, with $N_T = 10^3$, and $\Delta T = 1$ ns, the number of operations required for simulation is the same as the example above, and the total energy required to run the network for $\Delta T$ is the same (since $N_T \Delta T$ was held constant). However, in this case the electrical circuit has a speed advantage of 310 (4) over the GTX GPU (DGX Superpod). Obviously there is a limit to how much this can be exploited, since the evolution time of the circuit cannot be made arbitrarily short if we require non-trivial physical transformations and long-range coupling. Nonetheless, this illustrates how adding parallel physical processes to a physical system can significantly enhance its potential computational power.

Just like the previous section, we could also increase the self-simulation advantages by adding additional ``parameter" transistors that do not require fast input and output, and thus do not add to the DAC and ADC costs, but instead add only a smaller cost associated with providing enough total power to compensate for the dissipation of the additional transistors. If maintaining each additional such transistor requires an additional \SI{1}{\micro W}, one could increase the self-simulation advantages by nearly $10^6$ by adding $10^6$ such transistors before the cost of these additional transistors was comparable to the DAC/ADC cost.

Finally, we note that the above discussion has neglected the details of realizing controllable linear coupling between transistors (or, in the earlier section, oscillators). Our findings should hold for coupling techniques that require no energy to maintain, such as those based on switchable bi- or multi-stable circuit elements, even if some energy is required to change their value (e.g., such as elements used in many memory devices). If constant power is required, however, its cost could be significant, since the number of connection elements is $M_1$ times larger than the number of transistors. If connections were implemented using transistors operated as voltage-controlled resistors, for example, the self-simulation advantages predicted above would be up to $M_1$ times smaller (for energy, for speed there would be no change). For very large transistor networks, this would not preclude enormous computational advantages, since the self-simulation advantages found above can very easily exceed $10^{10}$ for realistic numbers of transistors and $M_1 \approx 10^3$. Nonetheless, passive, or nearly-passive, connectivity will be an important feature of scalable PNNs. In this regard, PAT and the PNN concept should be useful in using coupling mechanisms that do not necessarily exhibit ideal linearity, low noise, high-controllable-precision, or any other number of features that would be required if devices were assumed to \textit{perfectly} obey an equation like Eqn. \ref{transNN}.

\clearpage
\section{Supplementary discussion: when can PNNs provide technological benefits?}

Drawing on the previous section and other analysis we have performed in the preparation of this work, this section attempts to provide some heuristic guidelines for the types of PNNs that could provide real benefits for computation, and the conditions under which these might be achieved. This section is not intended to be exhaustive or detailed, but merely to summarize basic considerations for obtaining actual computational or other advantages from PNNs.

Many tasks one could ask a PNN to do will involve a number of read-in operations $N_{\text{RI}}$,  and read-out operations $N_{\text{RO}}$. These account for converting the data from the electronic digital domain to the PNN domain, and from the PNN domain to the electronic digital domain. Of course, there will also be a third set of operations, those associated with actually performing the computation, which may be performed entirely or in part by the physical system. If portions of the computation are performed with conventional hardware, those operations can be counted as $N_{\text{D}}$. 

To quantify the \textit{effective} number of operations performed by the physical system, $N_{\text{P,eff}}$, we can compare the PNN's performance to an equivalent algorithm executed on a conventional computer. For example, if a PNN performs MNIST classification with a certain accuracy, we would say it performs $N_{\text{P,eff}} = N_{\text{C}}$ operations, where $N_{\text{C}}$ is the number of operations required by a digital NN that reaches the same performance. If the PNN uses some digital operations to achieve the accuracy, the physical portion's effective operations count is $N_{\text{P,eff}} = N_{\text{C}}-N_{\text{D}}$. 

For computational acceleration, we ultimately care about having the total time and energy of the PNN computation (including all parts above) be significantly smaller than the equivalent time and energy of doing the calculation purely on a digital computer. We therefore want the following two statements to be true. First, to ensure an energy benefit from the PNN:

\begin{equation}
    e_{\text{RO}}N_{\text{RO}}+e_{\text{RI}}N_{\text{RI}}+e_{\text{D}}N_{\text{D}} + E_{\text{PNN}}\ll e_{\text{D}}N_{\text{C}} ,
\end{equation}
\nobreak where $e_{\text{RO}}$, $e_{\text{RI}}$, and $e_{\text{D}}$ are the per-operation energy cost of read-out, read-in and other digital operations, and $E_{\text{PNN}}$ is the total energy cost of executing physical operations. 

Second, to ensure a speed benefit from the PNN: 

\begin{equation}
    t_{\text{RO}}N_{\text{RO}}+t_{\text{RI}}N_{\text{RI}}+t_{\text{D}}N_{\text{D}} + T_{\text{PNN}}\ll t_{\text{D}}N_{\text{C}} ,
\end{equation}
\nobreak where $t_{\text{RO}}$, $t_{\text{RI}}$, and $t_{\text{D}}$ are the per-operation time to perform read-out, read-in and other digital operations, and $T_{\text{PNN}}$ is the total time to execute physical operations. 

To a first approximation, we can assume that all digital operations, including read-in and read-out, will take the same amount of energy and time, such that $e_{\text{RO}}=e_{\text{RI}}=e_{\text{D}}$ and $t_{\text{RO}}=t_{\text{RI}}=t_{\text{D}}$. 

The basic idea to gain a benefit with a PNN for computation is, we expect, the following: One wants to make sure that the task-PNN combination is operationally bottlenecked with respect to the electronic digital computer it is partnered with.

\textbf{Operational bottleneck}: A task-PNN combination is operationally bottlenecked if the vast majority of the operations (the `complexity’ of the calculation task) occurs within the PNN, with only a small amount happening at the input (read-in), output (read-out), or on other digital operations. A task in which we read-in $N$ numbers and find the maximum of those $N$ numbers cannot be operationally bottlenecked, since the read-in consumes $N$ operations and the computation only requires $\sim N$ operations. A task in which $N$ numbers are read-in and we need to perform $\sim N^2$ operation-equivalents (i.e., on the digital computer we’d have to perform $\sim N^2$ operations), as occurs in an arbitrary matrix-vector operation, could in principle be operationally bottlenecked. 

With our assumptions above in place, such an operational bottleneck means:
\begin{equation}
  N_{\text{RO}}+N_{\text{RI}}+N_{\text{D}} \ll N_{\text{C}}.
\end{equation}

This is because the operations on the left hand side have no potential for physical advantage: they add time and energy at the rate of any digital computer operations. Some of the operations in the right hand side can be executed with significant speed and/or energy advantages by the physics of the PNN's physical modules. Even if the energy and time required for executing the physical system is zero, an unconventional physical processor can only provide a benefit to the extent the above inequality is satisfied. 

This is equivalent to saying that the vast majority of the computational operations need to be performed by the physical system, since that is where speed-ups and energy gains can be realized:

\begin{equation}
    N_{\text{P,eff}} \gg N_{\text{RO}}+N_{\text{RI}}+N_{\text{D}}.
\end{equation}

For example, with MNIST, a 196-dimensional image is sent into the PNN, and the PNN produces a 10-D output, or possibly a 1-D output. If the PNN's performance is equivalent to a 196x10 single-layer perceptron (i.e., it gets about 90\% accuracy), then the number of equivalent-operations it performs is only ~10 times higher than the number of operations required for read-in. It is thus very hard for such a simple task to gain from a PNN: the PNN will always be far from being operationally bottlenecked, and will fail to offer much benefit, if any. Concretely, $N_{\text{RO}} = 10$, $N_{\text{RI}}=196$, and $N_{\text{D}} \sim 196$ (since we will almost always have to rescale data before sending it to a physical co-processor). Meanwhile, the total number of operations is at most 1960, so the maximum possible energy gain (and speed-up) possible with a physical co-processor is $1960/(10+196+196) \approx 5$.  

If on the other hand, the PNN is performing a much harder task, then it may be possible for the number of operations performed physically to dwarf the read-in and read-out operations, and any digital overhead operations. For CIFAR10, high accuracy classification of the 3-channel, 32 by 32 images would be equivalent to 1-100 billion operations \cite{hasanpour2016lets}. The maximum possible PNN advantage is then approximately $\sim 10^9 / (32\times 32 \times 3) \approx 10^5$.
\\[\baselineskip]
There are several ways to overcome these challenges, which are mostly not exclusive.
\\[\baselineskip]
\textbf{No- or few-parameter PNNs}: Suppose the PNN is used as a powerful, parameterized feature map that can be used many times within a given calculation, or provides a complex transformation that we do not need to control to the same level of complexity (such as if the PNN does some very complicated nonlinear transformation, which we control by a number of parameters much smaller than the number of equivalent digital operations that would be required to simulate the PNN). If this feature map is useful, it can replace a lot of digital computations that would ordinarily be required for a task, and it may be justifiable to perform digital computations (such as input and output linear layers) that allow the few-parameter PNN to be used in numerous different ways within a hybrid digital-PNN computation. 

This type of PNN resembles reservoir computing or kernel methods, and indeed, the success of these methods provides some motivation as to why it would be beneficial. However, it is important to appreciate why this is not equivalent to reservoir computing. In reservoir computing, the physical feature map is essentially untrained, and cannot be backpropagated through. Thus, any digital pre-processing prior to the physical feature map cannot be efficiently trained. In contrast, by using PAT, one may backpropagate through the physical feature map, and thus optimize pre-processing (such as the linear input layer used in Fig. 4K-O in the main article) to maximally utilize the physical feature map. Using PAT, one may use sequences of physical feature maps, possibly with digital operations between them, and train all parameters with backpropagation. 

At first glance, the optical Fourier transform performed by a lens seems to be a good candidate for this. However, on closer inspection, it does not meet the requirement, because one needs to read-in and read-out $N$ samples, but the optics only provides an operations-equivalent of $N\log N$. Thus, the physical benefit is at most around $(N\log N) /3N = (\log N)/3$, which is only significant for extremely large $N$. 

A better example is a PNN that implements a large fraction of the initial layers of a large image-classification or speech-classification network, generating mainly generic features \cite{sharif2014cnn}. This could have few trainable parameters (even zero, if an input layer was used), but still provide a significant benefit as long as the task would utilize the features produced by the PNN. Here, it would still be beneficial that the PNN be differentiable, in order to learn the input layer, or in order to train a small number of parameters. 
\\\\
\textbf{Parameter-in-place PNNs}: If the PNN’s parameters must be read-in for each inference, then the number of read-in operations can quickly approach the scale of the computational operations (unless the PNN somehow only requires a few parameters). Since read-in operations are limited by the energy and speed of conventional processors, it is optimal that most of a PNN's physical parameters are stored in-place. That is, the parameters that control the PNN’s input-output transformation may be stored in the PNN itself, rather than read-in through the same channel as the inputs. If computations the PNN performs are part of a typical NN inference, then the parameters do not need to be updated as often as the inputs, and the PNN can avoid a significant read-in cost. As was noted in the last section, it is helpful (but not essential) that in-place parameters be fully passive, in order to minimize their energy cost. 
\\\\
\textbf{PNNs based on imperfect physical operators}: In order to minimize digital overhead operations, one strategy would be to design physical transformations that approximately implement specific desired operations, such as matrix-vector multiplications, or whose controllable dynamics loosely resemble neural networks, such as the oscillator and transistor networks discussed in the last section. Such operations could be very approximate, with a myriad of complex deviations from the nominal design operation. However, as physical transformations in a PNN whose parameters (including digital pre-processing) are going to be trained anyway, these imperfect physical operations would likely serve their function well enough. 
\\\\
\textbf{Physical-feedforward PNNs}: In some PNNs, the outputs can be fed directly into another PNN layer without a digital intermediary. PNNs in which the output is directly measured (such as electronic PNNs where the output is a voltage, or possibly in optical cavities or arrangements, where the full optical field is fed-forward to subsequent physical processing) are in this category, as are PNNs which have a memory of previous inputs and states (as in optical cavities or other resonators). In this case, one can cascade PNNs over many layers, building up computational complexity, without incurring read-in or read-out costs except at the very beginning and end.

There are two other important scenarios where PNNs can have benefits, but that are broader than just computational energy or speed. 
\\\\
\textbf{Physical-data-domain-\textit{input} PNNs}: In some PNNs, the input data will already be in the physical domain of the PNN, such as light entering an optical PNN. In this case, the PNN not only avoids the read-in cost, but can additionally perform operations that a traditional device, which would first read-in the data to the electronic domain and then process it, could not. As in compressed sensing single-pixel cameras using spatial light modulators \cite{Duarte2008singlepixelimaging}, this could allow for higher-speed, higher-resolution, or lower-energy costs for performing measurements (either by improving the nature of the measurement itself, or by performing parts of the post-processing physically). In other words, one could use PAT to train physical `smart sensors'. 
\\\\
\textbf{Physical-\textit{output} PNNs}: PNNs may also provide a means of producing a desired physical quantity or functionality, in the domain of the physical system, rather than a computation result that is ultimately utilized in a downstream digital computer. In PNNs, the final output is normally obtained from measuring the physical system. However, we can instead treat the final state of the physical system as the output, and the measurement of it as merely an assessment of that physical state. For example, a nonlinear optical PNN may be trained to produce a desired output pulse, which would be assessed based on measurements of that pulse (such as the power spectrum used in our optical SHG-PNN). Alternatively, the physical functionality of the produced physical state might be assessed, such as by measuring the effects of the produced pulse on a chemical, the optical functionality of a complex nanostructure  \cite{peurifoy2018nanophotonic}, or the behavior of a small robot \cite{howison2021reality}. In all cases, measurements that assess the physical state need to be performed and transferred to the digital domain, in order to facilitate PAT, but the true output of the PNN is the physical object and its state. Thus, in PNNs of this kind, as with the `smart sensors' described above, PAT allows the backpropagation algorithm for gradient descent to be applied to produce a desired physical state or functionality, in the physical domain of the PNN. 

\bibliographystyle{npjqi}
\bibliography{sm_references}